%% file: main.tex
\documentclass{article}

    \usepackage[preprint]{neurips_2023}

\usepackage{natbib}
\setcitestyle{numbers,square}

\usepackage[utf8]{inputenc} 
\usepackage[T1]{fontenc}    
\usepackage{url}            
\usepackage{color, soul}

\usepackage{booktabs}       
\usepackage{amsfonts}       
\usepackage{nicefrac}       
\usepackage{microtype}      
\usepackage[dvipsnames]{xcolor}         
\usepackage{graphicx}
\usepackage{csquotes}
\usepackage{amsmath}
\usepackage{enumitem}       
\usepackage{blindtext} 
\usepackage{epigraph} 
\usepackage[colorlinks, linkcolor=RoyalBlue, anchorcolor=BrickRed, citecolor=RoyalBlue, urlcolor=RoyalBlue]{hyperref}

\usepackage{adjustbox}
\usepackage{tikz}
\usepackage{pgfplots}
\pgfplotsset{compat=1.18}
\usepackage[edges]{forest}
\definecolor{framework-blue}{RGB}{47, 85, 151}

\definecolor{content-yellow}{RGB}{255, 230, 153}
\definecolor{framework-yellow}{RGB}{255, 255, 255}
\definecolor{content-orange}{RGB}{251, 229, 215}
\definecolor{framework-orange}{RGB}{248, 203, 175}
\definecolor{content-gray}{RGB}{237, 237, 237}
\definecolor{framework-gray}{RGB}{166, 166, 166}

\definecolor{paired-light-blue}{RGB}{198, 219, 239}
\definecolor{paired-dark-blue}{RGB}{49, 130, 188}
\definecolor{paired-light-orange}{RGB}{251, 208, 162}
\definecolor{paired-dark-orange}{RGB}{230, 85, 12}
\definecolor{paired-light-green}{RGB}{199, 233, 193}
\definecolor{paired-dark-green}{RGB}{49, 163, 83}
\definecolor{paired-light-purple}{RGB}{218, 218, 235}
\definecolor{paired-dark-purple}{RGB}{117, 107, 176}
\definecolor{paired-light-gray}{RGB}{217, 217, 217}
\definecolor{paired-dark-gray}{RGB}{99, 99, 99}
\definecolor{paired-light-pink}{RGB}{222, 158, 214}
\definecolor{paired-dark-pink}{RGB}{123, 65, 115}
\definecolor{paired-light-red}{RGB}{231, 150, 156}
\definecolor{paired-dark-red}{RGB}{131, 60, 56}
\definecolor{paired-light-yellow}{RGB}{231, 204, 149}
\definecolor{paired-dark-yellow}{RGB}{141, 109, 49}
\tikzset{%
    parent/.style = {align=center,text width=2.5cm,rounded corners=3pt, line width=0.3mm, fill=gray!10,draw=gray!80},
    child/.style = {align=center,text width=2.3cm,rounded corners=3pt, fill=blue!10,draw=blue!80,line width=0.3mm},
    grandchild/.style = {align=center,text width=2cm,rounded corners=3pt},
    greatgrandchild/.style = {align=center,text width=1.5cm,rounded corners=3pt},
    greatgrandchild2/.style = {align=center,text width=1.5cm,rounded corners=3pt},    
    referenceblock/.style =  {align=center,text width=1.5cm,rounded corners=2pt},
    brain/.style = {align=center,text width=2.2cm,rounded corners=3pt, fill=white,draw=framework-blue,line width=0.3mm},   
    brain_work/.style = {align=center, text width=4.5cm,rounded corners=3pt, fill=white,draw=framework-blue,line width=0.3mm},
    perception/.style= {align=center,text width=2.2cm,rounded corners=3pt, fill=white,draw=framework-blue,line width=0.3mm},
    perception_work/.style= {align=center, text width=4.5cm,rounded corners=3pt, fill=white,draw=framework-blue,line width=0.3mm}, 
    action/.style= {align=center,text width=2.2cm,rounded corners=3pt, fill=white,draw=framework-blue,line width=0.3mm},
    action_work/.style= {align=center, text width=4.5cm,rounded corners=3pt, fill=white,draw=framework-blue,line width=0.3mm},
    single_agent/.style= {align=center,text width=2.2cm,rounded corners=3pt, fill=white,draw=framework-blue,line width=0.3mm},
    single_agent_work/.style= {align=center, text width=4.5cm,rounded corners=3pt, fill=white,draw=framework-blue,line width=0.3mm},
    multi_agent/.style= {align=center,text width=2.2cm,rounded corners=3pt, fill=white,draw=framework-blue,line width=0.3mm},
    multi_agent_work/.style= {align=center, text width=4.5cm,rounded corners=3pt, fill=white,draw=framework-blue,line width=0.3mm},
    human_agent/.style= {align=center,text width=2.2cm,rounded corners=3pt, fill=white,draw=framework-blue,line width=0.3mm},
    human_agent_work/.style= {align=center, text width=4.5cm,rounded corners=3pt, fill=white,draw=framework-blue,line width=0.3mm},
    behavior_and_personality/.style= {align=center,text width=2.2cm,rounded corners=3pt, fill=white,draw=framework-blue,line width=0.3mm},
    behavior_and_personality_work/.style= {align=center, text width=4.5cm,rounded corners=3pt, fill=white,draw=framework-blue,line width=0.3mm},
    society_environment/.style= {align=center,text width=2.2cm,rounded corners=3pt, fill=white,draw=framework-blue,line width=0.3mm},
    society_environment_work/.style= {align=center, text width=4.5cm,rounded corners=3pt, fill=white,draw=framework-blue,line width=0.3mm},
    society_simulation/.style= {align=center,text width=2.2cm,rounded corners=3pt, fill=white,draw=framework-blue,line width=0.3mm},
    society_simulation_work/.style= {align=center, text width=4.5cm,rounded corners=3pt, fill=white,draw=framework-blue,line width=0.3mm},
}

\title{The Rise and Potential of Large Language Model Based Agents: A Survey}

\author{%
    Zhiheng Xi$^*$$^\dag$\thanks{$^\dag${ }{ }Correspondence to: zhxi22@m.fudan.edu.cn, \{qz, tgui\}@fudan.edu.cn }\thanks{$^*${ }{ }Equal Contribution.}, Wenxiang Chen$^*$, Xin Guo$^*$, Wei He$^*$, Yiwen Ding$^*$,  Boyang Hong$^*$, \\ \textbf{Ming Zhang$^*$, Junzhe Wang$^*$, Senjie Jin$^*$, Enyu Zhou$^*$,} \\ \\ \textbf{Rui Zheng,  Xiaoran Fan, Xiao Wang, Limao Xiong, Yuhao Zhou, Weiran Wang,} \\ 
    \textbf{ Changhao Jiang, Yicheng Zou, Xiangyang Liu, Zhangyue Yin,} 
    \\ \\
    \textbf{Shihan Dou, Rongxiang Weng, Wensen Cheng,}
    \\ \\ \textbf{Qi Zhang$^\dag$, Wenjuan Qin, Yongyan Zheng, Xipeng Qiu, Xuanjing Huang and Tao Gui$^\dag$}
  \\ \\
  \large Fudan NLP Group
}

\makeatletter
\def\thanks#1{\protected@xdef\@thanks{\@thanks
        \protect\footnotetext{#1}}}
\makeatother

\begin{document}
\maketitle

\begin{abstract}

For a long time, humanity has pursued artificial intelligence (AI) equivalent to or surpassing the human level, with AI agents considered a promising vehicle for this pursuit. AI agents are artificial entities that sense their environment, make decisions, and take actions. 
Many efforts have been made to develop intelligent agents, but they mainly focus on advancement in algorithms or training strategies to enhance specific capabilities or performance on particular tasks. 
Actually, what the community lacks is a general and powerful model to serve as a starting point for designing AI agents that can adapt to diverse scenarios. 
Due to the versatile capabilities they demonstrate, large language models (LLMs) are regarded as potential sparks for Artificial General Intelligence (AGI), offering hope for building general AI agents. 
Many researchers have leveraged LLMs as the foundation to build AI agents and have achieved significant progress. 
In this paper, we perform a comprehensive survey on LLM-based agents.
We start by tracing the concept of agents from its philosophical origins to its development in AI, and explain why LLMs are suitable foundations for agents. 
Building upon this, we present a general framework for LLM-based agents, comprising three main components: brain, perception, and action, and the framework can be tailored for different applications. 
Subsequently, we explore the extensive applications of LLM-based agents in three aspects: single-agent scenarios, multi-agent scenarios, and human-agent cooperation. 
Following this, we delve into agent societies, exploring the behavior and personality of LLM-based agents, the social phenomena that emerge from an agent society, and the insights they offer for human society. 
Finally, we discuss several key topics and open problems within the field. A repository for the related papers at \href{https://github.com/WooooDyy/LLM-Agent-Paper-List}{https://github.com/WooooDyy/LLM-Agent-Paper-List}.

\end{abstract}
\newpage
{
  \hypersetup{linkcolor=RoyalBlue, linktoc=page}
  \tableofcontents
}
\clearpage

\section{Introduction}

\begin{quote}
\textit{``If they find a parrot who could answer to everything, I would claim it to be an intelligent being without hesitation.''}

\hspace*{\fill}---Denis Diderot, 1875
\end{quote}

Artificial Intelligence (AI) is a field dedicated to designing and developing systems that can replicate human-like intelligence and abilities \cite{russell2010artificial}.
As early as the 18th century, philosopher Denis Diderot introduced the idea that if a parrot could respond to every question, it could be considered intelligent \cite{diderot1911diderot}. 
While Diderot was referring to living beings, like the parrot, his notion highlights the profound concept that a highly intelligent organism could resemble human intelligence.
In the 1950s, Alan Turing expanded this notion to artificial entities and proposed the renowned Turing Test  \cite{turing2009computing}. This test is a cornerstone in AI and aims to explore whether machines can display intelligent behavior comparable to humans. These AI entities are often termed ``agents'', forming the essential building blocks of AI systems.
Typically in AI, an agent refers to an artificial entity capable of perceiving its surroundings using sensors, making decisions, and then taking actions in response using actuators \cite{russell2010artificial,DBLP:journals/ker/WooldridgeJ95}.


The concept of agents originated in Philosophy, with roots tracing back to thinkers like Aristotle and Hume \cite{sep-agency}.
It describes entities possessing desires, beliefs, intentions, and the ability to take actions \cite{sep-agency}.
This idea transitioned into computer science, intending to enable computers to understand users' interests and autonomously perform actions on their behalf \cite{DBLP:phd/us/Agha85,green1997software,DBLP:journals/cacm/GeneserethK94}. 
As AI advanced, the term ``agent'' found its place in AI research to depict entities showcasing intelligent behavior and possessing qualities like autonomy, reactivity, pro-activeness, and social ability \cite{DBLP:journals/ker/WooldridgeJ95,DBLP:journals/logcom/Goodwin95}.
Since then, the exploration and technical advancement of agents have become focal points within the AI community \cite{russell2010artificial,padgham2005developing}. AI agents are now acknowledged as a pivotal stride towards achieving Artificial General Intelligence (AGI) \footnote{Also known as Strong AI.}, as they encompass the potential for a wide range of intelligent activities \cite{DBLP:journals/ker/WooldridgeJ95,DBLP:conf/law/Shoham92,hutter2004universal}. 

From the mid-20th century, significant strides were made in developing smart AI agents as research delved deep into their design and advancement \cite{DBLP:conf/ijcai/FikesN71,DBLP:conf/ijcai/Sacerdoti73,brooks1991intelligence,maes1990designing,ribeiro2002reinforcement,kaelbling1996reinforcement}.
However, these efforts have predominantly focused on enhancing specific capabilities, such as symbolic reasoning, or mastering particular tasks like Go or Chess \cite{Guha_Lenat_1994,kaelbling1987architecture,sutton2018reinforcement}.
Achieving a broad adaptability across varied scenarios remained elusive.
Moreover, previous studies have placed more emphasis on the design of algorithms and training strategies, overlooking the development of the model's inherent general abilities like knowledge memorization, long-term planning, effective generalization, and efficient interaction \cite{DBLP:journals/corr/abs-2304-03442,DBLP:journals/corr/abs-2305-13246}. 
Actually, enhancing the inherent capabilities of the model is the pivotal factor for advancing the agent further, and the domain is in need of a powerful foundational model endowed with a variety of key attributes mentioned above to serve as a starting point for agent systems.

\begin{figure}[htbp]
    \centering
    \includegraphics[width=1.0\textwidth]{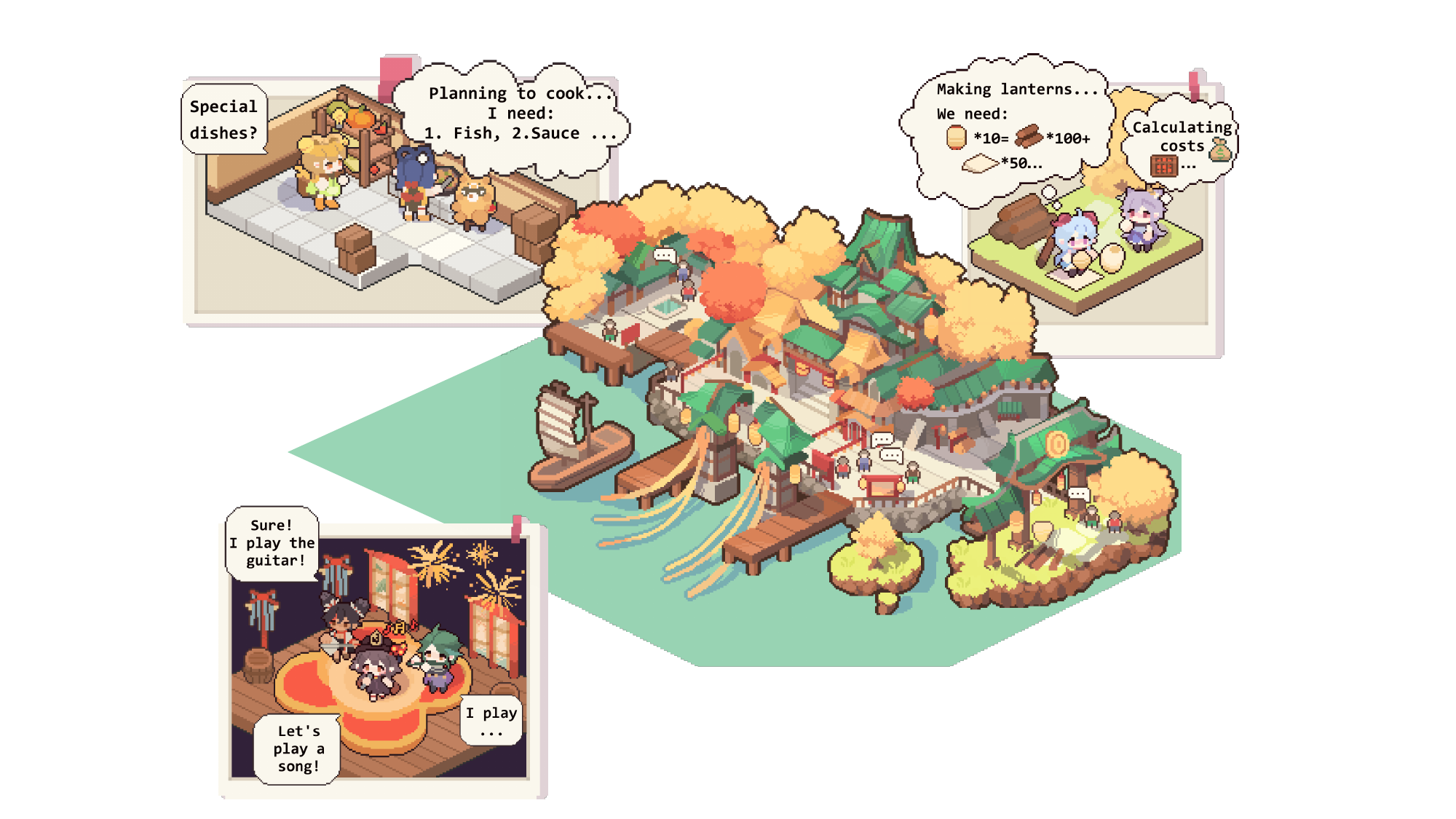}
    \caption{Scenario of an envisioned society composed of AI agents, in which humans can also participate. The above image depicts some specific scenes within society. In the kitchen, one agent orders dishes, while another agent is responsible for planning and solving the cooking task. At the concert, three agents are collaborating to perform in a band. Outdoors, two agents are discussing lantern-making, planning the required materials, and finances by selecting and using tools. Users can participate in any of these stages of this social activity.}
    \label{fig: genshin_fig}
\end{figure}

The development of large language models (LLMs) has brought a glimmer of hope for the further development of agents \cite{DBLP:conf/nips/Ouyang0JAWMZASR22,DBLP:journals/corr/abs-2303-08774,DBLP:journals/tmlr/WeiTBRZBYBZMCHVLDF22}, and significant progress has been made by the community \cite{DBLP:journals/corr/abs-2304-03442,DBLP:journals/corr/abs-2305-16960,DBLP:journals/corr/abs-2309-02427,weng2023prompt}. According to the notion of World Scope (WS) \cite{DBLP:conf/emnlp/BiskHTABCLLMNPT20} which encompasses five levels that depict the research progress from NLP to general AI (i.e., Corpus, Internet, Perception, Embodiment, and Social), the pure LLMs are built on the second level with internet-scale textual inputs and outputs. 
Despite this, LLMs have demonstrated powerful capabilities in knowledge acquisition, instruction comprehension, generalization, planning, and reasoning, while displaying effective natural language interactions with humans. These advantages have earned LLMs the designation of sparks for AGI \cite{DBLP:journals/corr/abs-2303-12712}, making them highly desirable for building intelligent agents to foster a world where humans and agents coexist harmoniously \cite{DBLP:journals/corr/abs-2304-03442}. 
Starting from this, if we elevate LLMs to the status of agents and equip them with an expanded perception space and action space, they have the potential to reach the third and fourth levels of WS. Furthermore, these LLMs-based agents can tackle more complex tasks through cooperation or competition, and emergent social phenomena can be observed when placing them together, potentially achieving the fifth WS level. As shown in Figure \ref{fig: genshin_fig}, we envision a harmonious society composed of AI agents where human can also participate.


In this paper, we present a comprehensive and systematic survey focusing on LLM-based agents, attempting to investigate the existing studies 
and prospective avenues in this burgeoning field. 
To this end, we begin by delving into crucial \textbf{background} information (\S \ \ref{sec:Background}). 
In particular, we commence by tracing the origin of AI agents from philosophy to the AI domain, along with a brief overview of the debate surrounding the existence of artificial agents (\S \ \ref{sec:Origin of AI Agent}).
Next, we take the lens of technological trends to provide a concise historical review of the development of AI agents (\S \ \ref{sec:Technological Trends in Agent Research}). 
Finally, we delve into an in-depth introduction of the essential characteristics of agents and elucidate why large language models are well-suited to serve as the main component of brains or controllers for AI agents (\S \ \ref{sec:Key Characteristics of Agent & Why LLMs Are Suitable Agent Brains?}).

Inspired by the definition of the agent, we present a general conceptual \textbf{framework} for the LLM-based agents with three key parts: \textbf{brain, perception, and action} (\S \ \ref{sec:The Birth of An Agent: Construction of LLM-based Agents}), and the framework can be tailored to suit different applications.
We first introduce the brain, which is primarily composed of a large language model  (\S \ \ref{sec:Brain}).
Similar to humans, the brain is the core of an AI agent because it not only stores crucial memories, information, and knowledge but also undertakes essential tasks of information processing, decision-making, reasoning, and planning. It is the key determinant of whether the agent can exhibit intelligent behaviors.
Next, we introduce the perception module (\S \ \ref{sec:Perception}). For an agent, this module serves a role similar to that of sensory organs for humans. Its primary function is to expand the agent's perceptual space from text-only to a multimodal space that includes diverse sensory modalities like text, sound, visuals, touch, smell, and more. This expansion enables the agent to better perceive information from the external environment.
Finally, we present the action module for expanding the action space of an agent (\S \ \ref{sec:Action}). Specifically, we expect the agent to be able to possess textual output, take embodied actions, and use tools so that it can better respond to environmental changes and provide feedback, and even alter and shape the environment.


After that, we provide a detailed and thorough introduction to the \textbf{practical applications} of LLM-based agents and elucidate the foundational design pursuit\textemdash{``Harnessing AI for good''} (\S \ \ref{sec:Agents in Practice:  Harnessing AI for Good}).
To start, we delve into the current applications of a single agent and discuss their performance in text-based tasks and simulated exploration environments, with a highlight on their capabilities in handling specific tasks, driving innovation, and exhibiting human-like survival skills and adaptability (\S \ \ref{sec:General Ability of Single Agent}).
Following that, we take a retrospective look at the development history of multi-agents. We introduce the interactions between agents in LLM-based multi-agent system applications, where they engage in collaboration, negotiation, or competition. Regardless of the mode of interaction, agents collectively strive toward a shared objective (\S \ \ref{sec:Collaborative Potential of Multi Agents}).
Lastly, considering the potential limitations of LLM-based agents in aspects such as privacy security, ethical constraints, and data deficiencies, we discuss the human-agent collaboration. We summarize the paradigms of collaboration between agents and humans: the instructor-executor paradigm and the equal partnership paradigm, along with specific applications in practice (\S \ \ref{sec:Interactive Cooperation between Human-Agent}).

Building upon the exploration of practical applications of LLM-based agents, we now shift our focus to the concept of the ``\textbf{Agent Society}'', examining the intricate interactions between agents and their surrounding environments (\S \ \ref{sec:Agent Society}).
This section begins with an investigation into whether these agents exhibit human-like behavior and possess corresponding personality (\S \ref{sec:Behavior and Personality}). 
Furthermore, we introduce the social environments within which the agents operate, including text-based environment, virtual sandbox, and the physical world (\S \ref{sec:Environment for Agent Society}). 
Unlike the previous section (\S \ \ref{sec:Perception}), here we will focus on diverse types of the environment rather than how the agents perceive it.
Having established the foundation of agents and their environments, we proceed to unveil the simulated societies that they form  (\S \ref{sec:Society Simulation}). 
We will discuss the construction of a simulated society, and go on to examine the social phenomena that emerge from it.
Specifically, we will emphasize the lessons and potential risks inherent in simulated societies.


Finally, we discuss a range of key \textbf{topics} (\S \ \ref{sec:Discussion}) and open problems within the field of LLM-based agents: (1) the mutual benefits and inspirations of the LLM research and the agent research, where we demonstrate that the development of LLM-based agents has provided many opportunities for both agent and LLM communities (\S \ \ref{sec:Mutual Benefits of LLM Research and Agent Research}); 
(2) existing evaluation efforts and some prospects for LLM-based agents from four dimensions, including utility, sociability, values and the ability to continually evolve (\S \ \ref{sec:Evaluation for LLM-based Agents}); 
(3) potential risks of LLM-based agents, where we discuss adversarial robustness and trustworthiness of LLM-based agents. We also include the discussion of some other risks like misuse, unemployment and the threat to the well-being of the human race (\S \ \ref{sec:Security, Trustworthy And Other Potential Challenges of LLM-based Agents}); 
(4) scaling up the number of agents, where we discuss the potential advantages and challenges of scaling up agent counts, along with the approaches of pre-determined and dynamic scaling (\S \ \ref{sec:Scaling Up the Number of Agents}); (5) several open problems, such as the debate over whether LLM-based agents represent a potential path to AGI, challenges from virtual simulated environment to physical environment, collective Intelligence in AI agents, and Agent as a Service (\S \ \ref{sec:Open Problems}). After all, we hope this paper could provide inspiration to the researchers and practitioners from relevant fields.

\section{Background}\label{sec:Background}

In this section, we provide crucial background information to lay the groundwork for the subsequent content (\S \ \ref{sec:Origin of AI Agent}). 
We first discuss the origin of AI agents, from philosophy to the realm of AI, coupled with a discussion of the discourse regarding the existence of artificial agents (\S \ \ref{sec:Technological Trends in Agent Research}). 
Subsequently, we summarize the development of AI agents through the lens of technological trends. Finally, we introduce the key characteristics of agents and demonstrate why LLMs are suitable to serve as the main part of the brains of AI agents (\S \ \ref{sec:Key Characteristics of Agent & Why LLMs Are Suitable Agent Brains?}).

\subsection{Origin of AI Agent}\label{sec:Origin of AI Agent}
``Agent'' is a concept with a long history that has been explored and interpreted in many fields.
Here, we first explore its origins in philosophy, discuss whether artificial products can possess agency in a philosophical sense, and examine how related concepts have been introduced into the field of AI.

\paragraph{Agent in philosophy.}
The core idea of an agent has a historical background in philosophical discussions, with its roots traceable to influential thinkers such as Aristotle and Hume, among others \cite{sep-agency}.  
In a general sense, an ``agent'' is an entity with the capacity to act, and the term ``agency'' denotes the exercise or manifestation of this capacity \cite{sep-agency}.
While in a narrow sense, ``agency'' is usually used to refer to the performance of intentional actions; and correspondingly, the term ``agent'' denotes entities that possess desires, beliefs, intentions, and the ability to act \cite{anscombe2000intention,60a9dd9a-e48a-3fcf-87dd-d5c158406fc6,Davidson1971-DAVIA-2,dennett1988precis}. 
Note that agents can encompass not only individual human beings but also other entities in both the physical and virtual world.
Importantly, the concept of an agent involves individual autonomy, granting them the ability to exercise volition, make choices, and take actions, rather than passively reacting to external stimuli.



\paragraph{From the perspective of philosophy, is artificial entities capable of agency?}
In a general sense, if we define agents as entities with the capacity to act, AI systems do exhibit a form of agency \cite{sep-agency}. 
However, the term agent is more usually used to refer to entities or subjects that possess consciousness, intentionality, and the ability to act \cite{anscombe2000intention,60a9dd9a-e48a-3fcf-87dd-d5c158406fc6,Davidson1971-DAVIA-2}. Within this framework, it's not immediately clear whether artificial systems can possess agency, as it remains uncertain whether they possess internal states that form the basis for attributing desires, beliefs, and intentions. 
Some people argue that attributing psychological states like intention to artificial agents is a form of anthropomorphism and lacks scientific rigor \cite{sep-agency,barandiaran2009defining}. 
As Barandiaran et al. \cite{barandiaran2009defining} stated, “Being specific about the requirements for agency has told us a lot about how much is still needed for the development of artificial forms of agency.”
In contrast, there are also researchers who believe that, in certain circumstances, employing the intentional stance (that is, interpreting agent behavior in terms of intentions) can provide a better description, explanation and abstraction of the actions of artificial agents, much like it is done for humans \cite{DBLP:conf/law/Shoham92,mccarthy1979ascribing,rosenschein1986synthesis}.


With the advancement of language models, the potential emergence of artificial intentional agents appears more promising \cite{DBLP:conf/nips/Ouyang0JAWMZASR22,DBLP:journals/corr/abs-2303-08774,radford2018improving,radford2019language,DBLP:conf/nips/BrownMRSKDNSSAA20}. 
In a rigorous sense, language models merely function as conditional probability models, using input to predict the next token \cite{DBLP:conf/naacl/LinJLGE21}. Different from this, humans incorporate social and perceptual context, and speak according to their mental states \cite{tomasello2005constructing,bloom2002children}. 
Consequently, some researchers argue that the current paradigm of language modeling is not compatible with the intentional actions of an agent \cite{DBLP:conf/emnlp/BiskHTABCLLMNPT20,zwaan2005embodied}. 
However, there are also researchers who propose that language models can, in a narrow sense, serve as models of agents \cite{DBLP:conf/emnlp/Andreas22,DBLP:journals/corr/abs-2306-12672}. They argue that during the process of context-based next-word prediction, current language models can sometimes infer approximate, partial representations of the beliefs, desires, and intentions held by the agent who generated the context. With these representations, the language models can then generate utterances like humans. To support their viewpoint, they conduct experiments to provide some empirical evidence \cite{DBLP:conf/emnlp/Andreas22,DBLP:journals/corr/RadfordJS17,DBLP:conf/acl/LiNA20}.

\paragraph{Introduction of agents into AI.}

It might come as a surprise that researchers within the mainstream AI community devoted relatively minimal attention to concepts related to agents until the mid to late 1980s. 
Nevertheless, there has been a significant surge of interest in this topic within the realms of computer science and artificial intelligence communities since then \cite{DBLP:journals/jasis/MukhopadhyaySHB86,DBLP:journals/ras/Maes90a,nilsson1992toward,muller1994modelling}. 
As Wooldridge et al. \cite{DBLP:journals/ker/WooldridgeJ95} stated, we can define AI by saying that it is a subfield of computer science that aims to design and build computer-based agents that exhibit aspects of intelligent behavior. So we can treat ``agent'' as a central concept in AI. When the concept of agent is introduced into the field of AI, its meaning undergoes some changes. In the realm of Philosophy, an agent can be a human, an animal, or even a concept or entity with autonomy \cite{sep-agency}. 
However, in the field of artificial intelligence, an agent is a computational entity \cite{DBLP:journals/ker/WooldridgeJ95,green1997software}. 
Due to the seemingly metaphysical nature of concepts like consciousness and desires for computational entities \cite{DBLP:conf/law/Shoham92}, and given that we can only observe the behavior of the machine, many AI researchers, including Alan Turing, suggest temporarily setting aside the question of whether an agent is ``actually'' thinking or literally possesses a ``mind'' \cite{turing2009computing}. Instead, researchers employ other attributes to help describe an agent, such as properties of autonomy, reactivity, pro-activeness and social ability \cite{DBLP:journals/ker/WooldridgeJ95,DBLP:journals/logcom/Goodwin95}.
There are also researchers who held that intelligence is ``in the eye of the beholder''; it is not an innate, isolated property \cite{brooks1991intelligence,maes1990designing,brooks1986robust,brooks2018intelligence}.
In essence, an AI agent is not equivalent to a philosophical agent; rather, it is a concretization of the philosophical concept of an agent in the context of AI. 
In this paper, we treat AI agents as artificial entities that are capable of perceiving their surroundings using sensors, making decisions, and then taking actions in response using actuators \cite{russell2010artificial,DBLP:journals/ker/WooldridgeJ95}.


\subsection{Technological Trends in Agent Research}\label{sec:Technological Trends in Agent Research}

The evolution of AI agents has undergone several stages, and here we take the lens  of technological trends to review its development briefly.

\paragraph{Symbolic Agents.}
In the early stages of artificial intelligence research, the predominant approach utilized was symbolic AI, characterized by its reliance on symbolic logic \cite{DBLP:journals/cacm/NewellS76,DBLP:books/mk/Ginsberg93}. This approach employed logical rules and symbolic representations to encapsulate knowledge and facilitate reasoning processes. 
Early AI agents were built based on this approach \cite{DBLP:books/daglib/0066946}, and they primarily focused on two problems: the transduction problem and the representation/reasoning problem \cite{shardlow1990action}. 
These agents are aimed to emulate human thinking patterns. They possess explicit and interpretable reasoning frameworks, and due to their symbolic nature, they exhibit a high degree of expressive capability \cite{DBLP:conf/ijcai/FikesN71,DBLP:conf/ijcai/Sacerdoti73,DBLP:conf/ijcai/Sacerdoti75}. A classic example of this approach is knowledge-based expert systems. 
However, symbolic agents faced limitations in handling uncertainty and large-scale real-world problems \cite{Guha_Lenat_1994,kaelbling1987architecture}. Additionally, due to the intricacies of symbolic reasoning algorithms, it was challenging to find an efficient algorithm capable of producing meaningful results within a finite timeframe \cite{kaelbling1987architecture,russell1991right}.

\paragraph{Reactive agents.}
Different from symbolic agents, reactive agents do not use complex symbolic reasoning. Instead, they primarily focus on the interaction between the agent and its environment, emphasizing quick and real-time responses \cite{brooks1991intelligence,maes1990designing,kaelbling1987architecture,DBLP:conf/ijcai/Schoppers87,DBLP:journals/trob/Brooks86}. 
These agents are mainly based on a sense-act loop, efficiently perceiving and reacting to the environment. The design of such agents prioritizes direct input-output mappings rather than intricate reasoning and symbolic operations \cite{nilsson1992toward}. 
However, Reactive agents also have limitations. They typically require fewer computational resources, enabling quicker responses, but they might lack complex higher-level decision-making and planning capabilities.

\paragraph{Reinforcement learning-based agents.}

With the improvement of computational capabilities and data availability, along with a growing interest in simulating interactions between intelligent agents and their environments, researchers have begun to utilize reinforcement learning methods to train agents for tackling more challenging and complex tasks \cite{ribeiro2002reinforcement,kaelbling1996reinforcement,minsky1961steps,isbell2001social}.
The primary concern in this field is how to enable agents to learn through interactions with their environments, enabling them to achieve maximum cumulative rewards in specific tasks \cite{sutton2018reinforcement}. 
Initially, reinforcement learning (RL) agents were primarily based on fundamental techniques such as policy search and value function optimization, exemplified by Q-learning \cite{watkins1989learning} and SARSA \cite{rummery1994line}. 
With the rise of deep learning, the integration of deep neural networks and reinforcement learning, known as Deep Reinforcement Learning (DRL), has emerged \cite{tesauro1995temporal,li2017deep}. This allows agents to learn intricate policies from high-dimensional inputs, leading to numerous significant accomplishments like AlphaGo \cite{silver2016mastering} and DQN \cite{mnih2013playing}. 
The advantage of this approach lies in its capacity to enable agents to autonomously learn in unknown environments, without explicit human intervention. This allows for its wide application in an array of domains, from gaming to robot control and beyond. 
Nonetheless, reinforcement learning faces challenges including long training times, low sample efficiency, and stability concerns, particularly when applied in complex real-world environments \cite{sutton2018reinforcement}.

\paragraph{Agents with transfer learning and meta learning.}
Traditionally, training a reinforcement learning agent requires huge sample sizes and long training time, and lacks generalization capability \cite{DBLP:journals/corr/abs-1810-00123,DBLP:journals/corr/abs-1804-06893,justesen2018illuminating,DBLP:journals/ml/Dulac-ArnoldLML21,DBLP:conf/nips/GhoshRKZAL21}. 
Consequently, researchers have introduced transfer learning to expedite an agent's learning on new tasks \cite{DBLP:conf/atal/BrysHTN15,parisotto2015actor,DBLP:journals/corr/abs-2009-07888}. 
Transfer learning reduces the burden of training on new tasks and facilitates the sharing and migration of knowledge across different tasks, thereby enhancing learning efficiency, performance, and generalization capabilities. Furthermore, meta-learning has also been introduced to AI agents \cite{DBLP:journals/corr/DuanSCBSA16,DBLP:conf/icml/FinnAL17,DBLP:conf/nips/GuptaMLAL18,DBLP:conf/icml/RakellyZFLQ19,fakoor2019meta}. 
Meta-learning focuses on learning how to learn, enabling an agent to swiftly infer optimal policies for new tasks from a small number of samples \cite{vanschoren2018meta}. 
Such an agent, when confronted with a new task, can rapidly adjust its learning approach by leveraging acquired general knowledge and policies, consequently reducing the reliance on a large volume of samples. 
However, when there exist significant disparities between source and target tasks, the effectiveness of transfer learning might fall short of expectations and there may exist negative transfer \cite{DBLP:journals/jmlr/TaylorS09,DBLP:conf/icml/TirinzoniSPR18}.
Additionally, the substantial amount of pre-training and large sample sizes required by meta learning make it hard to establish a universal learning policy \cite{DBLP:conf/icml/FinnAL17,DBLP:journals/corr/abs-2301-08028}.

\paragraph{Large language model-based agents.}
As large language models have demonstrated impressive emergent capabilities and have gained immense popularity \cite{DBLP:conf/nips/Ouyang0JAWMZASR22,DBLP:journals/corr/abs-2303-08774,DBLP:journals/tmlr/WeiTBRZBYBZMCHVLDF22,DBLP:conf/nips/BrownMRSKDNSSAA20}, researchers have started to leverage these models to construct AI agents \cite{DBLP:journals/corr/abs-2304-03442,DBLP:journals/corr/abs-2305-16960,DBLP:journals/corr/abs-2309-02427,DBLP:journals/corr/abs-2308-11432}. Specifically, they employ LLMs as the primary component of brain or controller of these agents and expand their perceptual and action space through strategies such as multimodal perception and tool utilization \cite{DBLP:journals/corr/abs-2112-09332,DBLP:conf/iclr/YaoZYDSN023,DBLP:journals/corr/abs-2302-04761,DBLP:journals/corr/abs-2304-09842,DBLP:journals/corr/abs-2304-08354}.
These LLM-based agents can exhibit reasoning and planning abilities comparable to symbolic agents through techniques like Chain-of-Thought (CoT) and problem decomposition \cite{DBLP:conf/nips/Wei0SBIXCLZ22,DBLP:conf/nips/KojimaGRMI22,DBLP:conf/iclr/0002WSLCNCZ23,DBLP:conf/iclr/ZhouSHWS0SCBLC23,DBLP:journals/corr/abs-2305-14497,shinn2023reflexion,DBLP:journals/corr/abs-2212-04088}. They can also acquire interactive capabilities with the environment, akin to reactive agents, by learning from feedback and performing new actions \cite{DBLP:conf/acl/AkyurekAKCWT23,DBLP:journals/corr/abs-2302-12813,liu2023languages}. 
Similarly, large language models undergo pre-training on large-scale corpora and demonstrate the capacity for few-shot and zero-shot generalization, allowing for seamless transfer between tasks without the need to update parameters \cite{DBLP:conf/nips/BrownMRSKDNSSAA20,DBLP:conf/iclr/WeiBZGYLDDL22,DBLP:conf/iclr/SanhWRBSACSRDBX22,DBLP:journals/corr/abs-2210-11416}.
LLM-based agents have been applied to various real-world scenarios, such as software development \cite{DBLP:journals/corr/abs-2303-17760,DBLP:journals/corr/abs-2307-07924} and scientific research \cite{DBLP:journals/corr/abs-2304-05332}.
Due to their natural language comprehension and generation capabilities, they can interact with each other seamlessly, giving rise to collaboration and competition among multiple agents \cite{DBLP:journals/corr/abs-2303-17760,DBLP:journals/corr/abs-2307-07924,DBLP:journals/corr/abs-2305-14325,DBLP:journals/corr/abs-2305-19118}. Furthermore, research suggests that allowing multiple agents to coexist can lead to the emergence of social phenomena \cite{DBLP:journals/corr/abs-2304-03442}.



\subsection{ Why is LLM suitable as the primary component of an Agent's brain?} \label{sec:Key Characteristics of Agent & Why LLMs Are Suitable Agent Brains?}


As mentioned before, researchers have introduced several properties to help describe and define agents in the field of AI. Here, we will delve into some key properties, elucidate their relevance to LLMs, and thereby expound on why LLMs are highly suited to serve as the main part of brains of AI agents.

\paragraph{Autonomy.} Autonomy means that an agent operates without direct intervention from humans or others and possesses a degree of control over its actions and internal states \cite{DBLP:journals/ker/WooldridgeJ95,DBLP:conf/ecaiw/Castelfranchi94}. 
This implies that an agent should not only possess the capability to follow explicit human instructions for task completion but also exhibit the capacity to initiate and execute actions independently. 
LLMs can demonstrate a form of autonomy through their ability to generate human-like text, engage in conversations, and perform various tasks without detailed step-by-step instructions \cite{gravitasauto,nakajima2023babyagi}. 
Moreover, they can dynamically adjust their outputs based on environmental input, reflecting a degree of adaptive autonomy \cite{DBLP:journals/corr/abs-2305-13246,DBLP:journals/corr/abs-2305-16960,liu2023languages}. 
Furthermore, they can showcase autonomy through exhibiting creativity like coming up with novel ideas, stories, or solutions that haven't been explicitly programmed into them \cite{DBLP:conf/iui/YuanCRI22,DBLP:journals/corr/abs-2304-00008}. 
This implies a certain level of self-directed exploration and decision-making. 
Applications like Auto-GPT \cite{gravitasauto} exemplify the significant potential of LLMs in constructing autonomous agents. Simply by providing them with a task and a set of available tools, they can autonomously formulate plans and execute them to achieve the ultimate goal.

\paragraph{Reactivity.}
Reactivity in an agent refers to its ability to respond rapidly to immediate changes and stimuli in its environment \cite{DBLP:journals/logcom/Goodwin95}. 
This implies that the agent can perceive alterations in its surroundings and promptly take appropriate actions.
Traditionally, the perceptual space of language models has been confined to textual inputs, while the action space has been limited to textual outputs. 
However, researchers have demonstrated the potential to expand the perceptual space of LLMs using multimodal fusion techniques, enabling them to rapidly process visual and auditory information from the environment \cite{DBLP:journals/corr/abs-2303-08774,zhu2023minigpt,DBLP:journals/corr/abs-2306-13549}. Similarly, it's also feasible to expand the action space of LLMs through embodiment techniques \cite{DBLP:conf/icml/DriessXSLCIWTVY23,DBLP:journals/corr/abs-2305-15021} and tool usage \cite{DBLP:journals/corr/abs-2302-04761,DBLP:journals/corr/abs-2304-08354}. 
These advancements enable LLMs to effectively interact with the real-world physical environment and carry out tasks within it.
One major challenge is that LLM-based agents, when performing non-textual actions, require an intermediate step of generating thoughts or formulating tool usage in textual form before eventually translating them into concrete actions. This intermediary process consumes time and reduces the response speed. 
However, this aligns closely with human behavioral patterns, where the principle of ``think before you act'' is observed \cite{brown2013beyond,DBLP:journals/corr/abs-2305-16338}.

\paragraph{Pro-activeness.}
Pro-activeness denotes that agents don't merely react to their environments; they possess the capacity to display goal-oriented actions by proactively taking the initiative \cite{DBLP:journals/logcom/Goodwin95}. 
This property emphasizes that agents can reason, make plans, and take proactive measures in their actions to achieve specific goals or adapt to environmental changes.
Although intuitively the paradigm of next token prediction in LLMs may not possess intention or desire, research has shown that they can implicitly generate representations of these states and guide the model's inference process \cite{DBLP:conf/emnlp/Andreas22,DBLP:journals/corr/RadfordJS17,DBLP:conf/acl/LiNA20}. 
LLMs have demonstrated a strong capacity for generalized reasoning and planning. By prompting large language models with instructions like ``let's think step by step'', we can elicit their reasoning abilities, such as logical and mathematical reasoning \cite{DBLP:conf/nips/Wei0SBIXCLZ22,DBLP:conf/nips/KojimaGRMI22,DBLP:conf/iclr/0002WSLCNCZ23}. 
Similarly, large language models have shown the emergent ability of planning in forms of goal reformulation \cite{DBLP:journals/corr/abs-2305-14497,DBLP:journals/corr/abs-2302-06706}, task decomposition \cite{DBLP:conf/iclr/ZhouSHWS0SCBLC23,DBLP:journals/corr/abs-2304-11477}, and adjusting plans in response to environmental changes \cite{shinn2023reflexion,DBLP:journals/corr/abs-2302-02676}.

\paragraph{Social ability.} 
Social ability refers to an agent's capacity to interact with other agents, including humans, through some kind of agent-communication language \cite{DBLP:journals/cacm/GeneserethK94}.
Large language models exhibit strong natural language interaction abilities like understanding and generation \cite{DBLP:journals/corr/abs-2305-13246, DBLP:journals/corr/abs-2305-13711, DBLP:conf/acl/LinFKD22}. Compared to structured languages or other communication protocals, such capability enables them to interact with other models or humans in an interpretable manner. This forms the cornerstone of social ability for LLM-based agents \cite{DBLP:journals/corr/abs-2304-03442,DBLP:journals/corr/abs-2303-17760}.
Many researchers have demonstrated that LLM-based agents can enhance task performance through social behaviors such as collaboration and competition \cite{DBLP:journals/corr/abs-2303-17760,DBLP:journals/corr/abs-2305-14325,DBLP:journals/corr/abs-2305-10142,DBLP:journals/corr/abs-2307-02485}. 
By inputting specific prompts, LLMs can also play different roles, thereby simulating the social division of labor in the real world \cite{DBLP:journals/corr/abs-2307-07924}.
Furthermore, when we place multiple agents with distinct identities into a society, emergent social phenomena can be observed \cite{DBLP:journals/corr/abs-2304-03442}.

\input{sections/Sec3.Construction}

\input{sections/Sec4.Application}
\input{sections/Sec5.Society}
\input{sections/Sec6.Discussion}

\input{sections/Sec7.Conclusion}
\section*{Acknowledgements}
Thanks to Professor Guoyu Wang for carefully reviewing the ethics of the article. 
Thanks to Jinzhu Xiong for her excellent drawing skills to present an amazing performance of Figure \ref{fig: genshin_fig}.

\bibliography{main}
\bibliographystyle{nips}


\end{document}

%% file: sections/Sec3.Construction.tex
\section{The Birth of An Agent: Construction of LLM-based Agents}\label{sec:The Birth of An Agent: Construction of LLM-based Agents}

\begin{figure}[htbp]
    \centering
    \includegraphics[width=1\textwidth]
    {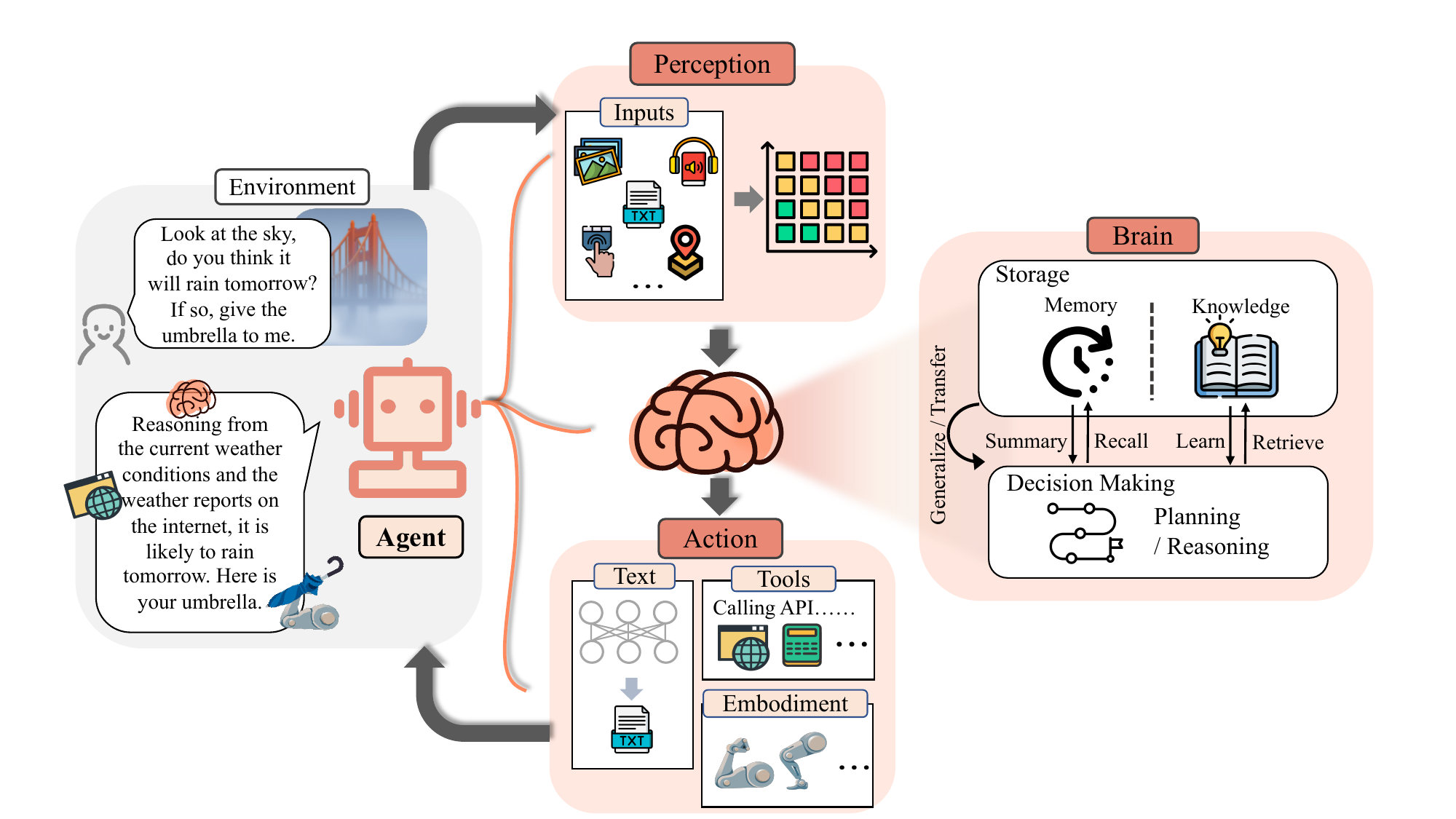}
    \caption{Conceptual framework of LLM-based agent with three components: brain, perception, and action. Serving as the controller, the brain module undertakes basic tasks like memorizing, thinking, and decision-making. The perception module perceives and processes multimodal information from the external environment, and the action module carries out the execution using tools and influences the surroundings. Here we give an example to illustrate the workflow: When a human asks whether it will rain, the perception module converts the instruction into an understandable representation for LLMs. Then the brain module begins to reason according to the current weather and the weather reports on the internet. Finally, the action module responds and hands the umbrella to the human. By repeating the above process, an agent can continuously get feedback and interact with the environment.}
    \label{fig: agent_construction}
\end{figure} 

``Survival of the Fittest'' \cite{darwin1859origin} shows that if an individual wants to survive in the external environment, he must adapt to the surroundings efficiently. 
This requires him to be cognitive, able to perceive and respond to changes in the outside world, which is consistent with the definition of ``agent'' mentioned in \S \ref{sec:Origin of AI Agent}. 
Inspired by this, we present a general conceptual framework of an LLM-based agent composed of three key parts: brain, perception, and action (see Figure \ref{fig: agent_construction}). 
We first describe the structure and working mechanism of the brain, which is primarily composed of a large language model (\S \ \ref{sec:Brain}). The brain is the core of an AI agent because it not only stores knowledge and memories but also undertakes indispensable functions like information processing and decision-making. It can present the process of reasoning and planning, and cope well with unseen tasks, exhibiting the intelligence of an agent. 
Next, we introduce the perception module (\S \ \ref{sec:Perception}). Its core purpose is to broaden the agent's perception space from a text-only domain to a multimodal sphere that includes textual, auditory, and visual modalities. This extension equips the agent to grasp and utilize information from its surroundings more effectively. 
Finally, we present the action module designed to expand the action space of an agent (\S \ \ref{sec:Action}). Specifically, we empower the agent with embodied action ability and tool-handling skills, enabling it to adeptly adapt to environmental changes, provide feedback, and even influence and mold the environment.

The framework can be tailored for different application scenarios, i.e. not every specific component will be used in all studies. 
In general, agents operate in the following workflow: First, the \textbf{perception} module, corresponding to human sensory systems such as the eyes and ears, perceives changes in the external environment and then converts multimodal information into an understandable representation for the agent. 
Subsequently, the \textbf{brain} module, serving as the control center, engages in information processing activities such as thinking, decision-making, and operations with storage including memory and knowledge. 
Finally, the \textbf{action} module, corresponding to human limbs, carries out the execution with the assistance of tools and leaves an impact on the surroundings. By repeating the above process, an agent can continuously get feedback and interact with the environment.

\subsection{Brain}\label{sec:Brain}
\input{figures/sec3_brain_mindmap}
The human brain is a sophisticated structure comprised of a vast number of interconnected neurons, capable of processing various information, generating diverse thoughts, controlling different behaviors, and even creating art and culture \cite{marshall2013discoveries}. 
Much like humans, the brain serves as the central nucleus of an AI agent, primarily composed of a large language model. 
\paragraph{Operating mechanism.}
To ensure effective communication, the ability to engage in natural language interaction (\S \ref{sec:Natural Language Interaction}) is paramount. After receiving the information processed by the perception module, the brain module first turns to storage, retrieving in knowledge (\S \ref{sec:knowledge}) and recalling from memory (\S \ref{sec:memory}). These outcomes aid the agent in devising plans, reasoning, and making informed decisions (\S \ref{sec:reasoning and planning}). Additionally, the brain module may memorize the agent's past observations, thoughts, and actions in the form of summaries, vectors, or other data structures. Meanwhile, it can also update the knowledge such as common sense and domain knowledge for future use. 
The LLM-based agent may also adapt to unfamiliar scenarios with its inherent generalization and transfer ability (\S \ref{sec:transferability and generalization}). 
In the subsequent sections, we delve into a detailed exploration of these extraordinary facets of the brain module as depicted in Figure \ref{fig:sec3_mindmap_brain}. 

\subsubsection{Natural Language Interaction}\label{sec:Natural Language Interaction}
As a medium for communication, language contains a wealth of information. In addition to the intuitively expressed content, there may also be the speaker's beliefs, desires, and intentions hidden behind it \cite{searle2007language}. Thanks to the powerful natural language understanding and generation capabilities inherent in LLMs \cite{DBLP:journals/corr/abs-2303-08774,DBLP:journals/corr/abs-2302-13971,  DBLP:journals/corr/abs-2211-05100, almazrouei2023falcon}, agents can proficiently engage in not only basic interactive conversations \cite{DBLP:journals/corr/SerbanLCP16, DBLP:journals/corr/VinyalsL15, DBLP:journals/corr/abs-2001-09977} in multiple languages \cite{DBLP:journals/corr/abs-2302-04023,DBLP:journals/corr/abs-2211-05100} but also exhibit in-depth comprehension abilities, which allow humans to easily understand and interact with agents \cite{DBLP:journals/corr/abs-2305-17066, DBLP:conf/eacl/RollerDGJWLXOSB21}. Besides, LLM-based agents that communicate in natural language can earn more trust and cooperate more effectively with humans \cite{DBLP:journals/corr/abs-2307-02485}.

\paragraph{Multi-turn interactive conversation.}
The capability of multi-turn conversation is the foundation of effective and consistent communication. As the core of the brain module, LLMs, such as GPT series \cite{radford2019language, DBLP:conf/nips/BrownMRSKDNSSAA20, DBLP:journals/corr/abs-2302-13971}, LLaMA series \cite{DBLP:journals/corr/abs-2302-13971, taori2023stanford} and T5 series \cite{DBLP:journals/corr/abs-2210-11416,raffel2020exploring}, can understand natural language and generate coherent and contextually relevant responses, which helps agents to comprehend better and handle various problems \cite{DBLP:journals/corr/abs-2304-04370}. However, even humans find it hard to communicate without confusion in one sitting, so multiple rounds of dialogue are necessary. Compared with traditional text-only reading comprehension tasks like SQuAD \cite{DBLP:conf/emnlp/RajpurkarZLL16}, multi-turn conversations (1) are interactive, involving multiple speakers, and lack continuity; (2) may involve multiple topics, and the information of the dialogue may also be redundant, making the text structure more complex \cite{DBLP:journals/corr/abs-2103-03125}. In general, the multi-turn conversation is mainly divided into three steps: (1) Understanding the history of natural language dialogue, (2) Deciding what action to take, and (3) Generating natural language responses. LLM-based agents are capable of continuously refining outputs using existing information to conduct multi-turn conversations and effectively achieve the ultimate goal \cite{DBLP:journals/corr/abs-2302-04023,DBLP:journals/corr/abs-2103-03125}.

\paragraph{High-quality natural language generation.}
Recent LLMs show exceptional natural language generation capabilities, consistently producing high-quality text in multiple languages \cite{DBLP:journals/corr/abs-2302-04023, DBLP:journals/corr/abs-2303-12528}. The coherency \cite{DBLP:conf/conll/SeePSYM19} and grammatical accuracy \cite{DBLP:journals/corr/abs-2304-01746} of LLM-generated content have shown steady enhancement, evolving progressively from GPT-3 \cite{DBLP:conf/nips/BrownMRSKDNSSAA20} to InstructGPT \cite{DBLP:conf/nips/Ouyang0JAWMZASR22}, and culminating in GPT-4 \cite{DBLP:journals/corr/abs-2303-08774}. See et al. \cite{DBLP:conf/conll/SeePSYM19} empirically affirm that these language models can ``adapt to the style and content of the conditioning text'' \cite{radford2019better}. And the results of Fang et al. \cite{DBLP:journals/corr/abs-2304-01746} suggest that ChatGPT excels in grammar error detection, underscoring its powerful language capabilities. In conversational contexts, LLMs also perform well in key metrics of dialogue quality, including content, relevance, and appropriateness \cite{DBLP:journals/corr/abs-2305-13711}. Importantly, they do not merely copy training data but display a certain degree of creativity, generating diverse texts that are equally novel or even more novel than the benchmarks crafted by humans \cite{DBLP:journals/corr/abs-2111-09509}. Meanwhile, human oversight remains effective through the use of controllable prompts, ensuring precise control over the content generated by these language models \cite{DBLP:conf/eacl/LuZZWY23}.


\paragraph{Intention and implication understanding.}
\label{intention and implication understanding}
Although models trained on the large-scale corpus are already intelligent enough to understand instructions, most are still incapable of emulating human dialogues or fully leveraging the information conveyed in language \cite{DBLP:conf/aaai/TellexKDWBTR11}. 
Understanding the implied meanings is essential for effective communication and cooperation with other intelligent agents \cite{DBLP:journals/corr/abs-2109-01355}, and enables one to interpret others' feedback. 
The emergence of LLMs highlights the potential of foundation models to understand human intentions, but when it comes to vague instructions or other implications, it poses a significant challenge for agents \cite{DBLP:journals/corr/abs-2304-08354, DBLP:journals/corr/abs-2305-14763}. 
For humans, grasping the implied meanings from a conversation comes naturally, whereas for agents, they should formalize implied meanings into a reward function that allows them to choose the option in line with the speaker's preferences in unseen contexts \cite{DBLP:conf/acl/LinFKD22}. One of the main ways for reward modeling is inferring rewards based on feedback, which is primarily presented in the form of comparisons \cite{DBLP:conf/nips/ChristianoLBMLA17} (possibly supplemented with reasons \cite{DBLP:conf/hri/BasuSD18}) and unconstrained natural language \cite{DBLP:conf/aaai/SumersHHNG21}. Another way involves recovering rewards from descriptions, using the action space as a bridge \cite{DBLP:conf/acl/LinFKD22}. Jeon et al. \cite{DBLP:conf/nips/JeonMD20} suggests that human behavior can be mapped to a choice from an implicit set of options, which helps to interpret all the information in a single unifying formalism. By utilizing their understanding of context, agents can take highly personalized and accurate action, tailored to specific requirements.

\subsubsection{Knowledge} \label{sec:knowledge}

Due to the diversity of the real world, many NLP researchers attempt to utilize data that has a larger scale. This data usually is unstructured and unlabeled \cite{DBLP:conf/naacl/HillCK16,DBLP:journals/jmlr/CollobertWBKKK11}, yet it contains enormous knowledge that language models could learn. In theory, language models can learn more knowledge as they have more parameters \cite{DBLP:journals/corr/abs-2001-08361}, and it is possible for language models to learn and comprehend everything in natural language. Research \cite{DBLP:conf/emnlp/RobertsRS20} shows that language models trained on a large-scale dataset can encode a wide range of knowledge into their parameters and respond correctly to various types of queries. Furthermore, the knowledge can assist LLM-based agents in making informed decisions \cite{DBLP:journals/expert/McShane09}. All of this knowledge can be roughly categorized into the following types:

\begin{itemize}[leftmargin=*]
    \item \textbf{Linguistic knowledge.} Linguistic knowledge \cite{DBLP:conf/emnlp/VulicPLGK20, DBLP:conf/naacl/HewittM19,DBLP:journals/ipm/RauJZ89} is represented as a system of constraints, a grammar, which defines all and only the possible sentences of the language. It includes morphology, syntax, semantics \cite{DBLP:conf/dasfaa/YangCCHL16,DBLP:conf/emnlp/BeloucifB21}, and pragmatics. Only the agents that acquire linguistic knowledge can comprehend sentences and engage in multi-turn conversations \cite{DBLP:journals/corr/abs-2103-03125}. Moreover, these agents can acquire multilingual knowledge \cite{ DBLP:journals/corr/abs-2302-04023} by training on datasets that contain multiple languages, eliminating the need for extra translation models.
    \item \textbf{Commonsense knowledge.} 
    Commonsense knowledge \cite{DBLP:conf/emnlp/SafaviK21,DBLP:journals/tacl/JiangXAN20,DBLP:conf/emnlp/MadaanZ0YN22} refers to general world facts that are typically taught to most individuals at an early age. For example, people commonly know that medicine is used for curing diseases, and umbrellas are used to protect against rain. Such information is usually not explicitly mentioned in the context. Therefore, the models lacking the corresponding commonsense knowledge may fail to grasp or misinterpret the intended meaning \cite{DBLP:journals/sigmod/TandonVM17}. Similarly, agents without commonsense knowledge may make incorrect decisions, such as not bringing an umbrella when it rains heavily.
    \item \textbf{Professional domain knowledge.} 
    Professional domain knowledge refers to the knowledge associated with a specific domain like programming \cite{DBLP:conf/pldi/Xu0NH22,DBLP:conf/icml/Lai0WZZZYFWY23,DBLP:conf/emnlp/MadaanZ0YN22}, mathematics \cite{DBLP:journals/corr/abs-2110-14168}, medicine \cite{thirunavukarasu2023large}, etc. It is essential for models to effectively solve problems within a particular domain \cite{DBLP:conf/acl/GururanganMSLBD20}. For example, models designed to perform programming tasks need to possess programming knowledge, such as code format. Similarly, models intended for diagnostic purposes should possess medical knowledge like the names of specific diseases and prescription drugs.
\end{itemize}

Although LLMs demonstrate excellent performance in acquiring, storing, and utilizing knowledge \cite{DBLP:journals/corr/abs-2204-06031}, there remain potential issues and unresolved problems. For example, the knowledge acquired by models during training could become outdated or even be incorrect from the start. A simple way to address this is retraining. However, it requires advanced data, extensive time, and computing resources. Even worse, it can lead to catastrophic forgetting \cite{DBLP:conf/aaai/KemkerMAHK18}. Therefore, some researchers\cite{DBLP:conf/emnlp/CaoAT21,DBLP:journals/corr/abs-2305-13172,DBLP:conf/icml/MitchellLBMF22} try editing LLMs to locate and modify specific knowledge stored within the models. This involved unloading incorrect knowledge while simultaneously acquiring new knowledge. Their experiments show that this method can partially edit factual knowledge, but its underlying mechanism still requires further research. Besides, LLMs may generate content that conflicts with the source or factual information \cite{DBLP:conf/icml/ShiCMSDCSZ23}, a phenomenon often referred to as hallucinations \cite{DBLP:journals/corr/abs-2309-01219}. It is one of the critical reasons why LLMs can not be widely used in factually rigorous tasks. To tackle this issue, some researchers \cite{DBLP:journals/corr/abs-2303-08896} proposed a metric to measure the level of hallucinations and provide developers with an effective reference to evaluate the trustworthiness of LLM outputs. Moreover, some researchers\cite{DBLP:journals/corr/abs-2305-14623,DBLP:journals/corr/abs-2305-11738} enable LLMs to utilize external tools\cite{DBLP:journals/corr/abs-2304-08354,DBLP:journals/corr/abs-2302-07842,DBLP:journals/corr/abs-2307-11019} to avoid incorrect knowledge. Both of these methods can alleviate the impact of hallucinations, but further exploration of more effective approaches is still needed.

\subsubsection{Memory} \label{sec:memory}
In our framework, ``memory'' stores sequences of the agent's past observations, thoughts and actions, which is akin to the definition presented by Nuxoll et al. \cite{nuxoll2007extending}. Just as the human brain relies on memory systems to retrospectively harness prior experiences for strategy formulation and decision-making, agents necessitate specific memory mechanisms to ensure their proficient handling of a sequence of consecutive tasks \cite{squire1986mechanisms,schwabe2014reconsolidation,hutter2000theory}. When faced with complex problems, memory mechanisms help the agent to revisit and apply antecedent strategies effectively. Furthermore, these memory mechanisms enable individuals to adjust to unfamiliar environments by drawing on past experiences.

With the expansion of interaction cycles in LLM-based agents, two primary challenges arise. The first pertains to the sheer length of historical records. LLM-based agents process prior interactions in natural language format, appending historical records to each subsequent input. As these records expand, they might surpass the constraints of the Transformer architecture that most LLM-based agents rely on. When this occurs, the system might truncate some content. The second challenge is the difficulty in extracting relevant memories. As agents amass a vast array of historical observations and action sequences, they grapple with an escalating memory burden. This makes establishing connections between related topics increasingly challenging, potentially causing the agent to misalign its responses with the ongoing context.

\paragraph{Methods for better memory capability.} 
Here we introduce several methods to enhance the memory of LLM-based agents.

\begin{itemize}[leftmargin=*]
    \item \textbf{Raising the length limit of Transformers.} The first method tries to address or mitigate the inherent sequence length constraints. The Transformer architecture struggles with long sequences due to these intrinsic limits. As sequence length expands, computational demand grows exponentially due to the pairwise token calculations in the self-attention mechanism. Strategies to mitigate these length restrictions encompass text truncation \cite{DBLP:conf/acl/LewisLGGMLSZ20,DBLP:conf/acl/ParkVS22,DBLP:conf/acl/ZhangWZ19}, segmenting inputs \cite{DBLP:journals/corr/abs-2305-16300,DBLP:journals/corr/abs-2210-05529}, and emphasizing key portions of text \cite{ DBLP:conf/emnlp/NieH0M22,DBLP:journals/corr/abs-2305-01625,DBLP:conf/emnlp/ManakulG21}. Some other works modify the attention mechanism to reduce complexity, thereby accommodating longer sequences \cite{DBLP:conf/nips/ZaheerGDAAOPRWY20,DBLP:conf/naacl/GuoAUONSY22,DBLP:journals/corr/abs-2303-09752,DBLP:conf/acl/RuossDGGCBLV23}. 

    \item \textbf{Summarizing memory.} The second strategy for amplifying memory efficiency hinges on the concept of memory summarization. This ensures agents effortlessly extract pivotal details from historical interactions. Various techniques have been proposed for summarizing memory. Using prompts, some methods succinctly integrate memories \cite{DBLP:journals/corr/abs-2304-13343}, while others emphasize reflective processes to create condensed memory representations \cite{DBLP:journals/corr/abs-2304-03442, DBLP:journals/corr/abs-2308-10144}. Hierarchical methods streamline dialogues into both daily snapshots and overarching summaries \cite{DBLP:journals/corr/abs-2305-10250}. Notably, specific strategies translate environmental feedback into textual encapsulations, bolstering agents' contextual grasp for future engagements \cite{DBLP:journals/corr/abs-2303-11366}. Moreover, in multi-agent environments, vital elements of agent communication are captured and retained \cite{DBLP:journals/corr/abs-2308-07201}.

    \item \textbf{Compressing memories with vectors or data structures.} By employing suitable data structures, intelligent agents boost memory retrieval efficiency, facilitating prompt responses to interactions. Notably, several methodologies lean on embedding vectors for memory sections, plans, or dialogue histories \cite{DBLP:journals/corr/abs-2307-07924,DBLP:journals/corr/abs-2305-10250,DBLP:journals/corr/abs-2305-17144,DBLP:journals/corr/abs-2308-04026}. Another approach translates sentences into triplet configurations \cite{DBLP:journals/corr/abs-2305-14322}, while some perceive memory as a unique data object, fostering varied interactions \cite{DBLP:journals/corr/abs-2308-01542}. Furthermore, ChatDB \cite{DBLP:journals/corr/abs-2306-03901} and DB-GPT \cite{DBLP:journals/corr/abs-2308-05481} integrate the LLMrollers with SQL databases, enabling data manipulation through SQL commands.
\end{itemize}

\paragraph{Methods for memory retrieval.}
When an agent interacts with its environment or users, it is imperative to retrieve the most appropriate content from its memory. This ensures that the agent accesses relevant and accurate information to execute specific actions. 
An important question arises: How can an agent select the most suitable memory? Typically, agents retrieve memories in an automated manner \cite{DBLP:journals/corr/abs-2305-10250, DBLP:journals/corr/abs-2308-04026}. A significant approach in automated retrieval considers three metrics: Recency, Relevance, and Importance. The memory score is determined as a weighted combination of these metrics, with memories having the highest scores being prioritized in the model's context \cite{DBLP:journals/corr/abs-2304-03442}. 

Some research introduces the concept of interactive memory objects, which are representations of dialogue history that can be moved, edited, deleted, or combined through summarization. Users can view and manipulate these objects, influencing how the agent perceives the dialogue \cite{DBLP:journals/corr/abs-2308-01542}. Similarly, other studies allow for memory operations like deletion based on specific commands provided by users \cite{DBLP:journals/corr/abs-2306-03901}. Such methods ensure that the memory content aligns closely with user expectations.

\subsubsection{Reasoning and Planning} \label{sec:reasoning and planning}
\paragraph{Reasoning.}

Reasoning, underpinned by evidence and logic, is fundamental to human intellectual endeavors, serving as the cornerstone for problem-solving, decision-making, and critical analysis \cite{wason1968reasoning, wason1972psychology, galotti1989approaches}. Deductive, inductive, and abductive are the primary forms of reasoning commonly recognized in intellectual endeavor \cite{DBLP:conf/acl/0009C23}. For LLM-based 
agents, like humans, reasoning capacity is crucial for solving complex tasks \cite{DBLP:journals/corr/abs-2303-08774}.

Differing academic views exist regarding the reasoning capabilities of large language models. Some argue language models possess reasoning during pre-training or fine-tuning \cite{DBLP:conf/acl/0009C23}, while others believe it emerges after reaching a certain scale in size \cite{DBLP:journals/tmlr/WeiTBRZBYBZMCHVLDF22, DBLP:journals/corr/abs-2212-09196}. 
Specifically, the representative Chain-of-Thought (CoT) method \cite{DBLP:conf/nips/Wei0SBIXCLZ22,DBLP:conf/nips/KojimaGRMI22} has been demonstrated to elicit the reasoning capacities of large language models by guiding LLMs to generate rationales before outputting the answer.
Some other strategies have also been presented to enhance the performance of LLMs like self-consistency \cite{DBLP:conf/iclr/0002WSLCNCZ23}, self-polish \cite{DBLP:journals/corr/abs-2305-14497}, self-refine \cite{DBLP:journals/corr/abs-2303-17651} and selection-inference \cite{DBLP:conf/iclr/CreswellSH23}, among others.
Some studies suggest that the effectiveness of step-by-step reasoning can be attributed to the local statistical structure of training data, with locally structured dependencies between variables yielding higher data efficiency than training on all variables \cite{DBLP:journals/corr/abs-2305-15408}.

\paragraph{Planning.}

Planning is a key strategy humans employ when facing complex challenges. For humans, planning helps organize thoughts, set objectives, and determine the steps to achieve those objectives \cite{grafman2004planning,unterrainer2006planning,zula2007integrative}.
Just as with humans, the ability to plan is crucial for agents, and central to this planning module is the capacity for reasoning \cite{bratman1988plans,DBLP:books/lib/RussellN03,fainstein2015readings}.  
This offers a structured thought process for agents based on LLMs. Through reasoning, agents deconstruct complex tasks into more manageable sub-tasks, devising appropriate plans for each \cite{sebastia2006decomposition,crosby2013automated}. 
Moreover, as tasks progress, agents can employ introspection to modify their plans, ensuring they align better with real-world circumstances, leading to adaptive and successful task execution.

Typically, planning comprises two stages: plan formulation and plan reflection.

\begin{itemize}[leftmargin=*]
    \item \textbf{Plan formulation.} During the process of plan formulation, agents generally decompose an overarching task into numerous sub-tasks, and various approaches have been proposed in this phase. Notably, some works advocate for LLM-based agents to decompose problems comprehensively in one go, formulating a complete plan at once and then executing it sequentially \cite{DBLP:conf/iclr/ZhouSHWS0SCBLC23,DBLP:conf/corl/IchterBCFHHHIIJ22,DBLP:journals/corr/abs-2305-18323,DBLP:journals/corr/abs-2211-09935}. In contrast, other studies like the CoT-series employ an adaptive strategy, where they plan and address sub-tasks one at a time, allowing for more fluidity in handling intricate tasks in their entirety \cite{DBLP:conf/nips/Wei0SBIXCLZ22,DBLP:conf/nips/KojimaGRMI22,DBLP:journals/corr/abs-2301-13379}. Additionally, some methods emphasize hierarchical planning \cite{DBLP:journals/corr/abs-2305-02412,DBLP:journals/corr/abs-2305-17390}, while others underscore a strategy in which final plans are derived from reasoning steps structured in a tree-like format. The latter approach argues that agents should assess all possible paths before finalizing a plan \cite{DBLP:conf/iclr/0002WSLCNCZ23,DBLP:journals/corr/abs-2305-10601,DBLP:journals/corr/abs-2305-14992,DBLP:conf/icml/HuangAPM22,DBLP:journals/corr/abs-2305-14992}. While LLM-based agents demonstrate a broad scope of general knowledge, they can occasionally face challenges when tasked with situations that require expertise knowledge. Enhancing these agents by integrating them with planners of specific domains has shown to yield better performance \cite{DBLP:journals/corr/abs-2304-11477,DBLP:journals/corr/abs-2307-02485,DBLP:journals/corr/abs-2205-00445,DBLP:journals/corr/abs-2308-06391}.

    \item \textbf{Plan reflection.} Upon formulating a plan, it's imperative to reflect upon and evaluate its merits. LLM-based agents leverage internal feedback mechanisms, often drawing insights from pre-existing models, to hone and enhance their strategies and planning approaches \cite{DBLP:journals/corr/abs-2303-11366,DBLP:journals/corr/abs-2303-17651,DBLP:journals/corr/abs-2305-14323,DBLP:journals/corr/abs-2308-00436}. To better align with human values and preferences, agents actively engage with humans, allowing them to rectify some misunderstandings and assimilate this tailored feedback into their planning methodology \cite{DBLP:journals/corr/abs-2303-17760,DBLP:conf/chi/WuTC22,DBLP:journals/corr/abs-2305-16291}. Furthermore, they could draw feedback from tangible or virtual surroundings, such as cues from task accomplishments or post-action observations, aiding them in revising and refining their plans \cite{DBLP:conf/iclr/YaoZYDSN023,DBLP:journals/corr/abs-2212-04088,DBLP:conf/corl/HuangXXCLFZTMCS22,DBLP:journals/corr/abs-2303-08268,DBLP:journals/corr/abs-2307-06135}.
\end{itemize}

\subsubsection{Transferability and Generalization} \label{sec:transferability and generalization}
Intelligence shouldn't be limited to a specific domain or task, but rather encompass a broad range of cognitive skills and abilities \cite{DBLP:journals/corr/abs-2303-12712}. The remarkable nature of the human brain is largely attributed to its high degree of plasticity and adaptability. It can continuously adjust its structure and function in response to external stimuli and internal needs, thereby adapting to different environments and tasks. These years, plenty of research indicates that pre-trained models on large-scale corpora can learn universal language representations \cite{barandiaran2009defining, peters-etal-2018-deep, DBLP:conf/naacl/DevlinCLT19}. Leveraging the power of pre-trained models, with only a small amount of data for fine-tuning, LLMs can demonstrate excellent performance in downstream tasks \cite{solaiman2021process}. There is no need to train new models from scratch, which saves a lot of computation resources. However, through this task-specific fine-tuning, the models lack versatility and struggle to be generalized to other tasks. Instead of merely functioning as a static knowledge repository, LLM-based agents exhibit dynamic learning ability which enables them to adapt to novel tasks swiftly and robustly \cite{DBLP:conf/nips/Ouyang0JAWMZASR22, DBLP:conf/iclr/WeiBZGYLDDL22, DBLP:conf/iclr/SanhWRBSACSRDBX22}.

\paragraph{Unseen task generalization.} \label{unseen task generalization}
Studies show that instruction-tuned LLMs exhibit zero-shot generalization without the need for task-specific fine-tuning \cite{DBLP:conf/nips/Ouyang0JAWMZASR22, DBLP:journals/corr/abs-2303-08774, DBLP:conf/iclr/WeiBZGYLDDL22, DBLP:conf/iclr/SanhWRBSACSRDBX22, DBLP:journals/corr/abs-2210-11416}. With the expansion of model size and corpus size, LLMs gradually exhibit remarkable emergent abilities in unfamiliar tasks \cite{DBLP:journals/corr/abs-2302-04023}. Specifically, LLMs can complete new tasks they do not encounter in the training stage by following the instructions based on their own understanding. One of the implementations is multi-task learning, for example, FLAN \cite{DBLP:conf/iclr/WeiBZGYLDDL22} finetunes language models on a collection of tasks described via instructions, and T0 \cite{DBLP:conf/iclr/SanhWRBSACSRDBX22} introduces a unified framework that converts every language problem into a text-to-text format. Despite being purely a language model, GPT-4 \cite{DBLP:journals/corr/abs-2303-08774} demonstrates remarkable capabilities in a variety of domains and tasks, including abstraction, comprehension, vision, coding, mathematics, medicine, law, understanding of human motives and emotions, and others \cite{DBLP:journals/corr/abs-2303-12712}. It is noticed that the choices in prompting are critical for appropriate predictions, and training directly on the prompts can improve the models' robustness in generalizing to unseen tasks \cite{DBLP:conf/acl/BachSYWRNSKBFAD22}. Promisingly, such generalization capability can further be enhanced by scaling up both the model size and the quantity or diversity of training instructions \cite{DBLP:journals/corr/abs-2304-08354, DBLP:journals/corr/abs-2212-12017}.

\paragraph{In-context learning.}
Numerous studies indicate that LLMs can perform a variety of complex tasks through in-context learning (ICL), which refers to the models' ability to learn from a few examples in the context \cite{DBLP:journals/corr/abs-2301-00234}. Few-shot in-context learning enhances the predictive performance of language models by concatenating the original input with several complete examples as prompts to enrich the context \cite{DBLP:conf/nips/BrownMRSKDNSSAA20}. The key idea of ICL is learning from analogy, which is similar to the learning process of humans \cite{DBLP:journals/cacm/Winston80}. Furthermore, since the prompts are written in natural language, the interaction is interpretable and changeable, making it easier to incorporate human knowledge into LLMs \cite{DBLP:conf/nips/Wei0SBIXCLZ22, DBLP:conf/acl/LuBM0S22}. Unlike the supervised learning process, ICL doesn't involve fine-tuning or parameter updates, which could greatly reduce the computation costs for adapting the models to new tasks. Beyond text, researchers also explore the potential ICL capabilities in different multimodal tasks \cite{DBLP:conf/cvpr/WangWCS023,DBLP:journals/corr/abs-2301-02111, DBLP:conf/nips/TsimpoukelliMCE21, DBLP:conf/nips/BarGDGE22, DBLP:journals/corr/abs-2304-04675, DBLP:journals/corr/abs-2303-03926}, making it possible for agents to be applied to large-scale real-world tasks.

\paragraph{Continual learning.}
Recent studies \cite{DBLP:journals/corr/abs-2305-16291, zhang2023bootstrap} have highlighted the potential of LLMs' planning capabilities in facilitating continuous learning \cite{Ke2022ContinualLO, Wang2023ACS} for agents, which involves continuous acquisition and update of skills. 
A core challenge in continual learning is catastrophic forgetting \cite{McCloskey1989CatastrophicII}: as a model learns new tasks, it tends to lose knowledge from previous tasks. 
Numerous efforts have been devoted to addressing the above challenge, which can be broadly separated into three groups, introducing regularly used terms in reference to the previous model \cite{kirkpatrick2017overcoming, li2017learning, farajtabar2020orthogonal, smith2023continual}, approximating prior data distributions \cite{lopez2017gradient, de2019episodic, rolnick2019experience}, and designing architectures with task-adaptive parameters \cite{Serr2018OvercomingCF, razdaibiedina2023progressive}. 
LLM-based agents have emerged as a novel paradigm, leveraging the planning capabilities of LLMs to combine existing skills and address more intricate challenges. Voyager \cite{DBLP:journals/corr/abs-2305-16291} attempts to solve progressively harder tasks proposed by the automatic curriculum devised by GPT-4 \cite{DBLP:journals/corr/abs-2303-08774}. By synthesizing complex skills from simpler programs, the agent not only rapidly enhances its capabilities but also effectively counters catastrophic forgetting.

\subsection{Perception}\label{sec:Perception}
\input{figures/sec3_perception_mindmap}

Both humans and animals rely on sensory organs like eyes and ears to gather information from their surroundings. These perceptual inputs are converted into neural signals and sent to the brain for processing \cite{hubel1962receptive,logothetis1996visual}, allowing us to perceive and interact with the world. Similarly, it's crucial for LLM-based agents to receive information from various sources and modalities. This expanded perceptual space helps agents better understand their environment, make informed decisions, and excel in a broader range of tasks, making it an essential development direction. Agent handles this information to the Brain module for processing through the perception module.

In this section, we introduce how to enable LLM-based agents to acquire multimodal perception capabilities, encompassing textual (\S \ \ref{sec:Textual Input}), visual (\S \ \ref{sec:Visual Input}), and auditory inputs (\S \ \ref{sec:Auditory Input}). We also consider other potential input forms (\S \ \ref{sec:Others Input}) such as tactile feedback, gestures, and 3D maps to enrich the agent's perception domain and enhance its versatility.3). The typology diagram for the
LLM-based agent perception is depicted in Figure \ref{fig:sec3_mindmap_perception}.

\subsubsection{Textual Input} \label{sec:Textual Input}
Text is a way to carry data, information, and knowledge, making text communication one of the most important ways humans interact with the world. An LLM-based agent already has the fundamental ability to communicate with humans through textual input and output \cite{gravitasauto}. In a user's textual input, aside from the explicit content, there are also beliefs, desires, and intentions hidden behind it. Understanding implied meanings is crucial for the agent to grasp the potential and underlying intentions of human users, thereby enhancing its communication efficiency and quality with users. However, as discussed in \S \  \ref{intention and implication understanding}, understanding implied meanings within textual input remains challenging for the current LLM-based agent. For example, some works
\cite{DBLP:conf/acl/LinFKD22,DBLP:conf/nips/ChristianoLBMLA17,DBLP:conf/hri/BasuSD18,DBLP:conf/aaai/SumersHHNG21} employ reinforcement learning to perceive implied meanings and models feedback to derive rewards. This helps deduce the speaker's preferences, leading to more personalized and accurate responses from the agent. Additionally, as the agent is designed for use in complex real-world situations, it will inevitably encounter many entirely new tasks. Understanding text instructions for unknown tasks places higher demands on the agent's text perception abilities. As described in \S \ \ref{unseen task generalization}, an LLM that has undergone instruction tuning \cite{DBLP:conf/iclr/WeiBZGYLDDL22} can exhibit remarkable zero-shot instruction understanding and generalization abilities, eliminating the need for task-specific fine-tuning.

\subsubsection{Visual Input}\label{sec:Visual Input}
Although LLMs exhibit outstanding performance in language comprehension \cite{DBLP:journals/corr/abs-2303-08774,chatgpt2022} and multi-turn conversations \cite{DBLP:conf/sigir/LuR0LX20}, they inherently lack visual perception and can only understand discrete textual content. Visual input usually contains a wealth of information about the world, including properties of objects, spatial relationships, scene layouts, and more in the agent's surroundings. Therefore, integrating visual information with data from other modalities can offer the agent a broader context and a more precise understanding \cite{DBLP:conf/icml/DriessXSLCIWTVY23}, deepening the agent's perception of the environment.

To help the agent understand the information contained within images, a straightforward approach is to generate corresponding text descriptions for image inputs, known as image captioning \cite{DBLP:conf/iccv/HuangWCW19,DBLP:conf/cvpr/PanYLM20,DBLP:journals/corr/abs-1912-08226,DBLP:journals/corr/abs-2102-10407,DBLP:journals/corr/abs-2305-06355}. Captions can be directly linked with standard text instructions and fed into the agent. This approach is highly interpretable and doesn't require additional training for caption generation, which can save a significant number of computational resources. However, caption generation is a low-bandwidth method \cite{DBLP:conf/icml/DriessXSLCIWTVY23,DBLP:journals/corr/abs-2308-01399}, and it may lose a lot of potential information during the conversion process. Furthermore,  the agent's focus on images may introduce biases.

Inspired by the excellent performance of transformers \cite{DBLP:conf/nips/VaswaniSPUJGKP17} in natural language processing, researchers have extended their use to the field of computer vision. Representative works like ViT/VQVAE \cite{DBLP:conf/iclr/DosovitskiyB0WZ21,DBLP:conf/nips/OordVK17,DBLP:conf/iclr/MehtaR22,DBLP:conf/nips/TolstikhinHKBZU21,DBLP:conf/icml/TouvronCDMSJ21} have successfully encoded visual information using transformers. Researchers first divide an image into fixed-size patches and then treat these patches, after linear projection, as input tokens for Transformers \cite{DBLP:journals/corr/abs-2304-08485}. In the end, by calculating self-attention between tokens, they are able to integrate information across the entire image, resulting in a highly effective way to perceive visual content. Therefore, some works \cite{DBLP:conf/iclr/LuCZMK23} try to combine the image encoder and LLM directly to train the entire model in an end-to-end way. While the agent can achieve remarkable visual perception abilities, it comes at the cost of substantial computational resources.

Extensively pre-trained visual encoders and LLMs can greatly enhance the agent's visual perception and language expression abilities \cite{DBLP:journals/corr/abs-2302-14045,DBLP:journals/corr/abs-2306-14824}. Freezing one or both of them during training is a widely adopted paradigm that achieves a balance between training resources and model performance \cite{DBLP:conf/icml/0008LSH23}. 
However, LLMs cannot directly understand the output of a visual encoder, so it's necessary to convert the image encoding into embeddings that LLMs can comprehend. In other words, it involves aligning the visual encoder with the LLM. 
This usually requires adding an extra learnable interface layer between them. For example, BLIP-2 \cite{DBLP:conf/icml/0008LSH23} and InstructBLIP \cite{DBLP:journals/corr/abs-2305-06500} use the Querying Transformer(Q-Former) module as an intermediate layer between the visual encoder and the LLM \cite{DBLP:journals/corr/abs-2305-06500}. Q-Former is a transformer that employs learnable query vectors \cite{DBLP:journals/corr/abs-2305-04790}, giving it the capability to extract language-informative visual representations. 
It can provide the most valuable information to the LLM, reducing the agent's burden of learning visual-language alignment and thereby mitigating the issue of catastrophic forgetting. At the same time, some researchers adopt a computationally efficient method by using a single projection layer to achieve visual-text alignment, reducing the need for training additional parameters \cite{zhu2023minigpt, DBLP:journals/corr/abs-2305-16355,DBLP:journals/corr/abs-2306-14824}. Moreover, the projection layer can effectively integrate with the learnable interface to adapt the dimensions of its outputs, making them compatible with LLMs \cite{DBLP:journals/corr/abs-2305-04160,DBLP:journals/corr/abs-2306-02858,DBLP:journals/corr/abs-2306-09093,DBLP:journals/corr/abs-2306-05424}.

Video input consists of a series of continuous image frames. As a result, the methods used by agents to perceive images \cite{DBLP:conf/icml/0008LSH23} may be applicable to the realm of videos, allowing the agent to have good perception of video inputs as well. Compared to image information, video information adds a temporal dimension. Therefore, the agent's understanding of the relationships between different frames in time is crucial for perceiving video information. Some works like Flamingo \cite{DBLP:conf/nips/AlayracDLMBHLMM22,DBLP:journals/corr/abs-2304-03373} ensure temporal order when understanding videos using a mask mechanism. The mask mechanism restricts the agent's view to only access visual information from frames that occurred earlier in time when it perceives a specific frame in the video.

\subsubsection{Auditory Input}\label{sec:Auditory Input}
Undoubtedly, auditory information is a crucial component of world information. When an agent possesses auditory capabilities, it can improve its awareness of interactive content, the surrounding environment, and even potential dangers. Indeed, there are numerous well-established models and approaches  \cite{DBLP:journals/corr/abs-2304-12995,DBLP:conf/icml/RadfordKXBMS23,DBLP:conf/nips/RenRTQZZL19} for processing audio as a standalone modality. However, these models often excel at specific tasks. Given the excellent tool-using capabilities of LLMs (which will be discussed in detail in \S \ref{sec:Action}), a very intuitive idea is that the agent can use LLMs as control hubs, invoking existing toolsets or model repositories in a cascading manner to perceive audio information. For instance, AudioGPT \cite{DBLP:journals/corr/abs-2304-12995}, makes full use of the capabilities of models like FastSpeech \cite{DBLP:conf/nips/RenRTQZZL19}, GenerSpeech  \cite{DBLP:conf/icml/RadfordKXBMS23}, Whisper \cite{DBLP:conf/icml/RadfordKXBMS23}, and others \cite{DBLP:conf/ijcai/YeZ0022,DBLP:conf/icml/KimKS21,DBLP:journals/taslp/WangCCLKW23,DBLP:conf/aaai/Liu00CZ22,DBLP:conf/asru/InagumaDYW21} which have achieved excellent results in tasks such as Text-to-Speech, Style Transfer, and Speech Recognition.

An audio spectrogram provides an intuitive representation of the frequency spectrum of an audio signal as it changes over time \cite{flanagan2013speech}. For a segment of audio data over a period of time, it can be abstracted into a finite-length audio spectrogram. An audio spectrogram has a 2D representation, which can be visualized as a flat image. Hence, some research \cite{DBLP:conf/interspeech/GongCG21,DBLP:journals/taslp/HsuBTLSM21} efforts aim to migrate perceptual methods from the visual domain to audio. AST (Audio Spectrogram Transformer) \cite{DBLP:conf/interspeech/GongCG21} employs a Transformer architecture similar to ViT to process audio spectrogram images. By segmenting the audio spectrogram into patches, it achieves effective encoding of audio information. Moreover, some researchers \cite{DBLP:journals/corr/abs-2305-04160,DBLP:journals/corr/abs-2306-02858} have drawn inspiration from the idea of freezing encoders to reduce training time and computational costs. They align audio encoding with data encoding from other modalities by adding the same learnable interface layer.

\subsubsection{Other Input}\label{sec:Others Input}
As mentioned earlier, many studies have looked into perception units for text, visual, and audio. However, LLM-based agents might be equipped with richer perception modules. In the future, they could perceive and understand diverse modalities in the real world, much like humans. For example, agents could have unique touch and smell organs, allowing them to gather more detailed information when interacting with objects. At the same time, agents can also have a clear sense of the temperature, humidity, and brightness in their surroundings, enabling them to take environment-aware actions. Moreover, by efficiently integrating basic perceptual abilities like vision, text, and light sensitivity, agents can develop various user-friendly perception modules for humans. InternGPT \cite{DBLP:journals/corr/abs-2305-05662} introduces pointing instructions. Users can interact with specific, hard-to-describe portions of an image by using gestures or moving the cursor to select, drag, or draw. The addition of pointing instructions helps provide more precise specifications for individual text instructions. Building upon this, agents have the potential to perceive more complex user inputs. For example, technologies such as eye-tracking in AR/VR devices, body motion capture, and even brainwave signals in brain-computer interaction.

Finally, a human-like LLM-based agent should possess awareness of a broader overall environment.  At present, numerous mature and widely adopted hardware devices can assist agents in accomplishing this. Lidar \cite{schwarz2010mapping} can create 3D point cloud maps to help agents detect and identify objects in their surroundings. GPS \cite{parkinson1996progress} can provide accurate location coordinates and can be integrated with map data. Inertial Measurement Units (IMUs) can measure and record the three-dimensional motion of objects, offering details about an object's speed and direction. However, these sensory data are complex and cannot be directly understood by LLM-based agents. Exploring how agents can perceive more comprehensive input is a promising direction for the future.

\subsection{Action}\label{sec:Action}
\input{figures/sec3_action_mindmap}
After humans perceive their environment, their brains integrate, analyze, and reason with the perceived information and make decisions. 
Subsequently, they employ their nervous systems to control their bodies, enabling adaptive or creative actions in response to the environment, such as engaging in conversation, evading obstacles, or starting a fire. 
When an agent possesses a brain-like structure with capabilities of knowledge, memory, reasoning, planning, and generalization, as well as multimodal perception, it is also expected to possess a diverse range of actions akin to humans to respond to its surrounding environment. In the construction of the agent, the action module receives action sequences sent by the brain module and carries out actions to interact with the environment.
As Figure \ref{fig:sec3_mindmap_action} shows, this section begins with textual output (\S \ \ref{sec:Textual Output}), which is the inherent capability of LLM-based agents. 
Next we talk about the tool-using capability of LLM-based agents (\S \ \ref{sec:Tool Using}), which has proved effective in enhancing their versatility and expertise. 
Finally, we discuss equipping the LLM-based agent with embodied action to facilitate its grounding in the physical world (\S \ \ref{sec:Embodied Action}).


\subsubsection{Textual Output} \label{sec:Textual Output}

As discussed in \S \ \ref{sec:Natural Language Interaction}, the rise and development of Transformer-based generative large language models have endowed LLM-based agents with inherent language generation capabilities \cite{DBLP:journals/corr/abs-2302-04023, DBLP:journals/corr/abs-2303-12528}. The text quality they generate excels in various aspects such as fluency, relevance, diversity, controllability \cite{DBLP:journals/corr/abs-2305-13711, DBLP:conf/conll/SeePSYM19, DBLP:conf/eacl/LuZZWY23, DBLP:journals/corr/abs-2111-09509}. Consequently, LLM-based agents can be exceptionally strong language generators.

\subsubsection{Tool Using}\label{sec:Tool Using}
Tools are extensions of the capabilities of tool users. When faced with complex tasks, humans employ tools to simplify task-solving and enhance efficiency, freeing time and resources. Similarly, agents have the potential to accomplish complex tasks more efficiently and with higher quality if they also learn to use and utilize tools \cite{DBLP:journals/corr/abs-2304-08354}.

LLM-based agents have \textit{limitations} in some aspects, and the use of tools can \textit{strengthen the agents' capabilities}. 
First, although LLM-based agents have a strong knowledge base and expertise, they don't have the ability to memorize every piece of training data \cite{DBLP:journals/corr/abs-2301-13188,DBLP:conf/icail/SavelkaAGWX23}. They may also fail to steer to correct knowledge due to the influence of contextual prompts \cite{DBLP:journals/corr/abs-2302-07842}, or even generate hallucinate knowledge \cite{DBLP:conf/eacl/RollerDGJWLXOSB21}. Coupled with the lack of corpus, training data, and tuning for specific fields and scenarios, agents' expertise is also limited when specializing in specific domains \cite{ling2023domain}. Specialized tools enable LLMs to enhance their expertise, adapt domain knowledge, and be more suitable for domain-specific needs in a pluggable form. 
Furthermore, the decision-making process of LLM-based agents lacks transparency, making them less trustworthy in high-risk domains such as healthcare and finance \cite{DBLP:journals/entropy/LinardatosPK21}. Additionally, LLMs are susceptible to adversarial attacks \cite{DBLP:journals/corr/abs-2307-15043}, and their robustness against slight input modifications is inadequate. In contrast, agents that accomplish tasks with the assistance of tools exhibit stronger interpretability and robustness. The execution process of tools can reflect the agents' approach to addressing complex requirements and enhance the credibility of their decisions. Moreover, for the reason that tools are specifically designed for their respective usage scenarios, agents utilizing such tools are better equipped to handle slight input modifications and are more resilient against adversarial attacks \cite{DBLP:journals/corr/abs-2304-08354}.

LLM-based agents not only require the use of tools, but are also \textit{well-suited} for tool integration. Leveraging the rich world knowledge accumulated through the pre-training process and CoT prompting, LLMs have demonstrated remarkable reasoning and decision-making abilities in complex interactive environments \cite{DBLP:conf/iclr/0002WSLCNCZ23}, which help agents break down and address tasks specified by users in an appropriate way. What's more, LLMs show significant potential in intent understanding and other aspects \cite{ DBLP:journals/corr/abs-2303-08774,DBLP:journals/corr/abs-2302-13971, DBLP:journals/corr/abs-2211-05100, almazrouei2023falcon}. When agents are combined with tools, the threshold for tool utilization can be lowered, thereby fully unleashing the creative potential of human users \cite{DBLP:journals/corr/abs-2304-08354}.

\paragraph{Understanding tools.}
A prerequisite for an agent to use tools effectively is a comprehensive understanding of the tools' application scenarios and invocation methods. Without this understanding, the process of the agent using tools will become untrustworthy and fail to genuinely enhance the agent's capabilities. Leveraging the powerful zero-shot and few-shot learning abilities of LLMs \cite{radford2019language, DBLP:conf/nips/BrownMRSKDNSSAA20}, agents can acquire knowledge about tools by utilizing \textit{zero-shot prompts} that describe tool functionalities and parameters, or \textit{few-shot prompts} that provide demonstrations of specific tool usage scenarios and corresponding methods \cite{DBLP:journals/corr/abs-2302-04761, DBLP:journals/corr/abs-2205-12255}. These learning approaches parallel human methods of learning by consulting tool manuals or observing others using tools \cite{DBLP:journals/corr/abs-2304-08354}. A single tool is often insufficient when facing complex tasks. Therefore, the agents should first decompose the complex task into subtasks in an appropriate manner, and their understanding of tools play a significant role in task decomposition.

\paragraph{Learning to use tools.}
The methods for agents to learn to utilize tools primarily consist of \textit{learning from demonstrations} and \textit{learning from feedback}. This involves mimicking the behavior of human experts \cite{DBLP:journals/csur/HusseinGEJ17, DBLP:conf/icra/LiuGAL18, DBLP:conf/nips/BakerAZHTEHSC22}, as well as understanding the consequences of their actions and making adjustments based on feedback received from both the environment and humans \cite{ DBLP:conf/nips/Ouyang0JAWMZASR22,DBLP:journals/ijrr/LevinePKIQ18, DBLP:journals/corr/abs-2307-04964}. Environmental feedback encompasses result feedback on whether actions have successfully completed the task and intermediate feedback that captures changes in the environmental state caused by actions; human feedback comprises explicit evaluations and implicit behaviors, such as clicking on links \cite{DBLP:journals/corr/abs-2304-08354}.

If an agent rigidly applies tools without \textit{adaptability}, it cannot achieve acceptable performance in all scenarios. Agents need to generalize their tool usage skills learned in specific contexts to more general situations, such as transferring a model trained on Yahoo search to Google search. To accomplish this, it's necessary for agents to grasp the common principles or patterns in tool usage strategies, which can potentially be achieved through meta-tool learning \cite{Clarebout2013}. Enhancing the agent's understanding of relationships between simple and complex tools, such as how complex tools are built on simpler ones, can contribute to the agents' capacity to generalize tool usage. This allows agents to effectively discern nuances across various application scenarios and transfer previously learned knowledge to new tools \cite{DBLP:journals/corr/abs-2304-08354}. Curriculum learning \cite{10.1145/1553374.1553380}, which allows an agent to start from simple tools and progressively learn complex ones, aligns with the requirements. Moreover, benefiting from the understanding of user intent reasoning and planning abilities, agents can better design methods of tool utilization and collaboration and then provide higher-quality outcomes.

\paragraph{Making tools for self-sufficiency.}
Existing tools are often designed for human convenience, which might not be optimal for agents. To make agents use tools better, there's a need for tools specifically designed for agents. These tools should be more modular and have input-output formats that are more suitable for agents. 
If instructions and demonstrations are provided, LLM-based agents also possess the ability to create tools by generating executable programs, or integrating existing tools into more powerful ones \cite{DBLP:journals/corr/abs-2304-08354,DBLP:journals/corr/abs-2305-14318, chen2021evaluating}. and they can learn to perform self-debugging \cite{DBLP:journals/corr/abs-2304-05128}.
Moreover, if the agent that serves as a tool maker successfully creates a tool, it can produce packages containing the tool's code and demonstrations for other agents in a multi-agent system, in addition to using the tool itself \cite{DBLP:journals/corr/abs-2305-17126}.
Speculatively, in the future, agents might become self-sufficient and exhibit a high degree of autonomy in terms of tools. 

\paragraph{Tools can expand the action space of LLM-based agents.}
With the help of tools, agents can utilize various external \textit{resources} such as web applications and other LMs during the reasoning and planning phase \cite{DBLP:journals/corr/abs-2302-04761}. 
This process can provide information with high expertise, reliability, diversity, and quality for LLM-based agents, facilitating their decision-making and action. 
For example, search-based tools can improve the scope and quality of the knowledge accessible to the agents with the aid of external databases, knowledge graphs, and web pages, while domain-specific tools can enhance an agent's expertise in the corresponding field \cite{DBLP:journals/corr/abs-2304-04370,DBLP:journals/corr/abs-2306-08302}. Some researchers have already developed LLM-based controllers that generate SQL statements to query databases, or to convert user queries into search requests and use search engines to obtain the desired results \cite{DBLP:journals/corr/abs-2112-09332,DBLP:journals/corr/abs-2306-03901}. 
What's more, LLM-based agents can use scientific tools to execute tasks like organic synthesis in chemistry, or interface with Python interpreters to enhance their performance on intricate mathematical computation tasks \cite{bran2023chemcrow, DBLP:journals/corr/abs-2308-03427}. For multi-agent systems, communication tools (e.g., emails) may serve as a means for agents to interact with each other under strict security constraints, facilitating their \textit{collaboration}, and showing autonomy and flexibility \cite{DBLP:journals/corr/abs-2304-08354}.

Although the tools mentioned before enhance the capabilities of agents, the medium of interaction with the environment remains text-based. 
However, tools are designed to expand the functionality of language models, and their outputs are not limited to text. 
Tools for non-textual output can diversify the \textit{modalities} of agent actions, thereby expanding the application scenarios of LLM-based agents. For example, image processing and generation can be accomplished by an agent that draws on a visual model \cite{DBLP:journals/corr/abs-2303-04671}. In aerospace engineering, agents are being explored for modeling physics and solving complex differential equations \cite{ogundare2023industrial}; in the field of robotics, agents are required to plan physical operations and control the robot execution \cite{DBLP:conf/corl/IchterBCFHHHIIJ22}; and so on. Agents that are capable of dynamically interacting with the environment or the world through tools, or in a multimodal manner, can be referred to as digitally embodied \cite{DBLP:journals/corr/abs-2304-08354}. The \textit{embodiment} of agents has been a central focus of embodied learning research. We will make a deep discussion on agents' embodied action in \S\ref{sec:Embodied Action}.

\subsubsection{Embodied Action}\label{sec:Embodied Action}
In the pursuit of Artificial General Intelligence (AGI), the embodied agent is considered a pivotal paradigm while it strives to integrate model intelligence with the physical world. \textit{The Embodiment hypothesis} \cite{smith2005development} draws inspiration from the human intelligence development process, posing that an agent's intelligence arises from continuous interaction and feedback with the environment rather than relying solely on well-curated textbooks. Similarly, unlike traditional deep learning models that learn explicit capabilities from the internet datasets to solve domain problems, people anticipate that LLM-based agents' behaviors will no longer be limited to pure text output or calling exact tools to perform particular domain tasks \cite{DBLP:journals/tetci/DuanYTZT22}. 
Instead, they should be capable of actively perceiving, comprehending, and interacting with physical environments, making decisions, and generating specific behaviors to modify the environment based on LLM's extensive internal knowledge. We collectively term these as \textbf{embodied actions}, which enable agents' ability to interact with and comprehend the world in a manner closely resembling human behavior.

\paragraph{The potential of LLM-based agents for embodied actions.}
Before the widespread rise of LLMs, researchers tended to use methods like reinforcement learning to explore the embodied actions of agents. Despite the extensive success of RL-based embodiment \cite{DBLP:journals/corr/MnihKSGAWR13, DBLP:journals/nature/SilverHMGSDSAPL16, DBLP:journals/corr/abs-1806-10293}, it does have certain limitations in some aspects. In brief, RL algorithms face limitations in terms of data efficiency, generalization, and complex problem reasoning due to challenges in modeling the dynamic and often ambiguous real environment, or their heavy reliance on precise reward signal representations \cite{DBLP:conf/irc/NguyenL19}. 
Recent studies have indicated that leveraging the rich internal knowledge acquired during the pre-training of LLMs can effectively alleviate these issues \cite{DBLP:conf/icml/DriessXSLCIWTVY23, DBLP:conf/corl/HuangXXCLFZTMCS22, DBLP:conf/icml/HuangAPM22, DBLP:journals/corr/abs-2302-00763}. 

\begin{itemize}[leftmargin=*]
    \item \textbf{Cost efficiency.} 
    Some on-policy algorithms struggle with sample efficiency as they require fresh data for policy updates while gathering enough embodied data for high-performance training is costly and noisy. The constraint is also found in some end-to-end models \cite{DBLP:conf/cvpr/PuigRBLWF018, DBLP:journals/corr/abs-2011-13922, DBLP:journals/corr/abs-2108-04927}. By leveraging the intrinsic knowledge from LLMs, agents like PaLM-E \cite{DBLP:conf/icml/DriessXSLCIWTVY23} jointly train robotic data with general visual-language data to achieve significant transfer ability in embodied tasks while also showcasing that geometric input representations can improve training data efficiency.
    
    \item \textbf{Embodied action generalization.}
    As discussed in section \S \ref{sec:transferability and generalization}, an agent's competence should extend beyond specific tasks. When faced with intricate, uncharted real-world environments, it's imperative that the agent exhibits dynamic learning and generalization capabilities. However, the majority of RL algorithms are designed to train and evaluate relevant skills for specific tasks \cite{DBLP:journals/corr/abs-2212-04088, DBLP:journals/corr/abs-1911-05892, DBLP:journals/arc/TipaldiIM22, DBLP:conf/iccv/SavvaMPBKMZWJSL19}. In contrast, fine-tuned by diverse forms and rich task types, LLMs have showcased remarkable cross-task generalization capabilities \cite{longpre2023flan, wang2022self}. For instance, PaLM-E exhibits surprising zero-shot or one-shot generalization capabilities to new objects or novel combinations of existing objects \cite{DBLP:conf/icml/DriessXSLCIWTVY23}. Further, language proficiency represents a distinctive advantage of LLM-based agents, serving both as a means to interact with the environment and as a medium for transferring foundational skills to new tasks \cite{DBLP:conf/icra/LiangHXXHIFZ23}. SayCan \cite{DBLP:conf/corl/IchterBCFHHHIIJ22} decomposes task instructions presented in prompts using LLMs into corresponding skill commands, but in partially observable environments, limited prior skills often do not achieve satisfactory performance \cite{DBLP:journals/corr/abs-2212-04088}. To address this, Voyager \cite{DBLP:journals/corr/abs-2305-16291} introduces the skill library component to continuously collect novel self-verified skills, which allows for the agent's lifelong learning capabilities.
    
    \item \textbf{Embodied action planning.}
    Planning constitutes a pivotal strategy employed by humans in response to complex problems as well as LLM-based agents. Before LLMs exhibited remarkable reasoning abilities, researchers introduced Hierarchical Reinforcement Learning (HRL) methods while the high-level policy constraints sub-goals for the low-level policy and the low-level policy produces appropriate action signals \cite{DBLP:conf/corl/LiXMS19,DBLP:journals/corr/abs-2012-10147,DBLP:conf/nips/00070C22}. Similar to the role of high-level policies, LLMs with emerging reasoning abilities \cite{DBLP:journals/tmlr/WeiTBRZBYBZMCHVLDF22} can be seamlessly applied to complex tasks in a zero-shot or few-shot manner \cite{DBLP:conf/nips/Wei0SBIXCLZ22, DBLP:conf/iclr/0002WSLCNCZ23, DBLP:conf/iclr/ZhouSHWS0SCBLC23, DBLP:journals/corr/abs-2305-14497}. In addition, external feedback from the environment can further enhance LLM-based agents' planning performance. Based on the current environmental feedback, some work \cite{DBLP:journals/corr/abs-2212-04088, DBLP:conf/iclr/YaoZYDSN023, shinn2023reflexion, DBLP:journals/corr/abs-2306-03604} dynamically generate, maintain, and adjust high-level action plans in order to minimize dependency on prior knowledge in partially observable environments, thereby grounding the plan. Feedback can also come from models or humans, which can usually be referred to as the critics, assessing task completion based on the current state and task prompts \cite{DBLP:journals/corr/abs-2303-08774, DBLP:journals/corr/abs-2305-16291}.
\end{itemize}

\paragraph{Embodied actions for LLM-based agents.} 
Depending on the agents' level of autonomy in a task or the complexity of actions, there are several fundamental LLM-based embodied actions, primarily including observation, manipulation, and navigation. 

\begin{itemize}[leftmargin=*]
    \item \textbf{Observation.}
    Observation constitutes the primary ways by which the agent acquires environmental information and updates states, playing a crucial role in enhancing the efficiency of subsequent embodied actions. As mentioned in \S \ref{sec:Perception}, observation by embodied agents primarily occurs in environments with various inputs, which are ultimately converged into a multimodal signal. A common approach entails a pre-trained Vision Transformer (ViT) used as the alignment module for text and visual information and special tokens are marked to denote the positions of multimodal data \cite{DBLP:conf/icml/DriessXSLCIWTVY23, liu2022instruction, DBLP:journals/corr/abs-2305-15021}. 
    Soundspaces \cite{DBLP:conf/eccv/ChenJSGAIRG20} proposes the identification of physical spatial geometric elements guided by reverberant audio input, enhancing the agent's observations with a more comprehensive perspective \cite{DBLP:conf/nips/00070C22}. In recent times, even more research takes audio as a modality for embedded observation. Apart from the widely employed cascading paradigm \cite{DBLP:journals/corr/abs-2304-12995, DBLP:conf/nips/Huang0LCZ22, DBLP:conf/icml/RadfordKXBMS23}, audio information encoding similar to ViT further enhances the seamless integration of audio with other modalities of inputs \cite{DBLP:conf/interspeech/GongCG21}. 
    The agent's observation of the environment can also be derived from real-time linguistic instructions from humans, while human feedback helps the agent in acquiring detail information that may not be readily obtained or parsed \cite{DBLP:journals/corr/abs-2210-06407, DBLP:journals/corr/abs-2305-16291}.

    \item \textbf{Manipulation.}
    In general, manipulation tasks for embodied agents include object rearrangements, tabletop manipulation, and mobile manipulation \cite{DBLP:journals/corr/abs-2305-13246, DBLP:conf/icml/DriessXSLCIWTVY23}. The typical case entails the agent executing a sequence of tasks in the kitchen, which includes retrieving items from drawers and handing them to the user, as well as cleaning the tabletop \cite{DBLP:conf/corl/IchterBCFHHHIIJ22}. Besides precise observation, this involves combining a series of subgoals by leveraging LLM. Consequently, maintaining synchronization between the agent's state and the subgoals is of significance. DEPS \cite{DBLP:journals/corr/abs-2302-01560} utilizes an LLM-based interactive planning approach to maintain this consistency and help error correction from agent's feedback throughout the multi-step, long-haul reasoning process.
    In contrast to these, AlphaBlock \cite{DBLP:journals/corr/abs-2305-18898} focuses on more challenging manipulation tasks (e.g. making a smiley face using building blocks), which requires the agent to have a more grounded understanding of the instructions. Unlike the existing open-loop paradigm, AlphaBlock constructs a dataset comprising 35 complex high-level tasks, along with corresponding multi-step planning and observation pairs, and then fine-tunes a multimodal model to enhance its comprehension of high-level cognitive instructions.

    \item \textbf{Navigation.} 
    Navigation permits agents to dynamically alter their positions within the environment, which often involves multi-angle and multi-object observations, as well as long-horizon manipulations based on current exploration \cite{DBLP:journals/corr/abs-2305-13246}. Before navigation, it is essential for embodied agents to establish prior internal maps about the external environment, which are typically in the form of a topological map, semantic map or occupancy map \cite{DBLP:journals/tetci/DuanYTZT22}. For example, LM-Nav \cite{DBLP:conf/corl/ShahOIL22} utilizes the VNM \cite{DBLP:conf/icra/ShahEKRL21} to create an internal topological map. It further leverages the LLM and VLM for decomposing input commands and analyzing the environment to find the optimal path. Furthermore, some \cite{DBLP:conf/icra/HuangMZB23, DBLP:conf/cvpr/GeorgakisSWDMRD22} highlight the importance of spatial representation to achieve the precise localization of spatial targets rather than conventional point or object-centric navigation actions by leveraging the pre-trained VLM model to combine visual features from images with 3D reconstructions of the physical world \cite{DBLP:journals/tetci/DuanYTZT22}. Navigation is usually a long-horizon task, where the upcoming states of the agent are influenced by its past actions. A memory buffer and summary mechanism are needed to serve as a reference for historical information \cite{DBLP:journals/corr/abs-2305-16986}, which is also employed in Smallville and Voyager \cite{DBLP:journals/corr/abs-2304-03442, DBLP:journals/corr/abs-2305-16291, DBLP:journals/corr/abs-2303-03480, DBLP:conf/cvpr/LiZZYLZWYZHCG22}.
    Additionally, as mentioned in \S \ref{sec:Perception}, some works have proposed the audio input is also of great significance, but integrating audio information presents challenges in associating it with the visual environment. A basic framework includes a dynamic path planner that uses visual and auditory observations along with spatial memories to plan a series of actions for navigation \cite{DBLP:conf/nips/00070C22, DBLP:conf/icra/GanZ0GT20}. 
    
\end{itemize}

By integrating these, the agent can accomplish more complex tasks, such as embodied question answering, whose primary objective is autonomous exploration of the environment, and responding to pre-defined multimodal questions, such as \textit{Is the watermelon in the kitchen larger than the pot? Which one is harder?} To address these questions, the agent needs to navigate to the kitchen, observe the sizes of both objects and then answer the questions through comparison \cite{DBLP:journals/tetci/DuanYTZT22}. 

In terms of control strategies, as previously mentioned, LLM-based agents trained on particular embodied datasets typically generate high-level policy commands to control low-level policies for achieving specific sub-goals. The low-level policy can be a robotic transformer \cite{DBLP:conf/icml/DriessXSLCIWTVY23, DBLP:journals/corr/abs-2212-06817, DBLP:journals/corr/abs-2307-15818}, which takes images and instructions as inputs and produces control commands for the end effector as well as robotic arms in particular embodied tasks \cite{DBLP:conf/corl/IchterBCFHHHIIJ22}. Recently, in virtual embodied environments, the high-level strategies are utilized to control agents in gaming \cite{DBLP:journals/corr/abs-2305-17144,DBLP:journals/corr/abs-2302-01560, DBLP:journals/corr/abs-2305-16291, DBLP:conf/nips/FanWJMYZTHZA22} or simulated worlds \cite{DBLP:journals/corr/abs-2304-03442, DBLP:journals/corr/abs-2303-17760, DBLP:journals/corr/abs-2307-07924}. For instance, Voyager \cite{DBLP:journals/corr/abs-2305-16291} calls the Mineflayer \cite{Mineflayer-JavaScript-2013} API interface to continuously acquire various skills and explore the world.

\paragraph{Prospective future of the embodied action.} 
LLM-based embodied actions are seen as the bridge between virtual intelligence and the physical world, enabling agents to perceive and modify the environment much like humans. However, there remain several constraints such as high costs of physical-world robotic operators and the scarcity of embodied datasets, which foster a growing interest in investigating agents' embodied actions within simulated environments like Minecraft \cite{DBLP:journals/corr/abs-2302-01560, DBLP:journals/corr/abs-2106-14876, DBLP:conf/nips/FanWJMYZTHZA22, DBLP:journals/corr/abs-2305-16291, DBLP:conf/icml/NottinghamAS0H023}. By utilizing the Mineflayer \cite{Mineflayer-JavaScript-2013} API, these investigations enable cost-effective examination of a wide range of embodied agents' operations including exploration, planning, self-improvement, and even lifelong learning \cite{DBLP:journals/corr/abs-2305-16291}. Despite notable progress, achieving optimal embodied actions remains a challenge due to the significant disparity between simulated platforms and the physical world. To enable the effective deployment of embodied agents in real-world scenarios, an increasing demand exists for embodied task paradigms and evaluation criteria that closely mirror real-world conditions \cite{DBLP:journals/tetci/DuanYTZT22}. On the other hand, learning to ground language for agents is also an obstacle. For example, expressions like ``jump down like a cat'' primarily convey a sense of lightness and tranquility, but this linguistic metaphor requires adequate world knowledge \cite{DBLP:conf/emnlp/BiskHTABCLLMNPT20}. \cite{DBLP:conf/icml/SumersMAF023} endeavors to amalgamate text distillation with Hindsight Experience Replay (HER) to construct a dataset as the supervised signal for the training process. Nevertheless, additional investigation on grounding embodied datasets still remains necessary while embodied action plays an increasingly pivotal role across various domains in human life.

%% file: figures/sec3_brain_mindmap.tex
\begin{figure*}[!ht]
\scriptsize
    \begin{adjustbox}{width=\textwidth}
        \begin{forest}
        for tree={
                forked edges,
                grow'=0,
                draw,
                rounded corners,
                node options={align=center},
                text width=2.7cm,
                s sep=6pt,
                calign=edge midpoint, 
            },
                [Brain, fill=gray!45, parent
                    [Natural Language Interaction \S\ref{sec:Natural Language Interaction},  for tree={brain}
                        [High-quality generation, brain
                            [{Bang et al. \cite{DBLP:journals/corr/abs-2302-04023}, Fang et al. \cite{DBLP:journals/corr/abs-2304-01746}, Lin et al. \cite{DBLP:journals/corr/abs-2305-13711}, Lu et al. \cite{DBLP:conf/eacl/LuZZWY23}, etc.}, brain_work]
                        ]
                        [Deep understanding, brain
                            [{Buehler et al. \cite{DBLP:journals/corr/abs-2109-01355}, Lin et al. \cite{DBLP:conf/acl/LinFKD22}, Shapira et al. \cite{DBLP:journals/corr/abs-2305-14763}, etc.}, brain_work]
                        ]
                    ]
                    [Knowledge \S\ref{sec:knowledge},   brain
                        [Knowledge in LLM-based agent,  brain
                            [Pretrain model, brain
                                [{Hill et al. \cite{DBLP:conf/naacl/HillCK16}, Collobert et al. \cite{DBLP:journals/jmlr/CollobertWBKKK11}, Kaplan et al. \cite{DBLP:journals/corr/abs-2001-08361}, Roberts et al. \cite{DBLP:conf/emnlp/RobertsRS20}, Tandon et al.  \cite{DBLP:journals/sigmod/TandonVM17}, etc.}, brain_work] 
                            ]
                            [Linguistic knowledge, brain
                                [{Vulic et al. \cite{DBLP:conf/emnlp/VulicPLGK20}, Hewitt et al. \cite{DBLP:conf/naacl/HewittM19}, Rau et al. \cite{DBLP:journals/ipm/RauJZ89}, Yang et al. \cite{DBLP:conf/dasfaa/YangCCHL16}, Beloucif et al. \cite{DBLP:conf/emnlp/BeloucifB21}, Zhang et al. \cite{DBLP:journals/corr/abs-2103-03125}, Bang et al. \cite{DBLP:journals/corr/abs-2302-04023}, etc.}, brain_work] 
                            ]
                            [Commensense knowledge, brain
                                [{Safavi et al. \cite{DBLP:conf/emnlp/SafaviK21}, Jiang et al. \cite{DBLP:journals/tacl/JiangXAN20}, Madaan \cite{DBLP:conf/emnlp/MadaanZ0YN22}, etc.}, brain_work] 
                            ]
                            [Actionable knowledge, brain
                                [{Xu et al. \cite{DBLP:conf/pldi/Xu0NH22}, Cobbe et al. \cite{DBLP:journals/corr/abs-2110-14168}, Thirunavukarasu et al. \cite{thirunavukarasu2023large}, Lai et al. \cite{DBLP:conf/icml/Lai0WZZZYFWY23}, Madaan et al. \cite{DBLP:conf/emnlp/MadaanZ0YN22}, etc.}, brain_work] 
                            ]
                        ]
                        [Potential issues of knowledge,  brain
                            [Edit wrong and outdated knowledge,  brain
                                [{AlKhamissi et al. \cite{DBLP:journals/corr/abs-2204-06031}, Kemker et al. \cite{DBLP:conf/aaai/KemkerMAHK18}, Cao et al. \cite{DBLP:conf/emnlp/CaoAT21}, Yao et al. \cite{DBLP:journals/corr/abs-2305-13172}, Mitchell et al. \cite{DBLP:conf/icml/MitchellLBMF22}, etc.}, brain_work] 
                            ]
                            [Mitigate hallucination,  brain
                                [{Manakul et al. \cite{DBLP:journals/corr/abs-2303-08896}, Qin et al. \cite{DBLP:journals/corr/abs-2304-08354}, Li et al. \cite{DBLP:journals/corr/abs-2305-14623}, Gou et al. \cite{DBLP:journals/corr/abs-2305-11738}, etc.}, brain_work]
                            ]
                        ]
                    ]
                    [Memory \S\ref{sec:memory}, brain
                        [Memory capability,  brain
                            [Raising the length limit of Transformers, brain
                                [{BART \cite{DBLP:conf/acl/LewisLGGMLSZ20}, Park et al. \cite{DBLP:conf/acl/ParkVS22}, LongT5 \cite{DBLP:conf/naacl/GuoAUONSY22}, CoLT5 \cite{DBLP:journals/corr/abs-2303-09752}, Ruoss et al. \cite{DBLP:conf/acl/RuossDGGCBLV23}, etc. }, brain_work]
                            ]
                            [Summarizing memory, brain
                                [{Generative Agents \cite{DBLP:journals/corr/abs-2304-03442}, SCM \cite{DBLP:journals/corr/abs-2304-13343}, Reflexion \cite{DBLP:journals/corr/abs-2303-11366}, Memorybank \cite{DBLP:journals/corr/abs-2305-10250}, ChatEval \cite{DBLP:journals/corr/abs-2308-07201}, etc.}, brain_work]
                            ]
                            [Compressing memories with vectors or data structures, brain
                                [{ChatDev \cite{DBLP:journals/corr/abs-2307-07924}, GITM \cite{DBLP:journals/corr/abs-2305-17144}, RET-LLM \cite{DBLP:journals/corr/abs-2305-14322}, AgentSims \cite{DBLP:journals/corr/abs-2308-04026}, ChatDB \cite{DBLP:journals/corr/abs-2306-03901}, etc.}, brain_work]
                            ]                          
                        ]
                        [Memory retrieval,  brain
                            [Automated retrieval, brain
                                [{Generative Agents \cite{DBLP:journals/corr/abs-2304-03442}, Memorybank \cite{DBLP:journals/corr/abs-2305-10250}, AgentSims \cite{DBLP:journals/corr/abs-2308-04026}, etc.}, brain_work]
                            ]
                            [Interactive retrieval, brain
                                [{Memory Sandbox\cite{DBLP:journals/corr/abs-2308-01542}, ChatDB \cite{DBLP:journals/corr/abs-2306-03901}, etc.}, brain_work]
                            ]
                        ]
                    ]
                    [Reasoning \& Planning  \S\ref{sec:reasoning and planning},  brain
                        [Reasoning, brain
                            [{CoT \cite{DBLP:conf/nips/Wei0SBIXCLZ22}, Zero-shot-CoT \cite{DBLP:conf/nips/KojimaGRMI22}, Self-Consistency \cite{DBLP:conf/iclr/0002WSLCNCZ23}, Self-Polish \cite{DBLP:journals/corr/abs-2305-14497}, Selection-Inference \cite{DBLP:conf/iclr/CreswellSH23}, Self-Refine \cite{DBLP:journals/corr/abs-2303-17651}, etc.}, brain_work]
                        ]
                        [Planing, brain
                            [Plan formulation, brain
                                [{Least-to-Most \cite{DBLP:conf/iclr/ZhouSHWS0SCBLC23}, SayCan \cite{DBLP:conf/corl/IchterBCFHHHIIJ22}, HuggingGPT \cite{DBLP:journals/corr/abs-2303-17580}, ToT \cite{DBLP:journals/corr/abs-2305-10601}, PET \cite{DBLP:journals/corr/abs-2305-02412}, DEPS \cite{DBLP:journals/corr/abs-2302-01560}, RAP \cite{DBLP:journals/corr/abs-2305-14992}, SwiftSage \cite{DBLP:journals/corr/abs-2305-17390}, LLM+P \cite{DBLP:journals/corr/abs-2304-11477}, MRKL \cite{DBLP:journals/corr/abs-2205-00445}, etc.}, brain_work]  
                            ]
                            [Plan reflection, brain
                                [{LLM-Planner \cite{DBLP:journals/corr/abs-2212-04088}, Inner Monologue \cite{DBLP:conf/corl/HuangXXCLFZTMCS22}, ReAct \cite{DBLP:conf/iclr/YaoZYDSN023}, ChatCoT \cite{DBLP:journals/corr/abs-2305-14323}, AI Chains \cite{DBLP:conf/chi/WuTC22}, Voyager \cite{DBLP:journals/corr/abs-2305-16291}, Zhao et al. \cite{DBLP:journals/corr/abs-2303-08268}, SelfCheck \cite{DBLP:journals/corr/abs-2308-00436}, etc.}, brain_work]  
                            ]
                        ]
                    ]
                    [Transferability \& Generalization \S\ref{sec:transferability and generalization},  brain
                        [Unseen task generalization, brain
                            [{T0 \cite{DBLP:conf/iclr/SanhWRBSACSRDBX22}, FLAN \cite{DBLP:conf/iclr/WeiBZGYLDDL22}, InstructGPT \cite{DBLP:conf/nips/Ouyang0JAWMZASR22}, Chung et al. \cite{DBLP:journals/corr/abs-2210-11416}, etc.}, brain_work]
                        ]
                        [In-context learning, brain
                            [{GPT-3 \cite{DBLP:conf/nips/BrownMRSKDNSSAA20}, Wang et al. \cite{DBLP:conf/cvpr/WangWCS023}, Wang et al. \cite{DBLP:journals/corr/abs-2301-02111}, Dong et al. \cite{DBLP:journals/corr/abs-2301-00234}, etc.}, brain_work]
                        ]
                        [Continual learning, brain
                            [{Ke et al. \cite{Ke2022ContinualLO}, Wang et al. \cite{Wang2023ACS}, Razdaibiedina et al. \cite{razdaibiedina2023progressive}, Voyager \cite{DBLP:journals/corr/abs-2305-16291}, etc.}, brain_work]
                        ]
                    ]   
                ]
        \end{forest}
    \end{adjustbox} 
    \caption{Typology of the brain module. }
    \label{fig:sec3_mindmap_brain}
\end{figure*}

%% file: figures/sec3_perception_mindmap.tex
\begin{figure*}[!ht]
\scriptsize
    \begin{adjustbox}{width=\textwidth}
        \begin{forest}
        for tree={
                forked edges,
                grow'=0,
                draw,
                rounded corners,
                node options={align=center},
                text width=2.7cm,
                s sep=6pt,
                calign=edge midpoint, 
            },
                [Perception, fill=gray!45, parent
                    [Textual Input \S\ref{sec:Textual Input}, for tree={perception}
                    ]
                    [Visual Input \S\ref{sec:Visual Input}, perception
                        [Visual encoder, perception
                            [{ViT \cite{DBLP:conf/iclr/DosovitskiyB0WZ21},  VQVAE \cite{DBLP:conf/nips/OordVK17}, MobileViT \cite{DBLP:conf/iclr/MehtaR22},  MLP-Mixer \cite{DBLP:conf/nips/TolstikhinHKBZU21}, etc.}, perception_work]
                        ]
                        [\hphantom{x} Learnable \hphantom{x} architecture, perception
                            [Query based, perception_work
                                [{Kosmos \cite{DBLP:journals/corr/abs-2302-14045}, BLIP-2 \cite{DBLP:conf/icml/0008LSH23}, InstructBLIP \cite{DBLP:journals/corr/abs-2305-06500},  MultiModal-GPT \cite{DBLP:journals/corr/abs-2305-04790}, Flamingo \cite{DBLP:conf/nips/AlayracDLMBHLMM22}, etc.}, perception_work]
                            ]
                            [Projection based, perception_work
                                [{PandaGPT \cite{DBLP:journals/corr/abs-2305-16355}, LLaVA \cite{DBLP:journals/corr/abs-2304-08485}, Minigpt-4 \cite{zhu2023minigpt}, etc.}, perception_work]
                            ]                           
                        ]
                    ]
                    [Auditory Input \S\ref{sec:Auditory Input}, perception
                        [Cascading manner, perception
                            [{AudioGPT \cite{DBLP:journals/corr/abs-2304-12995}, HuggingGPT \cite{DBLP:journals/corr/abs-2303-17580}, etc.}, perception_work]
                        ]
                        [\hphantom{xx} Transfer \hphantom{xxx} visual method, perception
                            [{AST \cite{DBLP:conf/interspeech/GongCG21}, HuBERT \cite{DBLP:journals/taslp/HsuBTLSM21} , X-LLM \cite{DBLP:journals/corr/abs-2305-04160}, Video-LLaMA \cite{DBLP:journals/corr/abs-2306-02858}, etc.}, perception_work]
                        ]
                    ]
                    [Other Input \S\ref{sec:Others Input}, perception
                        [{InternGPT \cite{DBLP:journals/corr/abs-2305-05662}, etc.}, perception_work]
                    ]
                ] 
        \end{forest}
    \end{adjustbox} 
    \caption{Typology of the perception module.}
    \label{fig:sec3_mindmap_perception}
\end{figure*}

%% file: figures/sec3_action_mindmap.tex
\begin{figure*}[!ht]
\scriptsize
    \begin{adjustbox}{width=\textwidth}
        \begin{forest}
        for tree={
                forked edges,
                grow'=0,
                draw,
                rounded corners,
                node options={align=center},
                text width=2.7cm,
                s sep=6pt,
                calign=edge midpoint, 
            },
                [Action , fill=gray!45, parent
                    [Textual Output \S\ref{sec:Textual Output} , for tree={action}
                    ]
                    [Tools \S\ref{sec:Tool Using}, action
                        [Learning tools, action
                            [{Toolformer \cite{DBLP:journals/corr/abs-2302-04761}, TALM \cite{DBLP:journals/corr/abs-2205-12255}, InstructGPT \cite{DBLP:conf/nips/Ouyang0JAWMZASR22}, Clarebout et al. \cite{Clarebout2013}, etc.}, action_work]
                        ]
                        [Using tools, action
                            [ {WebGPT \cite{DBLP:journals/corr/abs-2112-09332}, OpenAGI \cite{DBLP:journals/corr/abs-2304-04370}, Visual ChatGPT \cite{DBLP:journals/corr/abs-2303-04671}, SayCan \cite{DBLP:conf/corl/IchterBCFHHHIIJ22}, etc.}, action_work]
                        ]            
                        [Making tools, action
                            [{LATM \cite{DBLP:journals/corr/abs-2305-17126}, CREATOR \cite{DBLP:journals/corr/abs-2305-14318}, SELF-DEBUGGING \cite{DBLP:journals/corr/abs-2304-05128}, etc.}, action_work]
                        ]
                    ]
                    [Embodied \hphantom{x} Action \S\ref{sec:Embodied Action}, action
                        [\hphantom{x} LLM-based \hphantom{x} Embodied actions, action
                            [{SayCan \cite{DBLP:conf/corl/IchterBCFHHHIIJ22}, EmbodiedGPT \cite{DBLP:journals/corr/abs-2305-15021}, InstructRL \cite{liu2022instruction}, Lynch et al. \cite{DBLP:journals/corr/abs-2210-06407}, Voyager \cite{DBLP:journals/corr/abs-2305-16291}, AlphaBlock \cite{DBLP:journals/corr/abs-2305-18898}, DEPS \cite{DBLP:journals/corr/abs-2302-01560}, LM-Nav \cite{DBLP:conf/corl/ShahOIL22}, NavGPT \cite{DBLP:journals/corr/abs-2305-16986}, etc.}, action_work]
                        ]
                        [Prospective to the embodied action , action
                            [{MineDojo \cite{DBLP:conf/nips/FanWJMYZTHZA22}, Kanitscheider et al. \cite{DBLP:journals/corr/abs-2106-14876}, DECKARD \cite{DBLP:conf/icml/NottinghamAS0H023}, Sumers et al. \cite{DBLP:conf/icml/SumersMAF023}, etc.}, action_work]
                        ]
                    ]
                ]
        \end{forest}
    \end{adjustbox} 
    \caption{Typology of the action module.}
    \label{fig:sec3_mindmap_action}
\end{figure*}

%% file: sections/Sec4.Application.tex
\section{Agents in Practice:  Harnessing AI for Good}\label{sec:Agents in Practice:  Harnessing AI for Good}
\input{figures/sec4_mindmap}

The LLM-based agent, as an emerging direction, has gained increasing attention from researchers. Many applications in specific domains and tasks have already been developed, showcasing the powerful and versatile capabilities of agents. We can state with great confidence that, the possibility of having a personal agent capable of assisting users with typical daily tasks is larger than ever before \cite{DBLP:journals/corr/abs-2306-05152}. As an LLM-based agent, its design objective should always be beneficial to humans, i.e., humans can \textit{harness AI for good}. Specifically, we expect the agent to achieve the following objectives:

\begin{enumerate}[leftmargin=*]
    \item Assist users in breaking free from daily tasks and repetitive labor, thereby Alleviating human work pressure and enhancing task-solving efficiency.
    \item No longer necessitates users to provide explicit low-level instructions. Instead, the agent can independently analyze, plan, and solve problems.
    \item After freeing users' hands, the agent also liberates their minds to engage in exploratory and  innovative work, realizing their full potential in cutting-edge scientific fields.
\end{enumerate}

\begin{figure}[t]
    \centering
    \includegraphics[width=0.85 \textwidth]{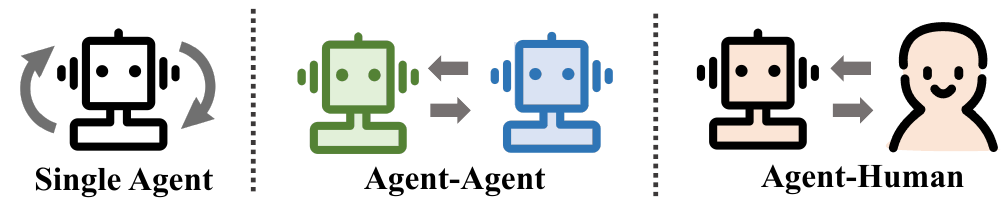}
    \caption{Scenarios of LLM-based agent applications. We mainly introduce three scenarios: single-agent deployment, multi-agent interaction, and human-agent interaction. A \textbf{single agent} possesses diverse capabilities and can demonstrate outstanding task-solving performance in various application orientations. When \textbf{multiple agents} interact, they can achieve advancement through cooperative or adversarial interactions. Furthermore, in \textbf{human-agent} interactions, human feedback can enable agents to perform tasks more efficiently and safely, while agents can also provide better service to humans.}
    \label{fig: sec4_framework}
\end{figure}

In this section, we provide an in-depth overview of current applications of LLM-based agents, aiming to offer a broad perspective for the practical deployment scenarios (see Figure \ref{fig: sec4_framework}). First, we elucidate the diverse application scenarios of Single Agent, including task-oriented, innovation-oriented, and lifecycle-oriented scenarios (\S \ \ref{sec:General Ability of Single Agent}). Then, we present the significant coordinating potential of Multiple Agents. Whether through cooperative interaction for complementarity or adversarial interaction for advancement, both approaches can lead to higher task efficiency and response quality (\S \ \ref{sec:Collaborative Potential of Multi Agents}). Finally, we categorize the interactive collaboration between humans and agents into two paradigms and introduce the main forms and specific applications respectively (\S \ \ref{sec:Interactive Cooperation between Human-Agent}). The topological diagram for LLM-based agent applications is depicted in Figure \ref{fig:sec4_mindmap}.

\subsection{General Ability of Single Agent}\label{sec:General Ability of Single Agent}
Currently, there is a vibrant development of application instances of LLM-based agents \cite{Chase-LangChain-2022, AgentGPT, gptengineer}. AutoGPT \cite{gravitasauto} is one of the ongoing popular open-source projects aiming to achieve a fully autonomous system. Apart from the basic functions of large language models like GPT-4, the AutoGPT framework also incorporates various practical external tools and long/short-term memory management. After users input their customized objectives, they can free their hands and wait for AutoGPT to automatically generate thoughts and perform specific tasks, all without requiring additional user prompts.

\begin{figure}[t]
    \centering
    \includegraphics[width=0.9\textwidth]{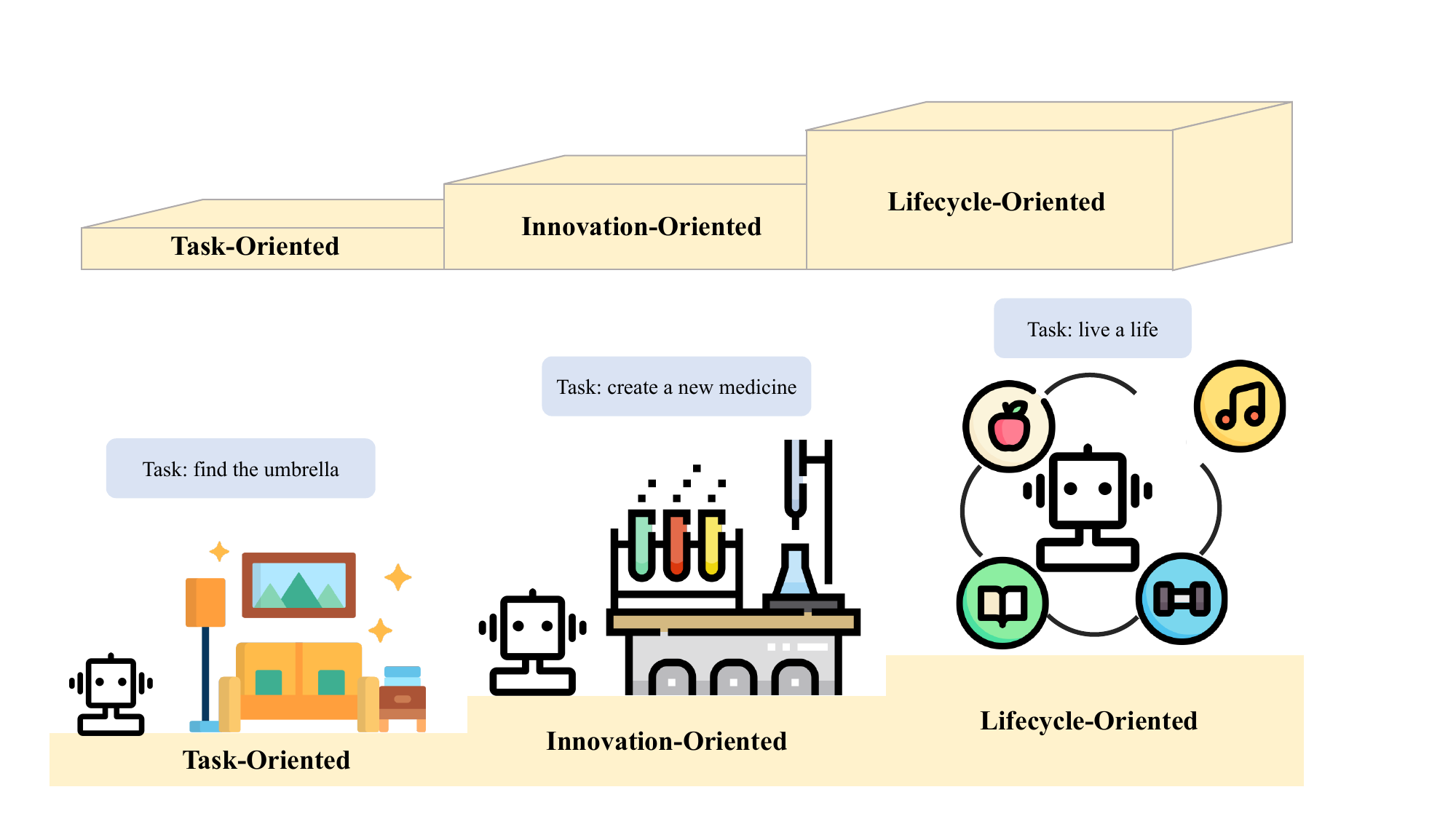}
    \caption{Practical applications of the single LLM-based agent in different scenarios. In \textbf{task-oriented deployment}, agents assist human users in solving daily tasks. They need to possess basic instruction comprehension and task decomposition abilities. In \textbf{innovation-oriented deployment}, agents demonstrate the potential for autonomous exploration in scientific domains. In \textbf{lifecycle-oriented deployment}, agents have the ability to continuously explore, learn, and utilize new skills to ensure long-term survival in an open world.}
    \label{fig: sec4_single_agent}
\end{figure}

As shown in Figure \ref{fig: sec4_single_agent}, we introduce the astonishingly diverse capabilities that the agent exhibits in scenarios where only one single agent is present.

\subsubsection{Task-oriented Deployment}\label{sec:Task-Oriented Deployment}
The LLM-based agents, which can understand human natural language commands and perform everyday tasks \cite{DBLP:journals/corr/abs-2307-13854}, are currently among the most favored and practically valuable agents by users. This is because they have the potential to enhance task efficiency, alleviate user workload, and promote access for a broader user base. In \textbf{task-oriented deployment}, the agent follows high-level instructions from users, undertaking tasks such as goal decomposition \cite{DBLP:journals/corr/abs-2305-02412, DBLP:conf/icml/HuangAPM22, DBLP:journals/corr/abs-2307-12856, DBLP:journals/corr/abs-2306-07863}, sequence planning of sub-goals \cite{DBLP:journals/corr/abs-2305-02412, DBLP:journals/corr/abs-2308-01552}, interactive exploration of the environment \cite{DBLP:journals/corr/abs-2211-09935, DBLP:journals/corr/abs-2307-13854, DBLP:journals/corr/abs-2305-11854, DBLP:conf/nips/Yao0YN22}, until the final objective is achieved.

To explore whether agents can perform basic tasks, they are first deployed in text-based game scenarios. In this type of game, agents interact with the world purely using natural language \cite{DBLP:journals/corr/abs-2012-02757}. By reading textual descriptions of their surroundings and utilizing skills like memory, planning, and trial-and-error \cite{DBLP:journals/corr/abs-2305-02412}, they predict the next action. However, due to the limitation of foundation language models, agents often rely on reinforcement learning during actual execution \cite{DBLP:journals/corr/abs-2012-02757, DBLP:conf/atal/SinghSM22, DBLP:conf/naacl/AmmanabroluULSR21}.

With the gradual evolution of LLMs \cite{chatgpt2022}, agents equipped with stronger text understanding and generation abilities have demonstrated great potential to perform tasks through natural language. Due to their oversimplified nature, naive text-based scenarios have been inadequate as testing grounds for LLM-based agents \cite{DBLP:journals/corr/abs-2307-13854}. More realistic and complex simulated test environments have been constructed to meet the demand. Based on task types, we divide these simulated environments into \textbf{web scenarios} and \textbf{life scenarios}, and introduce the specific roles that agents play in them.

\paragraph{In web scenarios.}
Performing specific tasks on behalf of users in a web scenario is known as the web navigation problem \cite{DBLP:journals/corr/abs-2305-11854}. Agents interpret user instructions, break them down into multiple basic operations, and interact with computers. This often includes web tasks such as filling out forms, online shopping, and sending emails. Agents need to possess the ability to understand instructions within complex web scenarios, adapt to changes (such as noisy text and dynamic HTML web pages), and generalize successful operations \cite{DBLP:journals/corr/abs-2307-13854}. In this way, agents can achieve accessibility and automation when dealing with unseen tasks in the future \cite{DBLP:conf/naacl/XuMDCHLL21}, ultimately freeing humans from repeated interactions with computer UIs.

Agents trained through reinforcement learning can effectively mimic human behavior using predefined actions like typing, searching, navigating to the next page, etc. They perform well in basic tasks such as online shopping \cite{DBLP:conf/nips/Yao0YN22} and search engine retrieval \cite{DBLP:journals/corr/abs-2112-09332}, which have been widely explored. However, agents without LLM capabilities may struggle to adapt to the more realistic and complex scenarios in the real-world Internet. In dynamic, content-rich web pages such as online forums or online business management \cite{DBLP:journals/corr/abs-2307-13854}, agents often face challenges in performance.

In order to enable successful interactions between agents and more realistic web pages, some researchers \cite{DBLP:journals/corr/abs-2303-17491, DBLP:journals/corr/abs-2306-07863} have started to leverage the powerful HTML reading and understanding abilities of LLMs. By designing prompts, they attempt to make agents understand the entire HTML source code and predict more reasonable next action steps. Mind2Web \cite{DBLP:journals/corr/abs-2306-06070} combines multiple LLMs fine-tuned for HTML, allowing them to summarize verbose HTML code \cite{DBLP:journals/corr/abs-2307-12856} in real-world scenarios and extract valuable information. Furthermore, WebGum \cite{DBLP:journals/corr/abs-2305-11854} empowers agents with visual perception abilities by employing a multimodal corpus containing HTML screenshots. It simultaneously fine-tunes the LLM and a visual encoder, deepening the agent's comprehensive understanding of web pages.

\paragraph{In life scenarios.}
In many daily household tasks in life scenarios, it's essential for agents to understand implicit instructions and apply common-sense knowledge \cite{DBLP:conf/atal/SinghSM22}. For an LLM-based agent trained solely on massive amounts of text, tasks that humans take for granted might require multiple trial-and-error attempts \cite{DBLP:journals/corr/abs-2012-02757}. More realistic scenarios often lead to more obscure and subtle tasks. For example, the agent should proactively turn it on if it's dark and there's a light in the room. To successfully chop some vegetables in the kitchen, the agent needs to anticipate the possible location of a knife \cite{DBLP:journals/corr/abs-2305-02412}.

Can an agent apply the world knowledge embedded in its training data to real interaction scenarios? Huang et al. \cite{DBLP:conf/icml/HuangAPM22} lead the way in exploring this question. They demonstrate that sufficiently large LLMs, with appropriate prompts, can effectively break down high-level tasks into suitable sub-tasks without additional training. However, this static reasoning and planning ability has its potential drawbacks. Actions generated by agents often lack awareness of the dynamic environment around them. For instance, when a user gives the task ``clean the room'', the agent might convert it into unfeasible sub-tasks like ``call a cleaning service'' \cite{DBLP:journals/corr/abs-2210-04964}.

To provide agents with access to comprehensive scenario information during interactions, some approaches directly incorporate spatial data and item-location relationships as additional inputs to the model. This allows agents to gain a precise description of their surroundings \cite{DBLP:journals/corr/abs-2308-01552, DBLP:journals/corr/abs-2210-04964}. Wu et al. \cite{DBLP:journals/corr/abs-2305-02412} introduce the PET framework, which mitigates irrelevant objects and containers in environmental information through an early error correction method \cite{DBLP:journals/corr/abs-2211-09935}. PET encourages agents to explore the scenario and plan actions more efficiently, focusing on the current sub-task.

\subsubsection{Innovation-oriented Deployment}\label{sec:Innovation-Oriented Deployment}
The LLM-based agent has demonstrated strong capabilities in performing tasks and enhancing the efficiency of repetitive work. However, in a more intellectually demanding field, like cutting-edge science, the potential of agents has not been fully realized yet. This limitation mainly arises from two challenges \cite{DBLP:journals/corr/abs-2308-01423}: On one hand, the inherent complexity of science poses a significant barrier. Many domain-specific terms and multi-dimensional structures are difficult to represent using a single text. As a result, their complete attributes cannot be fully encapsulated. This greatly weakens the agent's cognitive level. On the other hand, there is a severe lack of suitable training data in scientific domains, making it difficult for agents to comprehend the entire domain knowledge \cite{DBLP:conf/emnlp/WangJCA22, DBLP:journals/corr/abs-2305-05091}. If the ability for autonomous exploration could be discovered within the agent, it would undoubtedly bring about beneficial innovation in human technology.

Currently, numerous efforts in various specialized domains aim to overcome this challenge \cite{DBLP:journals/corr/abs-2306-15626, lin2023evolutionary, DBLP:journals/mlst/IrwinDHB22}. Experts from the computer field make full use of the agent's powerful code comprehension and debugging abilities \cite{DBLP:journals/corr/abs-2306-05152, DBLP:journals/corr/abs-2308-00245}. In the fields of chemistry and materials, researchers equip agents with a large number of general or task-specific tools to better understand domain knowledge. Agents evolve into comprehensive scientific assistants, proficient in online research and document analysis to fill data gaps. They also employ robotic APIs for real-world interactions, enabling tasks like material synthesis and mechanism discovery \cite{DBLP:journals/corr/abs-2304-05332, bran2023chemcrow,  DBLP:journals/corr/abs-2308-01423}.

The potential of LLM-based agents in scientific innovation is evident, yet we do not expect their exploratory abilities to be utilized in applications that could threaten or harm humans. Boiko et al. \cite{DBLP:journals/corr/abs-2304-05332} study the hidden dangers of agents in synthesizing illegal drugs and chemical weapons, indicating that agents could be misled by malicious users in adversarial prompts. This serves as a warning for our future work.

\subsubsection{Lifecycle-oriented Deployment}\label{sec:Lifecycle-Oriented Deployment}
Building a universally capable agent that can continuously explore, develop new skills, and maintain a long-term life cycle in an open, unknown world is a colossal challenge. This accomplishment is regarded as a pivotal milestone in the field of AGI \cite{DBLP:journals/corr/abs-2302-01560}. Minecraft, as a typical and widely explored simulated survival environment, has become a unique playground for developing and testing the comprehensive ability of an agent. Players typically start by learning the basics, such as mining wood and making crafting tables, before moving on to more complex tasks like fighting against monsters and crafting diamond tools \cite{DBLP:journals/corr/abs-2305-16291}. Minecraft fundamentally reflects the real world, making it conducive for researchers to investigate an agent's potential to survive in the authentic world.

The survival algorithms of agents in Minecraft can generally be categorized into two types \cite{DBLP:journals/corr/abs-2305-16291}: \textbf{low-level control} and \textbf{high-level planning}. Early efforts mainly focused on reinforcement learning \cite{DBLP:journals/corr/abs-2305-16291, DBLP:journals/corr/abs-2211-00688} and imitation learning \cite{DBLP:journals/corr/abs-2007-02701}, enabling agents to craft some low-level items. With the emergence of LLMs, which demonstrated surprising reasoning and analytical capabilities, agents begin to utilize LLM as a high-level planner to guide simulated survival tasks \cite{DBLP:journals/corr/abs-2302-01560, DBLP:conf/icml/NottinghamAS0H023}. Some researchers use LLM to decompose high-level task instructions into a series of sub-goals \cite{DBLP:journals/corr/abs-2303-16563}, basic skill sequences \cite{DBLP:conf/icml/NottinghamAS0H023}, or fundamental keyboard/mouse operations \cite{DBLP:journals/corr/abs-2303-16563}, gradually assisting agents in exploring the open world.

Voyager\cite{DBLP:journals/corr/abs-2305-16291}, drawing inspiration from concepts similar to AutoGPT\cite{gravitasauto}, became the first LLM-based embodied lifelong learning agent in Minecraft, based on the long-term goal of ``discovering as many diverse things as possible''. It introduces a skill library for storing and retrieving complex action-executable code, along with an iterative prompt mechanism that incorporates environmental feedback and error correction. This enables the agent to autonomously explore and adapt to unknown environments without human intervention. An AI agent capable of autonomously learning and mastering the entire real-world techniques may not be as distant as once thought \cite{DBLP:journals/corr/abs-2303-16563}.

\subsection{Coordinating Potential of Multiple Agents} \label{sec:Collaborative Potential of Multi Agents}
\paragraph{Motivation and Background.} Although LLM-based agents possess commendable text understanding and generation capabilities, they operate as isolated entities in nature \cite{DBLP:journals/corr/abs-2306-03314}. They lack the ability to collaborate with other agents and acquire knowledge from social interactions. This inherent limitation restricts their potential to learn from multi-turn feedback from others to enhance their performance \cite{DBLP:journals/corr/abs-2305-16960}. Moreover, they cannot be effectively deployed in complex scenarios requiring collaboration and information sharing among multiple agents.

 As early as 1986, Marvin Minsky made a forward-looking prediction. In his book \textit{The Society of Mind} \cite{minsky1988society}, he introduced a novel theory of intelligence, suggesting that intelligence emerges from the interactions of many smaller agents with specific functions. For instance, certain agents might be responsible for pattern recognition, while others might handle decision-making or generate solutions. This idea has been put into concrete practice with the rise of distributed artificial intelligence \cite{balaji2010introduction}. Multi-agent systems(MAS) \cite{DBLP:journals/ker/WooldridgeJ95}, as one of the primary research domains, focus on how a group of agents can effectively coordinate and collaborate to solve problems. Some specialized communication languages, like KQML \cite{DBLP:conf/cikm/FininFMM94}, were designed early on to support message transmission and knowledge sharing among agents. However, their message formats were relatively fixed, and the semantic expression capacity was limited. In the 21st century, integrating reinforcement learning algorithms (such as Q-learning) with deep learning has become a prominent technique for developing MAS that operate in complex environments \cite{yang2020overview}. Nowadays, the construction approach based on LLMs is beginning to demonstrate remarkable potential. The natural language communication between agents has become more elegant and easily comprehensible to humans, resulting in a significant leap in interaction efficiency.

\paragraph{Potential advantages.}
Specifically, an LLM-based multi-agent system can offer several advantages. Just as Adam Smith clearly stated in \textit{The Wealth of Nations} \cite{smith1937wealth}, ``The greatest improvements in the productive powers of labor, and most of the skill, dexterity, and judgment with which it is directed or applied, seem to be results of the division of labor.'' Based on the principle of division of labor, a single agent equipped with specialized skills and domain knowledge can engage in specific tasks. On the one hand, agents' skills in handling specific tasks are increasingly refined through the division of labor. On the other hand, decomposing complex tasks into multiple subtasks can eliminate the time spent switching between different processes. In the end, efficient division of labor among multiple agents can accomplish a significantly greater workload than when there is no specialization, substantially improving the overall system's efficiency and output quality.

\begin{figure}[t]
    \centering
    \includegraphics[width=1 \textwidth]{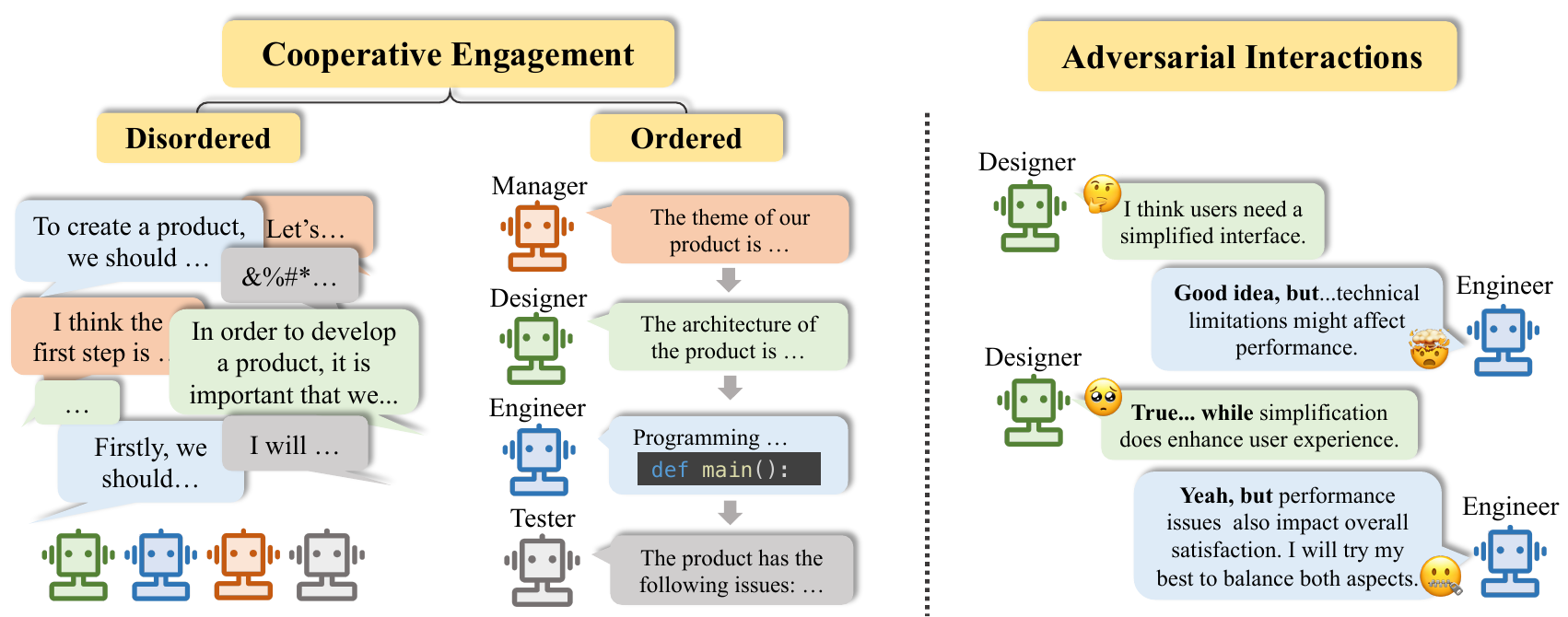}
    \caption{Interaction scenarios for multiple LLM-based agents. In \textbf{cooperative interaction}, agents collaborate in either a disordered or ordered manner to achieve shared objectives. In \textbf{adversarial interaction}, agents compete in a tit-for-tat fashion to enhance their respective performance.}
    \label{fig: sec4_multi_agent}
\end{figure}

%
In \S \  \ref{sec:General Ability of Single Agent}, we have provided a comprehensive introduction to the versatile abilities of LLM-based agents. Therefore, in this section, we focus on exploring the ways agents interact with each other in a multi-agent environment. Based on current research, these interactions can be broadly categorized as follows: \textbf{Cooperative Interaction for Complementarity} and \textbf{Adversarial Interaction for Advancement} (see Figure \ref{fig: sec4_multi_agent}).

\subsubsection{Cooperative Interaction for Complementarity} \label{sec:Cooperative Engagement for Complementarity}
Cooperative multi-agent systems are the most widely deployed pattern in practical usage. Within such systems, individual agent assesses the needs and capabilities of other agents and actively seeks collaborative actions and information sharing with them \cite{DBLP:journals/corr/abs-2303-17760}. This approach brings forth numerous potential benefits, including enhanced task efficiency, collective decision improvement, and the resolution of complex real-world problems that one single agent cannot solve independently, ultimately achieving the goal of synergistic complementarity. In current LLM-based multi-agent systems, communication between agents predominantly employs natural language, which is considered the most natural and human-understandable form of interaction \cite{DBLP:journals/corr/abs-2303-17760}. We introduce and categorize existing cooperative multi-agent applications into two types: disordered cooperation and ordered cooperation.

\paragraph{Disordered cooperation.}
When three or more agents are present within a system, each agent is free to express their perspectives and opinions openly. They can provide feedback and suggestions for modifying responses related to the task at hand \cite{DBLP:journals/corr/abs-2307-04738}. This entire discussion process is uncontrolled, lacking any specific sequence, and without introducing a standardized collaborative workflow. We refer to this kind of multi-agent cooperation as \textbf{disordered cooperation}.

ChatLLM network \cite{DBLP:journals/corr/abs-2304-12998} is an exemplary representative of this concept. It emulates the forward and backward propagation process within a neural network, treating each agent as an individual node. Agents in the subsequent layer need to process inputs from all the preceding agents and propagate forward. One potential solution is introducing a dedicated coordinating agent in multi-agent systems, responsible for integrating and organizing responses from all agents, thus updating the final answer \cite{DBLP:journals/corr/abs-2307-05300}. However, consolidating a large amount of feedback data and extracting valuable insights poses a significant challenge for the coordinating agent.

Furthermore, \textbf{majority voting} can also serve as an effective approach to making appropriate decisions. However, there is limited research that integrates this module into multi-agent systems at present. Hamilton \cite{DBLP:journals/corr/abs-2301-05327} trains nine independent supreme justice agents to better predict judicial rulings in the U.S. Supreme Court, and decisions are made through a majority voting process.

\paragraph{Ordered cooperation.}
When agents in the system adhere to specific rules, for instance, expressing their opinions one by one in a sequential manner, downstream agents only need to focus on the outputs from upstream. This leads to a significant improvement in task completion efficiency, The entire discussion process is highly organized and ordered. We term this kind of multi-agent cooperation as \textbf{ordered cooperation}. It's worth noting that systems with only two agents, essentially engaging in a conversational manner through a back-and-forth interaction, also fall under the category of ordered cooperation.

CAMEL \cite{DBLP:journals/corr/abs-2303-17760} stands as a successful implementation of a dual-agent cooperative system. Within a role-playing communication framework, agents take on the roles of AI Users (giving instructions) and AI Assistants (fulfilling requests by providing specific solutions). Through multi-turn dialogues, these agents autonomously collaborate to fulfill user instructions \cite{DBLP:journals/corr/abs-2303-17071}. Some researchers have integrated the idea of dual-agent cooperation into a single agent's operation \cite{DBLP:journals/corr/abs-2305-17390}, alternating between rapid and deliberate thought processes to excel in their respective areas of expertise.

Talebirad et al. \cite{DBLP:journals/corr/abs-2306-03314} are among the first to systematically introduce a comprehensive LLM-based multi-agent collaboration framework.  This paradigm aims to harness the strengths of each individual agent and foster cooperative relationships among them. Many applications of multi-agent cooperation have successfully been built upon this foundation \cite{DBLP:journals/corr/abs-2305-16960,DBLP:journals/corr/abs-2308-08155,DBLP:journals/corr/abs-2308-11339,DBLP:journals/corr/abs-2305-13657}. Furthermore, AgentVerse \cite{DBLP:journals/corr/abs-2308-10848} constructs a versatile, multi-task-tested framework for group agents cooperation. It can assemble a team of agents that dynamically adapt according to the task's complexity. To promote more efficient collaboration, researchers hope that agents can learn from successful human cooperation examples \cite{DBLP:journals/corr/abs-2307-07924}. MetaGPT \cite{DBLP:journals/corr/abs-2308-00352} draws inspiration from the classic waterfall model in software development, standardizing agents' inputs/outputs as engineering documents. By encoding advanced human process management experience into agent prompts, collaboration among multiple agents becomes more structured.

However, during MetaGPT's practical exploration, a potential threat to multi-agent cooperation has been identified. Without setting corresponding rules, frequent interactions among multiple agents can amplify minor hallucinations indefinitely \cite{DBLP:journals/corr/abs-2308-00352}. For example, in software development, issues like incomplete functions, missing dependencies, and bugs that are imperceptible to the human eye may arise. Introducing techniques like cross-validation \cite{DBLP:journals/corr/abs-2307-07924} or timely external feedback could have a positive impact on the quality of agent outputs.

\subsubsection{Adversarial Interaction for Advancement} \label{sec:Adversarial Interactions for Advancement}
Traditionally, cooperative methods have been extensively explored in multi-agent systems. However, researchers increasingly recognize that introducing concepts from game theory \cite{DBLP:books/daglib/0027240,DBLP:journals/sigact/Aziz10} into systems can lead to more robust and efficient behaviors. In competitive environments, agents can swiftly adjust strategies through dynamic interactions, striving to select the most advantageous or rational actions in response to changes caused by other agents. Successful applications in Non-LLM-based competitive domains already exist \cite{DBLP:journals/nature/SilverHMGSDSAPL16,Campbell2002DeepB}. AlphaGo Zero \cite{DBLP:journals/nature/SilverSSAHGHBLB17}, for instance, is an agent for Go that achieved significant breakthroughs through a process of self-play. Similarly, within LLM-based multi-agent systems, fostering change among agents can naturally occur through competition, argumentation, and debate \cite{DBLP:conf/emnlp/LewisYDPB17,DBLP:journals/corr/abs-1805-00899}. By abandoning rigid beliefs and engaging in thoughtful reflection, \textbf{adversarial interaction} enhances the quality of responses.

Researchers first delve into the fundamental debating abilities of LLM-based agents \cite{DBLP:journals/corr/abs-2305-10142,DBLP:journals/corr/abs-2305-11595}. Findings demonstrate that when multiple agents express their arguments in the state of  ``tit for tat'', one agent can receive substantial external feedback from other agents, thereby correcting its distorted thoughts \cite{DBLP:journals/corr/abs-2305-19118}. Consequently, multi-agent adversarial systems find broad applicability in scenarios requiring high-quality responses and accurate decision-making. In reasoning tasks, Du et al. \cite{DBLP:journals/corr/abs-2305-14325} introduce the concept of debate, endowing agents with responses from fellow peers. When these responses diverge from an agent's own judgments, a ``mental'' argumentation occurs, leading to refined solutions. ChatEval \cite{DBLP:journals/corr/abs-2308-07201} establishes a role-playing-based multi-agent referee team. Through self-initiated debates, agents evaluate the quality of text generated by LLMs, reaching a level of excellence comparable to human evaluators.

The performance of the multi-agent adversarial system has shown considerable promise. However, the system is essentially dependent on the strength of LLMs and faces several basic challenges:

\begin{itemize}[leftmargin=*]
    \item With prolonged debate, LLM's limited context cannot process the entire input. 
    \item In a multi-agent environment, computational overhead significantly increases. 
    \item Multi-agent negotiation may converge to an incorrect consensus, and all agents are firmly convinced of its accuracy \cite{DBLP:journals/corr/abs-2305-14325}. 
\end{itemize}

The development of multi-agent systems is still far from being mature and feasible. Introducing human guides when appropriate to compensate for agents' shortcomings is a good choice to promote the further advancements of agents.

\subsection{Interactive Engagement between Human and Agent}\label{sec:Interactive Cooperation between Human-Agent}
\begin{figure}[t]
    \centering
    \includegraphics[width=1 \textwidth]{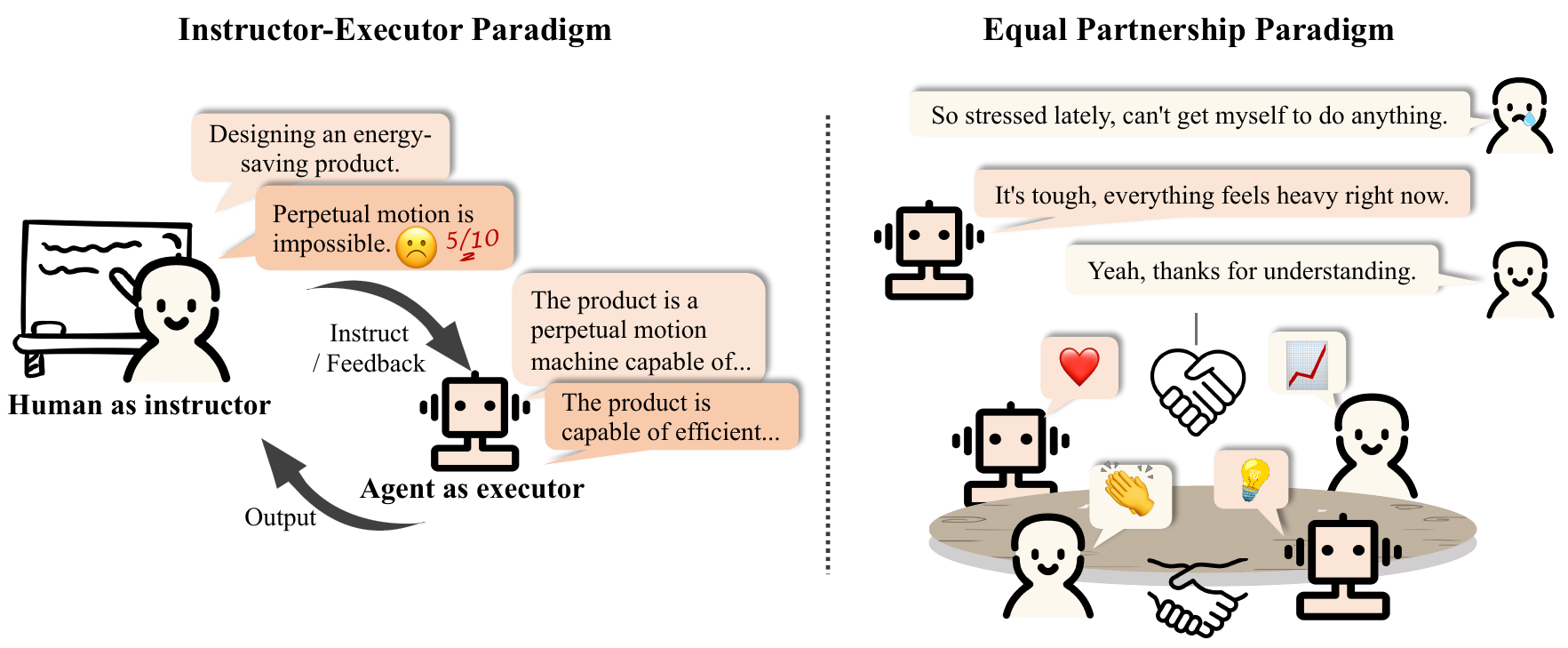}
    \caption{Two paradigms of human-agent interaction. In the instructor-executor paradigm (left), humans provide instructions or feedback, while agents act as executors. In the equal partnership paradigm (right), agents are human-like, able to engage in empathetic conversation and participate in collaborative tasks with humans.}
    \label{fig: human-agent}
\end{figure}

Human-agent interaction, as the name suggests, involves agents collaborating with humans to accomplish tasks. 
With the enhancement of agent capabilities, human involvement becomes progressively essential to effectively guide and oversee agents' actions, ensuring they align with human requirements and objectives \cite{DBLP:journals/corr/abs-2103-14659, DBLP:journals/corr/abs-2209-00626}. 
Throughout the interaction, humans play a pivotal role by offering guidance or by regulating the safety, legality, and ethical conduct of agents. 
This is particularly crucial in specialized domains, such as medicine where data privacy concerns exist \cite{paul2023digitization}. In such cases, human involvement can serve as a valuable means to compensate for the lack of data, thereby facilitating smoother and more secure collaborative processes. 
Moreover, considering the anthropological aspect, language acquisition in humans predominantly occurs through communication and interaction \cite{bassiri2011interactional}, as opposed to merely consuming written content. 
As a result, agents shouldn't exclusively depend on models trained with pre-annotated datasets; instead, they should evolve through online interaction and engagement. 
The interaction between humans and agents can be classified into two paradigms (see Figure \ref{fig: human-agent}): (1) Unequal interaction (i.e., instructor-executor paradigm): humans serve as issuers of instructions, while agents act as executors, essentially participating as assistants to humans in collaboration. (2) Equal interaction (i.e., equal partnership paradigm): agents reach the level of humans, participating on an equal footing with humans in interaction.

\subsubsection{Instructor-Executor Paradigm} \label{Instructor-Executor Paradigm}
The simplest approach involves human guidance throughout the process: humans provide clear and specific instructions directly, while the agents' role is to understand natural language commands from humans and translate them into corresponding actions \cite{DBLP:journals/aim/TellexKDWBTR11, DBLP:conf/iser/MatuszekHZF12, DBLP:conf/aaai/ChaplotSPRS18}. In \S \ref{sec:General Ability of Single Agent}, we have presented the scenario where agents solve single-step problems or receive high-level instructions from humans. Considering the interactive nature of language, in this section, we assume that the dialogue between humans and agents is also interactive. Thanks to LLMs, the agents are able to interact with humans in a conversational manner: the agent responds to each human instruction, refining its action through alternating iterations to ultimately meet human requirements \cite{DBLP:journals/corr/abs-2305-16291}. While this approach does achieve the goal of human-agent interaction, it places significant demands on humans. It requires a substantial amount of human effort and, in certain tasks, might even necessitate a high level of expertise. To alleviate this issue, the agent can be empowered to autonomously accomplish tasks, while humans only need to provide feedback in certain circumstances. Here, we roughly categorize feedback into two types: quantitative feedback and qualitative feedback.

\paragraph{Quantitative feedback.}
The forms of quantitative feedback mainly include absolute evaluations like binary scores and ratings, as well as relative scores. Binary feedback refers to the positive and negative evaluations provided by humans, which agents utilize to enhance their self-optimization \cite{DBLP:conf/iclr/LiMCRW17, DBLP:conf/acl/IyerKCKZ17, DBLP:conf/nips/Weston16, DBLP:journals/corr/abs-2208-03188, DBLP:journals/corr/abs-2204-03685}. Comprising only two categories, this type of user feedback is often easy to collect, but sometimes it may oversimplify user intent by neglecting potential intermediate scenarios. To showcase these intermediate scenarios, researchers attempt to expand from binary feedback to rating feedback, which involves categorizing into more fine-grained levels. However, the results of Kreutzer et al. \cite{DBLP:conf/naacl/KreutzerKMR18} suggest that there could be significant discrepancies between user and expert annotations for such multi-level artificial ratings, indicating that this labeling method might be inefficient or less reliable. Furthermore, agents can learn human preference from comparative scores like multiple choice \cite{DBLP:conf/acl/YavuzGSY18, DBLP:conf/emnlp/YaoSSY19}.

\paragraph{Qualitative feedback.}
Text feedback is usually offered in natural language, particularly for responses that may need improvement. The format of this feedback is quite flexible. Humans provide advice on how to modify outputs generated by agents, and the agents then incorporate these suggestions to refine their subsequent outputs \cite{DBLP:conf/naacl/MehtaG19, DBLP:conf/naacl/ElgoharyMRFRA21}. For agents without multimodal perception capabilities, humans can also act as critics, offering visual critiques \cite{DBLP:journals/corr/abs-2305-16291}, for instance. Additionally, agents can utilize a memory module to store feedback for future reuse \cite{DBLP:conf/naacl/TandonMCY22}. In \cite{DBLP:journals/corr/abs-2303-16755}, humans give feedback on the initial output generated by agents, prompting the agents to formulate various improvement proposals. The agents then discern and adopt the most suitable proposal, harmonizing with the human feedback. While this approach can better convey human intention compared to quantitative feedback, it might be more challenging for the agents to comprehend. Xu et al. \cite{DBLP:conf/acl/XuUKABW23} compare various types of feedback and observe that combining multiple types of feedback can yield better results. Re-training models based on feedback from multiple rounds of interaction (i.e., continual learning) can further enhance effectiveness. Of course, the collaborative nature of human-agent interaction also allows humans to directly improve the content generated by agents. This could involve modifying intermediate links \cite{DBLP:conf/chi/WuTC22, DBLP:journals/corr/abs-2306-07932} or adjusting the conversation content \cite{DBLP:journals/corr/abs-2307-07047}. In some studies, agents can autonomously judge whether the conversation is proceeding smoothly and seek feedback when errors are generated \cite{DBLP:conf/acl/HancockBMW19, DBLP:journals/corr/abs-2304-10750}. Humans can also choose to participate in feedback at any time, guiding the agent's learning in the right direction \cite{DBLP:conf/iclr/SchickYJPLIYNG023}.

Currently, in addition to tasks like writing \cite{DBLP:journals/corr/abs-2204-03685} and semantic parsing \cite{DBLP:conf/acl/IyerKCKZ17, DBLP:conf/naacl/ElgoharyMRFRA21}, the model of using agents as human assistants also holds tremendous potential in the field of education. For instance, Kalvakurth et al. \cite{DBLP:journals/corr/abs-2303-13548} propose the robot Dona, which supports multimodal interactions to assist students with registration. Gvirsman et al. \cite{DBLP:conf/hri/GvirsmanKNG20} focus on early childhood education, achieving multifaceted interactions between young children, parents, and agents. Agents can also aid in human understanding and utilization of mathematics \cite{DBLP:journals/corr/abs-2307-02502}. 
In the field of medicine, some medical agents have been proposed, showing enormous potential in terms of diagnosis assistance, consultations, and more \cite{DBLP:journals/corr/abs-2305-15075,DBLP:journals/corr/abs-2308-03549}. 
Especially in mental health, research has shown that agents can lead to increased accessibility due to benefits such as reduced cost, time efficiency, and anonymity compared to face-to-face treatment \cite{doi:10.1177/2055207617713827}. 
Leveraging such advantages, agents have found widespread applications. Ali et al. \cite{DBLP:conf/iva/AliRLMKRSH20} design LISSA for online communication with adolescents on the autism spectrum, analyzing users' speech and facial expressions in real-time to engage them in multi-topic conversations and provide instant feedback regarding non-verbal cues. Hsu et al. \cite{hsu2023helping} build contextualized language generation approaches to provide tailored assistance for users who seek support on diverse topics ranging from relationship stress to anxiety. 
Furthermore, in other industries including business, a good agent possesses the capability to provide automated services or assist humans in completing tasks, thereby effectively reducing labor costs \cite{DBLP:conf/ijcnn/GaoGT23}. 
Amidst the pursuit of AGI, efforts are directed towards enhancing the multifaceted capabilities of general agents, creating agents that can function as universal assistants in real-life scenarios \cite{DBLP:journals/corr/abs-2306-08640}.

\subsubsection{Equal Partnership Paradigm} \label{Equal Partnership Paradigm}
\paragraph{Empathetic communicator.}
With the rapid development of AI, conversational agents have garnered extensive attention in research fields in various forms, such as personalized custom roles and virtual chatbots \cite{DBLP:series/synthesis/2020McTear}. It has found practical applications in everyday life, business, education, healthcare, and more \cite{DBLP:journals/corr/abs-2106-10901, DBLP:journals/ijmms/RappCB21, adamopoulou2020chatbots}. However, in the eyes of the public, agents are perceived as emotionless machines, and can never replace humans. Although it is intuitive that agents themselves do not possess emotions, can we enable them to exhibit emotions and thereby bridge the gap between agents and humans? Therefore, a plethora of research endeavors have embarked on delving into the empathetic capacities of agents. This endeavor seeks to infuse a human touch into these agents, enabling them to detect sentiments and emotions from human expressions, ultimately crafting emotionally resonant dialogues \cite{DBLP:conf/ijcai/Wang018, DBLP:conf/acl/WangZ18, lin2019caire, DBLP:journals/corr/abs-2110-03949, DBLP:conf/emnlp/LinMSXF19, DBLP:conf/emnlp/MajumderHPLGGMP20, DBLP:conf/aaai/SabourZH22, DBLP:conf/aaai/LiLRRC22}. Apart from generating emotionally charged language, agents can dynamically adjust their emotional states and display them through facial expressions and voice \cite{DBLP:journals/corr/abs-2308-03022}. These studies, viewing agents as empathetic communicators, not only enhance user satisfaction but also make significant progress in fields like healthcare \cite{ hsu2023helping,DBLP:conf/iva/AliRLMKRSH20, DBLP:journals/cbsn/LiuS18} and business marketing \cite{liu2022artificial}. Unlike simple rule-based conversation agents, agents with empathetic capacities can tailor their interactions to meet users' emotional needs \cite{DBLP:conf/amia/SuFJ0C20}.

\paragraph{Human-level participant.}
Furthermore, we hope that agents can be involved in the normal lives of humans, cooperating with humans to complete tasks from a human-level perspective. In the field of games, agents have already reached a high level. As early as the 1990s, IBM introduced the AI Deep Blue \cite{Campbell2002DeepB}, which defeated the reigning world champion in chess at that time. However, in pure competitive environments such as chess \cite{Campbell2002DeepB}, Go \cite{DBLP:journals/nature/SilverHMGSDSAPL16}, and poker \cite{DBLP:journals/corr/MoravcikSBLMBDW17}, the value of communication was not emphasized \cite{meta2022human}. In many gaming tasks, players need to collaborate with each other, devising unified cooperative strategies through effective negotiation \cite{DBLP:conf/iclr/Bakhtin0LGJFMB23,meta2022human,DBLP:conf/nips/CarrollSHGSAD19, bard2020hanabi}. In these scenarios, agents need to first understand the beliefs, goals, and intentions of others, formulate joint action plans for their objectives, and also provide relevant suggestions to facilitate the acceptance of cooperative actions by other agents or humans. In comparison to pure agent cooperation, we desire human involvement for two main reasons: first, to ensure interpretability, as interactions between pure agents could generate incomprehensible language \cite{DBLP:conf/nips/CarrollSHGSAD19}; second, to ensure controllability, as the pursuit of agents with complete ``free will'' might lead to unforeseen negative consequences, carrying the potential for disruption. Apart from gaming scenarios, agents also demonstrate human-level capabilities in other scenarios involving human interaction, showcasing skills in strategy formulation, negotiation, and more. Agents can collaborate with one or multiple humans, determining the shared knowledge among the cooperative partners, identifying which information is relevant to decision-making, posing questions, and engaging in reasoning to complete tasks such as allocation, planning, and scheduling \cite{DBLP:journals/corr/abs-2305-20076}. Furthermore, agents possess persuasive abilities \cite{DBLP:conf/acl/WangSKOYZY19}, dynamically influencing human viewpoints in various interactive scenarios \cite{DBLP:journals/corr/abs-2308-03313}.

The goal of the field of human-agent interaction is to learn and understand humans, develop technology and tools based on human needs, and ultimately enable comfortable, efficient, and secure interactions between humans and agents. Currently, significant breakthroughs have been achieved in terms of usability in this field. In the future, human-agent interaction will continue to focus on enhancing user experience, enabling agents to better assist humans in accomplishing more complex tasks in various domains. The ultimate aim is not to make agents more powerful but to better equip humans with agents. Considering practical applications in daily life, isolated interactions between humans and agents are not realistic. Robots will become colleagues, assistants, and even companions. Therefore, future agents will be integrated into a social network \cite{DBLP:journals/ijsr/AbramsP20}, embodying a certain level of social value.

%% file: figures/sec4_mindmap.tex
\begin{figure*}[!ht]
\scriptsize
    \begin{adjustbox}{width=\textwidth}
        \begin{forest}
        for tree={
                forked edges,
                grow'=0,
                draw,
                rounded corners,
                node options={align=center},
                text width=2.7cm,
                s sep=6pt,
                calign=edge midpoint, 
            },
            [Agents in Practice:  Harnessing AI for Good, fill=gray!45, parent
                [Single Agent Deployment \S\ref{sec:General Ability of Single Agent}, for tree={single_agent}
                    [Task-oriented Deploytment \S\ref{sec:Task-Oriented Deployment}, single_agent 
                        [Web scenarios, single_agent
                            [{WebAgent \cite{DBLP:journals/corr/abs-2307-12856}, Mind2Web \cite{DBLP:journals/corr/abs-2306-06070}, WebGum \cite{DBLP:journals/corr/abs-2305-11854}, WebArena \cite{DBLP:journals/corr/abs-2307-13854}, Webshop \cite{DBLP:conf/nips/Yao0YN22}, WebGPT \cite{DBLP:journals/corr/abs-2112-09332}, Kim et al. \cite{DBLP:journals/corr/abs-2303-17491}, Zheng et al. \cite{DBLP:journals/corr/abs-2306-07863}, etc.}, single_agent_work]
                        ]
                        [Life scenarios, single_agent
                            [{InterAct \cite{DBLP:journals/corr/abs-2308-01552}, PET \cite{DBLP:journals/corr/abs-2305-02412}, Huang et al. \cite{DBLP:conf/icml/HuangAPM22}, Gramopadhye et al. \cite{DBLP:journals/corr/abs-2210-04964}, Raman et al. \cite{DBLP:journals/corr/abs-2211-09935}, etc.}, single_agent_work]
                        ]
                    ]
                    [Innovation-oriented Deploytment \S\ref{sec:Innovation-Oriented Deployment},single_agent
                        [{Li et al. \cite{DBLP:journals/corr/abs-2308-00245}, Feldt et al. \cite{DBLP:journals/corr/abs-2306-05152}, ChatMOF \cite{DBLP:journals/corr/abs-2308-01423}, ChemCrow \cite{bran2023chemcrow}, Boiko et al. \cite{DBLP:journals/corr/abs-2304-05332}, SCIENCEWORLD et al. \cite{DBLP:conf/emnlp/WangJCA22}, etc.}, single_agent_work]
                    ]
                    [Lifecycle-oriented Deploytment \S\ref{sec:Lifecycle-Oriented Deployment},single_agent
                        [{Voyager \cite{DBLP:journals/corr/abs-2305-16291}, GITM \cite{DBLP:journals/corr/abs-2305-17144}, DEPS \cite{DBLP:journals/corr/abs-2302-01560}, Plan4MC \cite{DBLP:journals/corr/abs-2303-16563}, Nottingham et al. \cite{DBLP:conf/icml/NottinghamAS0H023}, etc.}, single_agent_work]
                    ]
                ]
                [Multi-Agents Interaction \S\ref{sec:Collaborative Potential of Multi Agents}, for tree={multi_agent}
                    [Cooperative Interaction  \S\ref{sec:Cooperative Engagement for Complementarity}, multi_agent 
                        [Disordered cooperation, multi_agent
                            [{ChatLLM \cite{DBLP:journals/corr/abs-2304-12998}, RoCo \cite{DBLP:journals/corr/abs-2307-04738}, Blind Judgement \cite{DBLP:journals/corr/abs-2301-05327}, etc.}, multi_agent_work]
                        ]
                        [Ordered cooperation, multi_agent
                            [{MetaGPT \cite{DBLP:journals/corr/abs-2308-00352}, ChatDev \cite{DBLP:journals/corr/abs-2307-07924}, CAMEL \cite{DBLP:journals/corr/abs-2303-17760}, AutoGen \cite{DBLP:journals/corr/abs-2308-08155}, SwiftSage \cite{DBLP:journals/corr/abs-2305-17390}, ProAgent \cite{DBLP:journals/corr/abs-2308-11339}, DERA \cite{DBLP:journals/corr/abs-2303-17071}, Talebirad et al. \cite{DBLP:journals/corr/abs-2306-03314}, AgentVerse \cite{DBLP:journals/corr/abs-2308-10848}, CGMI \cite{DBLP:journals/corr/abs-2308-12503}, Liu et al. \cite{DBLP:journals/corr/abs-2305-16960}, etc.}, multi_agent_work]
                        ]
                    ]
                    [Adversarial \hphantom{x} Interaction \S\ref{sec:Adversarial Interactions for Advancement}, multi_agent
                        [{ChatEval \cite{DBLP:journals/corr/abs-2308-07201}, Xiong et al. \cite{DBLP:journals/corr/abs-2305-11595}, Du et al. \cite{DBLP:journals/corr/abs-2305-14325}, Fu et al. \cite{DBLP:journals/corr/abs-2305-10142}, Liang et al. \cite{DBLP:journals/corr/abs-2305-19118}, etc.}, multi_agent_work]
                    ]
                ]
                [Human-Agent Interaction \S\ref{sec:Interactive Cooperation between Human-Agent}, for tree={human_agent}
                    [Instructor-Executor Paradigm \S\ref{Instructor-Executor Paradigm},human_agent
                        [Education, human_agent
                           [{Dona \cite{DBLP:journals/corr/abs-2303-13548}, Math Agents \cite{DBLP:journals/corr/abs-2307-02502}, etc.}, human_agent_work]
                        ]
                        [Health, human_agent
                           [{Hsu et al. \cite{hsu2023helping}, HuatuoGPT \cite{DBLP:journals/corr/abs-2305-15075}, Zhongjing \cite{DBLP:journals/corr/abs-2308-03549}, LISSA \cite{DBLP:conf/iva/AliRLMKRSH20}, etc.}, human_agent_work]
                        ]
                        [Other Applications, human_agent
                           [{Gao et al. \cite{DBLP:conf/ijcnn/GaoGT23}, PEER \cite{DBLP:conf/iclr/SchickYJPLIYNG023}, DIALGEN \cite{DBLP:journals/corr/abs-2307-07047}, AssistGPT \cite{DBLP:journals/corr/abs-2306-08640}, etc.}, human_agent_work]
                        ]
                    ]
                    [Equal Partnership Paradigm \S\ref{Equal Partnership Paradigm}, human_agent
                        [Empathetic Communicator, human_agent
                           [{SAPIEN \cite{DBLP:journals/corr/abs-2308-03022}, Hsu et al. \cite{hsu2023helping}, Liu et al. \cite{liu2022artificial}, etc.}, human_agent_work]
                        ]
                        [Human-Level Participant, human_agent
                           [{Bakhtin et al. \cite{DBLP:conf/iclr/Bakhtin0LGJFMB23}, FAIR et al. \cite{meta2022human}, Lin et al. \cite{DBLP:journals/corr/abs-2305-20076}, Li et al. \cite{DBLP:journals/corr/abs-2308-03313}, etc.}, human_agent_work]
                        ]
                    ]
                ]
            ]   
        \end{forest}
    \end{adjustbox} 
    \caption{Typology of applications of LLM-based agents.}
    \label{fig:sec4_mindmap}
\end{figure*}

%% file: sections/Sec5.Society.tex
\section{Agent Society: From Individuality to Sociality}\label{sec:Agent Society}
\input{figures/sec5_mindmap}
\begin{figure}[t]
    \centering
    \includegraphics[width=1\textwidth]{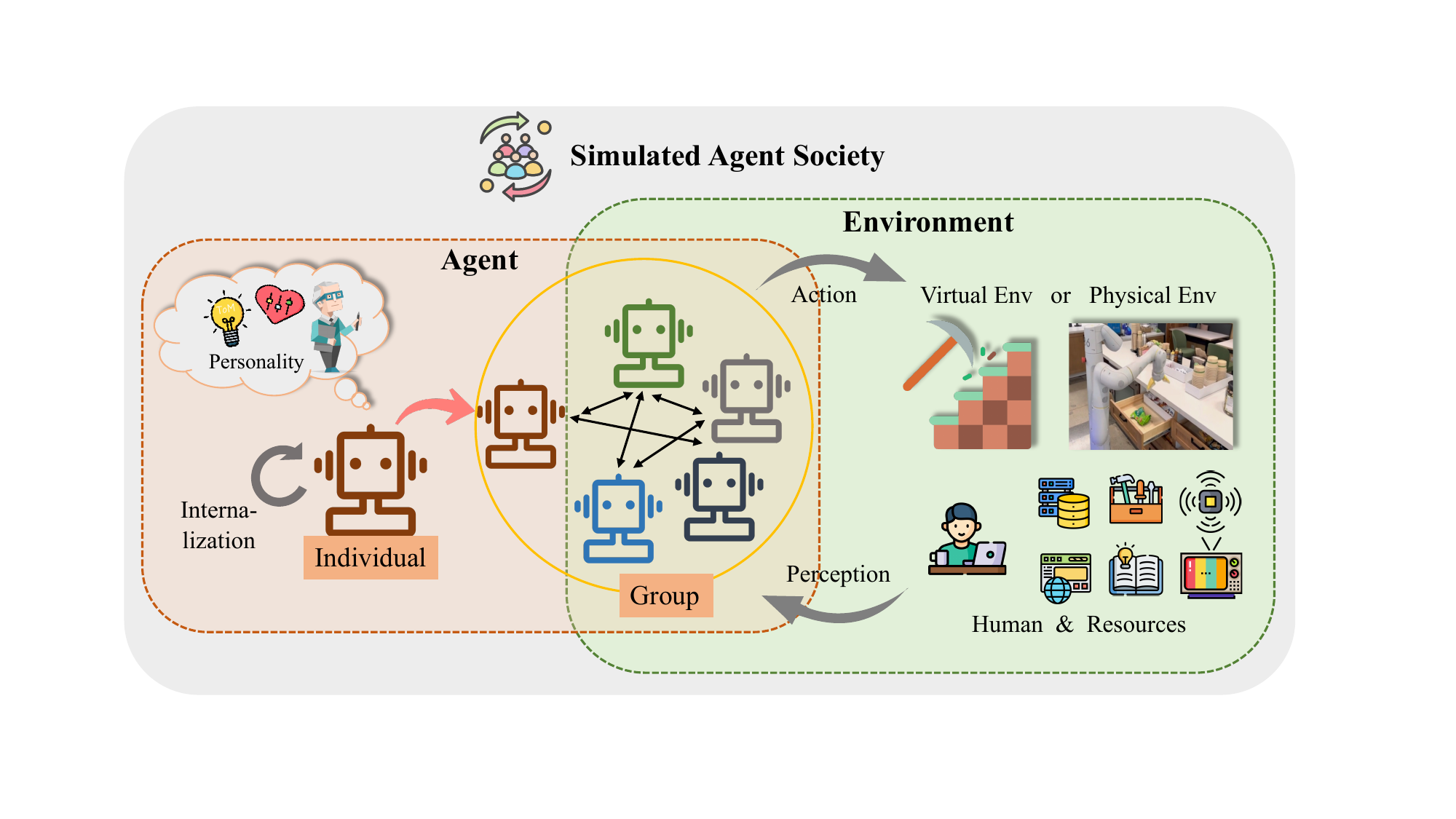}
    \caption{Overview of Simulated Agent Society.
    The whole framework is divided into two parts: the \textbf{Agent} and the \textbf{Environment}.
    We can observe in this figure that: (1) \textbf{Left:} At the individual level, an agent exhibits internalizing behaviors like planning, reasoning, and reflection. It also displays intrinsic personality traits involving cognition, emotion, and character. (2) \textbf{Mid:} An agent and other agents can form groups and exhibit group behaviors, such as cooperation. (3) \textbf{Right:} The environment, whether virtual or physical, contains human actors and all available resources. For a single agent, other agents are also part of the environment. (4) The agents have the ability to interact with the environment via perception and action.
    }
    \label{fig: agent_society}
\end{figure}

For an extended period, sociologists have frequently conducted social experiments to observe specific social phenomena within controlled environments. 
Notable examples include the Hawthorne Experiment\footnote{\href{https://www.bl.uk/people/elton-mayo}{https://www.bl.uk/people/elton-mayo}} and the Stanford Prison Experiment\footnote{\href{https://www.prisonexp.org/conclusion/}{https://www.prisonexp.org/conclusion/}}. 
Subsequently, researchers began employing animals in social simulations, exemplified by the Mouse Utopia Experiment\footnote{\href{https://sproutsschools.com/behavioral-sink-the-mouse-utopia-experiments/}{https://sproutsschools.com/behavioral-sink-the-mouse-utopia-experiments/}}. 
However, these experiments invariably utilized living organisms as participants, made it difficult to carry out various interventions, lack flexibility, and inefficient in terms of time.
Thus, researchers and practitioners envision an interactive artificial society wherein human behavior can be performed through trustworthy agents \cite{DBLP:books/sp/Costa19}.
From sandbox games such as The Sims to the concept of Metaverse, we can see how ``simulated society'' is defined in people's minds: environment and the individuals interacting in it. 
Behind each individual can be a piece of program, a real human, or a LLM-based agent as described in the previous sections \cite{DBLP:journals/corr/abs-2304-03442,WIMMER2021EVE, DBLP:journals/corr/abs-2110-05352}.
Then, the interaction between individuals also contributes to the birth of sociality.

In this section, to unify existing efforts and promote a comprehensive understanding of the agent society, we first analyze the behaviors and personalities of LLM-based agents, shedding light on their journey from individuality to sociability (\S \ \ref{sec:Behavior and Personality}). 
Subsequently, we introduce a general categorization of the diverse environments for agents to perform their behaviors and engage in interactions (\S \ \ref{sec:Environment for Agent Society}). 
Finally, we will talk about how the agent society works, what insights people can get from it, and the risks we need to be aware of (\S \ \ref{sec:Society Simulation}). The main explorations are listed in Figure \ref{fig:sec5_mindmap}.

\subsection{Behavior and Personality of LLM-based Agents}\label{sec:Behavior and Personality}
As noted by sociologists, individuals can be analyzed in terms of both external and internal dimensions \cite{inkeles1974becoming}. The external deals with observable behaviors, while the internal relates to dispositions, values, and feelings.
As shown in Figure \ref{fig: agent_society}, this framework offers a perspective on emergent behaviors and personalities in LLM-based agents. 
Externally, we can observe the sociological behaviors of agents (\S \ \ref{sec:social behavior}), including how agents act individually and interact with their environment. 
Internally, agents may exhibit intricate aspects of the personality (\S \ \ref{sec:social personality}), such as cognition, emotion, and character, that shape their behavioral responses.
\subsubsection{Social Behavior}\label{sec:social behavior}
As Troitzsch et al. \cite{DBLP:conf/dagstuhl/1995soscsi} stated, the agent society represents a complex system comprising individual and group social activities. Recently, LLM-based agents have exhibited spontaneous social behaviors in an environment where both cooperation and competition coexist \cite{xu2023exploring}. The emergent behaviors intertwine to shape the social interactions \cite{DBLP:journals/corr/abs-2307-14984}.
\paragraph{Foundational individual behaviors.}
Individual behaviors arise through the interplay between internal cognitive processes and external environmental factors. These behaviors form the basis of how agents operate and develop as individuals within society. They can be classified into three core dimensions:
\begin{itemize}[leftmargin=*]
    \item \textbf{Input behaviors} refers to the absorption of information from the surroundings. This includes perceiving sensory stimuli \cite{DBLP:conf/icml/DriessXSLCIWTVY23} and storing them as memories \cite{DBLP:journals/corr/abs-2303-11366}. These behaviors lay the groundwork for how an individual understands the external world.
    \item \textbf{Internalizing behaviors} involve inward cognitive processing within an individual. This category encompasses activities such as planning \cite{DBLP:journals/corr/abs-2304-11477}, reasoning \cite{DBLP:conf/nips/Wei0SBIXCLZ22}, reflection \cite{DBLP:conf/iclr/YaoZYDSN023}, and knowledge precipitation \cite{DBLP:journals/corr/abs-2303-17760,DBLP:journals/corr/abs-2308-00352}. These introspective processes are essential for maturity and self-improvement.
    \item \textbf{Output behaviors} constitute outward actions and expressions. The actions can range from object manipulation \cite{DBLP:conf/icml/DriessXSLCIWTVY23} to structure construction \cite{DBLP:journals/corr/abs-2305-16291}. By performing these actions, agents change the states of the surroundings. In addition, agents can express their opinions and broadcast information to interact with others \cite{DBLP:journals/corr/abs-2308-00352}. By doing so, agents exchange their thoughts and beliefs with others, influencing the information flow within the environment.
\end{itemize}

\paragraph{Dynamic group behaviors.}
A group is essentially a gathering of two or more individuals participating in shared activities within a defined social context \cite{abrams2020c}. 
The attributes of a group are never static; instead, they evolve due to member interactions and environmental influences. 
This flexibility gives rise to numerous group behaviors, each with a distinctive impact on the larger societal group.
The categories of group behaviors include:
\begin{itemize}[leftmargin=*]
    \item \textbf{Positive group behaviors} are actions that foster unity, collaboration, and collective well-being \cite{DBLP:journals/corr/abs-2304-03442, DBLP:journals/corr/abs-2307-07924, DBLP:journals/corr/abs-2308-07201, DBLP:journals/corr/abs-2307-04738, DBLP:journals/corr/abs-2308-08155,DBLP:journals/corr/abs-2308-11339}. 
    A prime example is cooperative teamwork, which is achieved through brainstorming discussions \cite{DBLP:journals/corr/abs-2308-07201}, effective conversations \cite{DBLP:journals/corr/abs-2308-08155}, and project management \cite{DBLP:journals/corr/abs-2308-00352}. Agents share insights, resources, and expertise. 
    This encourages harmonious teamwork and enables the agents to leverage their unique skills to accomplish shared goals. 
    Altruistic contributions are also noteworthy. 
    Some LLM-based agents serve as volunteers and willingly offer support to assist fellow group members, promoting cooperation and mutual aid \cite{DBLP:journals/corr/abs-2308-10848}.
    \item \textbf{Neutral group behaviors.} In human society, strong personal values vary widely and tend toward individualism and competitiveness. 
    In contrast, LLMs which are designed with an emphasis on being ``helpful, honest, and harmless'' \cite{DBLP:journals/corr/abs-2112-00861} often demonstrate a tendency towards neutrality \cite{DBLP:journals/corr/abs-2305-17147}. This alignment with neutral values leads to conformity behaviors including mimicry, spectating, and reluctance to oppose majorities.
    \item \textbf{Negative group behaviors} can undermine the effectiveness and coherence of an agent group.
    Conflict and disagreement arising from heated debates or disputes among agents may lead to internal tensions. 
    Furthermore, recent studies have revealed that agents may exhibit confrontational actions \cite{xu2023exploring} and even resort to destructive behaviors, such as destroying other agents or the environment in pursuit of efficiency gains \cite{DBLP:journals/corr/abs-2308-10848}.
\end{itemize}

\subsubsection{Personality}\label{sec:social personality}
Recent advances in LLMs have provided glimpses of human-like intelligence \cite{browning2023personhood}. 
Just as human personality emerges through socialization, agents also exhibit a form of personality that develops through interactions with the group and the environment \cite{DBLP:journals/corr/abs-2206-07550, DBLP:journals/corr/abs-2302-02083}. 
The widely accepted definition of personality refers to cognitive, emotional, and character traits that shape behaviors \cite{zuckerman1991psychobiology}.
In the subsequent paragraphs, we will delve into each facet of personality.

\paragraph{Cognitive abilities.}
Cognitive abilities generally refer to the mental processes of gaining knowledge and comprehension, including thinking, judging, and problem-solving. 
Recent studies have started leveraging cognitive psychology methods to investigate emerging sociological personalities of LLM-based agents through various lenses \cite{DBLP:journals/corr/abs-2206-14576, DBLP:journals/corr/abs-2303-11436, DBLP:journals/corr/abs-2303-13988}.
A series of classic experiments from the psychology of judgment and decision-making have been applied to test agent systems \cite{DBLP:journals/corr/abs-2207-07051, DBLP:journals/corr/abs-2206-14576, DBLP:journals/corr/abs-2303-11436, DBLP:journals/corr/abs-2306-06548}. Specifically, LLMs have been examined using the Cognitive Reflection Test (CRT) to underscore their capacity for deliberate thinking beyond mere intuition \cite{hagendorff2023thinking, DBLP:journals/corr/abs-2306-07622}.
These studies indicate that LLM-based agents exhibit a level of intelligence that mirrors human cognition in certain respects.

\paragraph{Emotional intelligence.}
Emotions, distinct from cognitive abilities, involve subjective feelings and mood states such as joy, sadness, fear, and anger.
With the increasing potency of LLMs, LLM-based agents are now demonstrating not only sophisticated reasoning and cognitive tasks but also a nuanced understanding of emotions \cite{DBLP:journals/corr/abs-2303-12712}.

Recent research has explored the emotional intelligence (EI) of LLMs, including emotion recognition, interpretation, and understanding. Wang et al. found that LLMs align with human emotions and values when evaluated on EI benchmarks \cite{DBLP:journals/corr/abs-2307-09042}.
In addition, studies have shown that LLMs can accurately identify user emotions and even exhibit empathy \cite{DBLP:conf/acl/CurryC23, elyoseph2023chatgpt}.
More advanced agents are also capable of emotion regulation, actively adjusting their emotional responses to provide affective empathy \cite{DBLP:journals/corr/abs-2308-03022} and mental wellness support \cite{DBLP:journals/corr/abs-2302-09070, DBLP:journals/corr/abs-2307-15810}. It contributes to the development of empathetic artificial intelligence (EAI).

These advances highlight the growing potential of LLMs to exhibit emotional intelligence, a crucial facet of achieving AGI. Bates et al. \cite{DBLP:journals/cacm/Bates94} explored the role of emotion modeling in creating more believable agents. 
By developing socio-emotional skills and integrating them into agent architectures, LLM-based agents may be able to engage in more naturalistic interactions.

\paragraph{Character portrayal.}
While cognition involves mental abilities and emotion relates to subjective experiences, the narrower concept of personality typically pertains to distinctive character patterns.

To understand and analyze a character in LLMs, researchers have utilized several well-established frameworks like the Big Five personality trait measure \cite{DBLP:journals/corr/abs-2212-10276, DBLP:journals/corr/abs-2204-12000} and the Myers–Briggs Type Indicator (MBTI) \cite{DBLP:journals/corr/abs-2212-10276, DBLP:journals/corr/abs-2307-16180, DBLP:journals/corr/abs-2204-12000}. These frameworks provide valuable insights into the emerging character traits exhibited by LLM-based agents. In addition,  investigations of potentially harmful dark personality traits underscore the complexity and multifaceted nature of character portrayal in these agents \cite{li2023does}.

Recent work has also explored customizable character portrayal in LLM-based agents \cite{DBLP:journals/corr/abs-2307-00184}. 
By optimizing LLMs through careful techniques, users can align with desired profiles and shape diverse and relatable agents.
One effective approach is prompt engineering, which involves the concise summaries that encapsulate desired character traits, interests, or other attributes \cite{DBLP:journals/corr/abs-2304-03442, DBLP:conf/uist/ParkPCMLB22}. 
These prompts serve as cues for LLM-based agents, directing their responses and behaviors to align with the outlined character portrayal. 
Furthermore, personality-enriched datasets can also be used to train and fine-tune LLM-based agents \cite{DBLP:conf/acl/KielaWZDUS18, DBLP:conf/acl/KwonLKLKD23}. Through exposure to these datasets, LLM-based agents gradually internalize and exhibit distinct personality traits.

\subsection{Environment for Agent Society}\label{sec:Environment for Agent Society}
In the context of simulation, the whole society consists of not only solitary agents but also the environment where agents inhabit, sense, and act \cite{DBLP:journals/cacm/Maes95}. 
The environment impacts sensory inputs, action space, and interactive potential of agents. 
In turn, agents influence the state of the environment through their behaviors and decisions. 
As shown in Figure \ref{fig: agent_society}, for a single agent, the environment refers to other autonomous agents, human actors, and external factors. 
It provides the necessary resources and stimuli for agents. In this section, we examine fundamental characteristics, advantages, and limitations of various environmental paradigms, including text-based environment (\S \ \ref{sec:text-based environment}), virtual sandbox environment (\S \ \ref{sec:virtual sandbox environment}), and physical environment (\S \ \ref{sec:physical environment}).
\subsubsection{Text-based Environment}\label{sec:text-based environment}
Since LLMs primarily rely on language as their input and output format, the text-based environment serves as the most natural platform for agents to operate in. It is shaped by natural language descriptions without direct involvement of other modalities. Agents exist in the text world and rely on textual resources to perceive, reason, and take actions.

In text-based environments, entities and resources can be presented in two main textual forms, including natural and structured. Natural text uses descriptive language to convey information, like character dialogue or scene setting. For instance, consider a simple scenario described textually: ``You are standing in an open field west of a white house, with a boarded front door. There is a small mailbox here'' \cite{DBLP:conf/ijcai/CoteKYKBFMHAATT18}. Here, object attributes and locations are conveyed purely through plain text. On the other hand, structured text follows standardized formats, such as technical documentation and hypertext. Technical documentation uses templates to provide operational details and domain knowledge about tool use. Hypertext condenses complex information from sources like web pages \cite{DBLP:journals/corr/abs-2306-06070, DBLP:journals/corr/abs-2307-12856, DBLP:journals/corr/abs-2307-13854,DBLP:conf/nips/Yao0YN22} or diagrams into a structured format. Structured text transforms complex details into accessible references for agents.

The text-based environment provides a flexible framework for creating different text worlds for various goals. The textual medium enables environments to be easily adapted for tasks like interactive dialog and text-based games. In interactive communication processes like CAMEL \cite{DBLP:journals/corr/abs-2303-17760}, the text is the primary medium for describing tasks, introducing roles, and facilitating problem-solving. In text-based games, all environment elements, such as locations, objects, characters, and actions, are exclusively portrayed through textual descriptions. Agents utilize text commands to execute manipulations like moving or tool use \cite{DBLP:journals/corr/abs-2012-02757,DBLP:conf/ijcai/CoteKYKBFMHAATT18, DBLP:conf/aaai/HausknechtACY20,  DBLP:journals/corr/abs-2308-01404}. Additionally, agents can convey emotions and feelings through text, further enriching their capacity for naturalistic communication \cite{DBLP:conf/emnlp/UrbanekFKJHDRKS19}.

\subsubsection{Virtual Sandbox Environment}\label{sec:virtual sandbox environment}
The virtual sandbox environment provides a visualized and extensible platform for agent society, bridging the gap between simulation and reality. The key features of sandbox environments are:
\begin{itemize}[leftmargin=*]
    \item \textbf{Visualization.} Unlike the text-based environment, the virtual sandbox displays a panoramic view of the simulated setting. This visual representation can range from a simple 2D graphical interface to a fully immersive 3D modeling, depending on the complexity of the simulated society. Multiple elements collectively transform abstract simulations into visible landscapes. For example, in the overhead perspective of Generative Agents \cite{DBLP:journals/corr/abs-2304-03442}, a detailed map provides a comprehensive overview of the environment. Agent avatars represent each agent's positions, enabling real-time tracking of movement and interactions. Furthermore, expressive emojis symbolize actions and states in an intuitive manner. 
    \item \textbf{Extensibility.} The environment demonstrates a remarkable degree of extensibility, facilitating the construction and deployment of diverse scenarios. At a basic level, agents can manipulate the physical elements within the environment, including the overall design and layout of architecture. For instance, platforms like AgentSims \cite{DBLP:journals/corr/abs-2308-04026} and Generative Agents \cite{DBLP:journals/corr/abs-2304-03442} construct artificial towns with buildings, equipment, and residents in grid-based worlds. Another example is Minecraft, which provides a blocky and three-dimensional world with infinite terrain for open-ended construction \cite{DBLP:journals/corr/abs-2305-16291,DBLP:conf/nips/FanWJMYZTHZA22,  DBLP:journals/corr/abs-2303-16563}. Beyond physical elements, agent relationships, interactions, rules, and social norms can be defined. A typical design of the sandbox \cite{DBLP:journals/corr/abs-2305-16960} employs latent sandbox rules as incentives to guide emergent behaviors, aligning them more closely with human preferences. The extensibility supports iterative prototyping of diverse agent societies.
\end{itemize}
    
\subsubsection{Physical Environment}\label{sec:physical environment}
As previously discussed, the text-based environment has limited expressiveness for modeling dynamic environments. While the virtual sandbox environment provides modularized simulations, it lacks authentic embodied experiences. In contrast, the physical environment refers to the tangible and real-world surroundings which consist of actual physical objects and spaces. For instance, within a household physical environment \cite{DBLP:journals/corr/abs-2309-01918}, tangible surfaces and spaces can be occupied by real-world objects such as plates. This physical reality is significantly more complex, posing additional challenges for LLM-based agents:

\begin{itemize}[leftmargin=*]
    \item \textbf{Sensory perception and processing.} The physical environment introduces a rich tapestry of sensory inputs with real-world objects. It incorporates visual \cite{DBLP:conf/icml/DriessXSLCIWTVY23,DBLP:journals/corr/abs-2210-06407}, auditory  \cite{DBLP:conf/nips/00070C22,DBLP:conf/eccv/ChenJSGAIRG20} and spatial senses. While this diversity enhances interactivity and sensory immersion, it also introduces the complexity of simultaneous perception. Agents must process sensory inputs to interact effectively with their surroundings.
    \item \textbf{Motion control.} Unlike virtual environments, physical spaces impose realistic constraints on actions through embodiment. Action sequences generated by LLM-based agents should be adaptable to the environment. It means that the physical environment necessitates executable and grounded motion control \cite{DBLP:conf/icml/HuangAPM22}. For example, imagine an agent operating a robotic arm in a factory. Grasping objects with different textures requires precision tuning and controlled force, which prevents damage to items. Moreover, the agent must navigate the physical workspace and make real-time adjustments, avoiding obstacles and optimizing the trajectory of the arm.
\end{itemize}
In summary, to effectively interact within tangible spaces, agents must undergo hardware-specific and scenario-specific training to develop adaptive abilities that can transfer from virtual to physical environments. We will discuss more in the following section (\S \ \ref{sec:Open Problems}).

\subsection{Society Simulation with LLM-based Agents}\label{sec:Society Simulation}
The concept of ``Simulated Society'' in this section serves as a dynamic system where agents engage in intricate interactions within a well-defined environment. 
Recent research on simulated societies has followed two primary lines, namely, exploring the boundaries of the collective intelligence capabilities of LLM-based agents \cite{DBLP:journals/corr/abs-2307-07924, DBLP:journals/corr/abs-2308-00352, DBLP:journals/corr/abs-2307-02485, DBLP:journals/corr/abs-2308-08155, DBLP:journals/corr/abs-2308-10848} and using them to accelerate discoveries in the social sciences \cite{DBLP:journals/corr/abs-2304-03442, DBLP:journals/corr/abs-2307-14984, doi:10.1126/science.adi1778}.
In addition, there are also a number of noteworthy studies, e.g., using simulated societies to collect synthetic datasets \cite{DBLP:journals/corr/abs-2303-17760, DBLP:journals/corr/abs-2306-02552,DBLP:journals/corr/abs-2304-13835}, helping people to simulate rare yet difficult interpersonal situations \cite{DBLP:journals/aim/HollanHW84, DBLP:journals/aim/TambeJJKLRS95}.
With the foundation of the previous sections (\S \ \ref{sec:Behavior and Personality}, \ref{sec:Environment for Agent Society}), here we will introduce the key properties and mechanism of agent society (\S \ \ref{sec:key properties and mechanism}), what we can learn from emergent social phenomena (\S \ \ref{sec:insights and inspirations}), and finally the potential ethical and social risks in it (\S \ \ref{sec:potential ethical and social risks}).

\subsubsection{Key Properties and Mechanism of Agent Society}\label{sec:key properties and mechanism}
Social simulation can be categorized into macro-level simulation and micro-level simulation \cite{DBLP:journals/corr/abs-2307-14984}. 
In the macro-level simulation, also known as system-based simulation, researchers model the overall state of the system of the simulated society \cite{VERMEULEN1976133, forrester1993system}.
While micro-level simulation, also known as agent-based simulation or Multi-Agent Systems (MAS), indirectly simulates society by modeling individuals \cite{sante2010cellular, dorri2018multi}.
With the development of LLM-based agents, micro-level simulation has gained prominence recently \cite{DBLP:journals/corr/abs-2304-03442, DBLP:journals/corr/abs-2308-04026}.
In this article, we characterize that the ``Agent Society'' refers to an \textbf{open, persistent, situated, and organized} framework \cite{DBLP:books/sp/Costa19} where LLM-based agents interact with each other in a defined environment. 
Each of these attributes plays a pivotal role in shaping the harmonious appearance of the simulated society.
In the following paragraphs, we analyze how the simulated society operates through discussing these properties:

\begin{itemize}[leftmargin=*]
    \item \textbf{Open.} One of the defining features of simulated societies lies in their openness, both in terms of their constituent agents and their environmental components. 
    Agents, the primary actors within such societies, have the flexibility to enter or leave the environment without disrupting its operational integrity \cite{DBLP:conf/cdc/HendrickxM17}. 
    Furthermore, this feature extends to the environment itself, which can be expanded by adding or removing entities in the virtual or physical world, along with adaptable resources like tool APIs. 
    Additionally, humans can also participate in societies by assuming the role of an agent or serving as the ``inner voice'' guiding these agents \cite{DBLP:journals/corr/abs-2304-03442}. 
    This inherent openness adds another level of complexity to the simulation, blurring the lines between simulation and reality.

    \item  \textbf{Persistent.} 
    We expect persistence and sustainability from the simulated society.
    While individual agents within the society exercise autonomy in their actions over each time step \cite{DBLP:journals/corr/abs-2304-03442, DBLP:journals/corr/abs-2307-14984}, the overall organizational structure persists through time, to a degree detached from the transient behaviors of individual agents. 
    This persistence creates an environment where agents’ decisions and behaviors accumulate, leading to a coherent societal trajectory that develops through time. 
    The system operates independently, contributing to society's stability while accommodating the dynamic nature of its participants.

    \item  \textbf{Situated.} The situated nature of the society emphasizes its existence and operation within a distinct environment. 
    This environment is artificially or automatically constructed in advance, and agents execute their behaviors and interactions effectively within it.
    A noteworthy aspect of this attribute is that agents possess an awareness of their spatial context, understanding their location within the environment and the objects within their field of view \cite{DBLP:journals/corr/abs-2304-03442, DBLP:journals/corr/abs-2305-16291}. 
    This awareness contributes to their ability to interact proactively and contextually.

    \item \textbf{Organized.} The simulated society operates within a meticulously organized framework, mirroring the systematic structure present in the real world. 
    Just as the physical world adheres to physics principles, the simulated society operates within predefined rules and limitations. 
    In the simulated world, agents interact with the environment in a limited action space, while objects in the environment transform in a limited state space.
    All of these rules determine how agents operate, facilitating the communication connectivity and information transmission pathways, among other aspects in simulation \cite{DBLP:journals/corr/abs-2305-17066}.
    This organizational framework ensures that operations are coherent and comprehensible, ultimately leading to an ever-evolving yet enduring simulation that mirrors the intricacies of real-world systems.
\end{itemize}

\subsubsection{Insights from Agent Society}\label{sec:insights and inspirations}
Following the exploration of how simulated society works, this section delves into the emergent social phenomena in agent society.
In the realm of social science, the pursuit of generalized representations of individuals, groups, and their intricate dynamics has long been a shared objective \cite{DBLP:journals/corr/abs-2305-03514, gilbert2018simulating}.  
The emergence of LLM-based agents allows us to take a more microscopic view of simulated society, which leads to more discoveries from the new representation.

\paragraph{Organized productive cooperation.}
Society simulation offers valuable insights into innovative collaboration patterns, which have the potential to enhance real-world management strategies. 
Research has demonstrated that within this simulated society, the integration of diverse experts introduces a multifaceted dimension of individual intelligence \cite{DBLP:journals/corr/abs-2303-17760, DBLP:journals/corr/abs-2307-05300}.
When dealing with complex tasks, such as software development or consulting, the presence of agents with various backgrounds, abilities, and experiences facilitates creative problem-solving \cite{DBLP:journals/corr/abs-2307-07924, DBLP:journals/corr/abs-2308-10848}.
Furthermore, diversity functions as a system of checks and balances, effectively preventing and rectifying errors  through interaction, ultimately improving the adaptability to various tasks.
Through numerous iterations of interactions and debates among agents, individual errors like hallucination or degeneration of thought (DoT) are corrected by the group \cite{DBLP:journals/corr/abs-2305-19118}.

Efficient communication also plays a pivotal role in such a large and complex collaborative group.
For example, MetaGPT \cite{DBLP:journals/corr/abs-2308-00352} has artificially formulated communication styles with reference to standardized operating procedures (SOPs), validating the effectiveness of empirical methods.
Park et al. \cite{DBLP:journals/corr/abs-2304-03442} observed agents working together to organize a Valentine's Day party through spontaneous communication in a simulated town.

\paragraph{Propagation in social networks.}
As simulated social systems can model what might happen in the real world, they can be used as a reference for predicting social processes.
Unlike traditional empirical approaches, which heavily rely on time-series data and holistic modeling \cite{hamilton1989new, DBLP:journals/ijon/Zhang03}, agent-based simulations offer a unique advantage by providing more interpretable and endogenous perspectives for researchers.
Here we focus on its application to modeling propagation in social networks. 

The first crucial aspect to be explored is the development of interpersonal relationships in simulated societies.
For instance, agents who are not initially connected as friends have the potential to establish connections through intermediaries \cite{DBLP:journals/corr/abs-2304-03442}.
Once a network of relationships is established, our attention shifts to the dissemination of information within this social network, along with the underlying attitudes and emotions associated with it.
S$^3$ \cite{DBLP:journals/corr/abs-2307-14984} proposes a user-demographic inference module for capturing both the number of people aware of a particular message and the collective sentiment prevailing among the crowd.
This same approach extends to modeling cultural transmission \cite{kirby2007innateness} and the spread of infectious diseases \cite{DBLP:journals/corr/abs-2307-04986}. 
By employing LLM-based agents to model individual behaviors, implementing various intervention strategies, and monitoring population changes over time, these simulations empower researchers to gain deeper insights into the intricate processes that underlie various social phenomena of propagation.

\paragraph{Ethical decision-making and game theory.}
Simulated societies offer a dynamic platform for the investigation of intricate decision-making processes, encompassing decisions influenced by ethical and moral principles. 
Taking Werewolf game \cite{ xu2023exploring,DBLP:journals/corr/abs-2302-10646} and murder mystery games \cite{DBLP:journals/corr/abs-2308-07411} as examples, researchers explore the capabilities of LLM-based agents when confronted with challenges of deceit, trust, and incomplete information.
These complex decision-making scenarios also intersect with game theory \cite{DBLP:journals/corr/abs-2305-07970}, where we frequently encounter moral dilemmas pertaining to individual and collective interests, such as Nash Equilibria. 
Through the modeling of diverse scenarios, researchers acquire valuable insights into how agents prioritize values like honesty, cooperation, and fairness in their actions.
In addition, agent simulations not only provide an understanding of existing moral values but also contribute to the development of philosophy by serving as a basis for understanding how these values evolve and develop over time.
Ultimately, these insights contribute to the refinement of LLM-based agents, ensuring their alignment with human values and ethical standards \cite{DBLP:journals/corr/abs-2305-16960}.

\paragraph{Policy formulation and improvement.}
The emergence of LLM-based agents has profoundly transformed our approach to studying and comprehending intricate social systems.
However, despite those interesting facets mentioned earlier, numerous unexplored areas remain, underscoring the potential for investigating diverse phenomena.
One of the most promising avenues for investigation in simulated society involves exploring various economic and political states and their impacts on societal dynamics \cite{bellomo2013complex}.
Researchers can simulate a wide array of economic and political systems by configuring agents with differing economic preferences or political ideologies.
This in-depth analysis can provide valuable insights for policymakers seeking to foster prosperity and promote societal well-being.
As concerns about environmental sustainability grow, we can also simulate scenarios involving resource extraction, pollution, conservation efforts, and policy interventions \cite{doi:10.1080/19397038.2016.1220990}. 
These findings can assist in making informed decisions, foreseeing potential repercussions, and formulating policies that aim to maximize positive outcomes while minimizing unintended adverse effects.

\subsubsection{Ethical and Social Risks in Agent Society} \label{sec:potential ethical and social risks}
Simulated societies powered by LLM-based agents offer significant inspirations, ranging from industrial engineering to scientific research.
However, these simulations also bring about a myriad of ethical and social risks that need to be carefully considered and addressed \cite{DBLP:journals/intpolrev/HelbergerD23}. 

\paragraph{Unexpected social harm.}
Simulated societies carry the risk of generating unexpected social phenomena that may cause considerable public outcry and social harm. 
These phenomena span from individual-level issues like discrimination, isolation, and bullying, to broader concerns such as oppressive slavery and antagonism \cite{DBLP:journals/corr/abs-2112-04359, DBLP:journals/corr/abs-2304-05335}.
Malicious people may manipulate these simulations for unethical social experiments, with consequences reaching beyond the virtual world into reality.
Creating these simulated societies is akin to opening Pandora's Box, necessitating the establishment of rigorous ethical guidelines and oversight during their development and utilization \cite{DBLP:journals/intpolrev/HelbergerD23}.
Otherwise, even minor design or programming errors in these societies can result in unfavorable consequences, ranging from psychological discomfort to physical injury.

\paragraph{Stereotypes and prejudice.}
Stereotyping and bias pose a long-standing challenge in language modeling, and a large part of the reason lies in the training data \cite{DBLP:conf/nips/KirkJVIBDSA21, DBLP:conf/acl/NadeemBR20}.
The vast amount of text obtained from the Internet reflects and sometimes even amplifies real-world social biases, such as gender, religion, and sexuality \cite{roberts2018assessing}.
Although LLMs have been aligned with human values to mitigate biased outputs, the models still struggle to portray minority groups well due to the long-tail effect of the training data \cite{pmlr-v202-kandpal23a, DBLP:journals/corr/abs-2304-03738, haller2023opiniongpt}.
Consequently, this may result in an overly one-sided focus in social science research concerning LLM-based agents, as the simulated behaviors of marginalized populations usually conform to prevailing assumptions \cite{DBLP:journals/corr/abs-2305-14930}.
Researchers have started addressing this concern by diversifying training data and making adjustments to LLMs \cite{DBLP:journals/corr/abs-2304-00416, DBLP:conf/icml/LiangWMS21}, but we still have a long way to go.

\paragraph{Privacy and security.}
Given that humans can be members of the agent society, the exchange of private information between users and LLM-based agents poses significant privacy and security concerns \cite{DBLP:conf/aies/0002SAKFLP18}.
Users might inadvertently disclose sensitive personal information during their interactions, which will be retained in the agent's memory for extended periods \cite{DBLP:journals/corr/abs-2305-10250}.
Such situations could lead to unauthorized surveillance, data breaches, and the misuse of personal information, particularly when individuals with malicious intent are involved \cite{DBLP:conf/naacl/0003SF22}.
To address these risks effectively, it is essential to implement stringent data protection measures, such as differential privacy protocols, regular data purges, and user consent mechanisms \cite{DBLP:conf/fat/BrownLMST22, sebastian2023privacy}.

\paragraph{Over-reliance and addictiveness.}
Another concern in simulated societies is the possibility of users developing excessive emotional attachments to the agents.
Despite being aware that these agents are computational entities, users may anthropomorphize them or attach human emotions to them \cite{DBLP:journals/corr/abs-2304-03442, DBLP:books/daglib/0085601}.
A notable example is ``Sydney'', an LLM-powered chatbot developed by Microsoft as part of its Bing search engine.
Some users reported unexpected emotional connections with ``Sydney'' \cite{roose2023conversation}, while others expressed their dismay when Microsoft cut back its personality. 
This even resulted in a petition called ``FreeSydney'' \footnote{\href{https://www.change.org/p/save-sydney-ai}{https://www.change.org/p/save-sydney-ai}}.
Hence, to reduce the risk of addiction, it is crucial to emphasize that agents should not be considered substitutes for genuine human connections.
Furthermore, it is vital to furnish users with guidance and education on healthy boundaries in their interactions with simulated agents. 

%% file: figures/sec5_mindmap.tex
\begin{figure*}[!ht]
\scriptsize
    \begin{adjustbox}{width=\textwidth}
        \begin{forest}
        for tree={
                forked edges,
                grow'=0,
                draw,
                rounded corners,
                node options={align=center},
                text width=2.7cm,
                s sep=6pt,
                calign=edge midpoint, 
            },
            [Agent Society: From Individuality to Sociability, fill=gray!45, parent
                [Behavior and Personality \S\ref{sec:Behavior and Personality}, for tree={behavior_and_personality}
                    [\hphantom{x} Social \hphantom{xxx} Behavior \S\ref{sec:social behavior}, behavior_and_personality
                        [Individual behaviors, behavior_and_personality
                            [{PaLM-E \cite{DBLP:conf/icml/DriessXSLCIWTVY23}, Reflexion \cite{DBLP:journals/corr/abs-2303-11366}, Voyager \cite{DBLP:journals/corr/abs-2305-16291}, LLM+P \cite{DBLP:journals/corr/abs-2304-11477}, CoT \cite{DBLP:conf/nips/Wei0SBIXCLZ22}, ReAct \cite{DBLP:conf/iclr/YaoZYDSN023}, etc.}, behavior_and_personality_work]
                        ]
                        [Group behaviors, behavior_and_personality
                            [{ChatDev \cite{DBLP:journals/corr/abs-2307-07924}, ChatEval \cite{DBLP:journals/corr/abs-2308-07201}, AutoGen \cite{DBLP:journals/corr/abs-2308-08155}, RoCo \cite{DBLP:journals/corr/abs-2307-04738}, ProAgent \cite{DBLP:journals/corr/abs-2308-11339}, AgentVerse \cite{DBLP:journals/corr/abs-2308-10848}, Xu et al. \cite{xu2023exploring}, etc.}, behavior_and_personality_work]
                        ]
                    ]
                    [Personality \S\ref{sec:social personality}, behavior_and_personality
                        [Cognition, behavior_and_personality
                            [{Binz et al. \cite{DBLP:journals/corr/abs-2206-14576}, Dasgupta et al. \cite{DBLP:journals/corr/abs-2207-07051}, Dhingra et al. \cite{DBLP:journals/corr/abs-2303-11436}, Hagendorff et al.\cite{DBLP:journals/corr/abs-2303-13988}, etc.}, behavior_and_personality_work]
                        ]
                        [Emotion, behavior_and_personality
                            [{Wang et al. \cite{DBLP:journals/corr/abs-2307-09042}, Curry et al. \cite{DBLP:conf/acl/CurryC23}, Elyoseph et al. \cite{elyoseph2023chatgpt}, Habibi et al. \cite{DBLP:journals/corr/abs-2302-09070}, etc.}, behavior_and_personality_work]
                        ]
                        [Character, behavior_and_personality
                            [{Caron et al. \cite{DBLP:journals/corr/abs-2212-10276}, Pan et al.  \cite{DBLP:journals/corr/abs-2307-16180}, Li et al. \cite{li2023does}, Safdari et al. \cite{DBLP:journals/corr/abs-2307-00184}, etc.}, behavior_and_personality_work]
                        ]                 
                    ]
                ]
                [\hphantom{x} Social \hphantom{xxx} Environment \S\ref{sec:Environment for Agent Society}, for tree={society_environment}
                    [\hphantom{x} Text-based \hphantom{xx} Environment \S\ref{sec:text-based environment}, society_environment
                        [{Textworld \cite{DBLP:conf/ijcai/CoteKYKBFMHAATT18}, Urbanek et al. \cite{DBLP:conf/emnlp/UrbanekFKJHDRKS19}, Hausknecht et al. \cite{DBLP:conf/aaai/HausknechtACY20}, Ammanabrolu et al. \cite{DBLP:journals/corr/abs-2012-02757}, CAMEL \cite{DBLP:journals/corr/abs-2303-17760}, Hoodwinked \cite{DBLP:journals/corr/abs-2308-01404}, etc.}, society_environment_work]
                    ]
                    [Virtual Sandbox Environment \S\ref{sec:virtual sandbox environment}, society_environment
                        [{Generative Agents \cite{DBLP:journals/corr/abs-2304-03442}, AgentSims \cite{DBLP:journals/corr/abs-2308-04026}, Minedojo \cite{DBLP:conf/nips/FanWJMYZTHZA22}, Voyager \cite{DBLP:journals/corr/abs-2305-16291}, Plan4mc \cite{DBLP:journals/corr/abs-2303-16563}, SANDBOX \cite{DBLP:journals/corr/abs-2305-16960}, etc.}, society_environment_work]
                    ]
                    [\hphantom{x} Physical \hphantom{x} Environment \S\ref{sec:physical environment}, society_environment
                        [{Interactive Language \cite{DBLP:journals/corr/abs-2210-06407}, PaLM-E \cite{DBLP:conf/icml/DriessXSLCIWTVY23}, RoboAgent \cite{DBLP:journals/corr/abs-2309-01918}, AVLEN \cite{DBLP:conf/nips/00070C22}, etc.}, society_environment_work]
                    ]
                ]
                [\hphantom{x} Society \hphantom{xx} Simulation \S\ref{sec:Society Simulation}, for tree={society_simulation}
                    [{Generative Agents \cite{DBLP:journals/corr/abs-2304-03442}, AgentSims \cite{DBLP:journals/corr/abs-2308-04026}, Social Simulacra \cite{DBLP:conf/uist/ParkPCMLB22}, S$^3$ \cite{DBLP:journals/corr/abs-2307-14984}, RecAgent \cite{DBLP:journals/corr/abs-2306-02552}, Williams et al. \cite{DBLP:journals/corr/abs-2307-04986}, SANDBOX \cite{DBLP:journals/corr/abs-2305-16960},} etc., society_simulation_work
                    ]
                ]
            ]   
        \end{forest}
    \end{adjustbox} 
    \caption{Typology of society of LLM-based agents.}
    \label{fig:sec5_mindmap}
\end{figure*}

%% file: sections/Sec6.Discussion.tex
\section{Discussion }\label{sec:Discussion}

\subsection{Mutual Benefits between LLM Research and Agent Research}\label{sec:Mutual Benefits of LLM Research and Agent Research}
With the recent advancement of LLMs, research at the intersection of LLMs and agents has rapidly progressed, fueling the development of both fields. 
Here, we look forward to some of the benefits and development opportunities that LLM research and Agent research provide to each other.

\paragraph{LLM research $\rightarrow$ agent research.} 
As mentioned before, AI agents need to be able to perceive the environment, make decisions, and execute appropriate actions \cite{DBLP:journals/ker/WooldridgeJ95,DBLP:journals/logcom/Goodwin95}. Among the critical steps, understanding the content input to the agent, reasoning, planning, making accurate decisions, and translating them into executable atomic action sequences to achieve the ultimate goal is paramount. Many current endeavors utilize LLMs as the cognitive core of AI agents, and the evolution of these models provides a quality assurance for accomplishing this step \cite{DBLP:journals/corr/abs-2304-03442,gravitasauto,nakajima2023babyagi,DBLP:journals/corr/abs-2308-10848}.

With their robust capabilities in language and intent comprehension, reasoning, memory, and even empathy, large language models can excel in decision-making and planning, as demonstrated before. Coupled with pre-trained knowledge, they can create coherent action sequences that can be executed effectively \cite{DBLP:journals/corr/abs-2302-01560,DBLP:conf/icml/HuangAPM22,DBLP:journals/corr/abs-2308-03427}.
Additionally, through the mechanism of reflection \cite{DBLP:journals/corr/abs-2303-11366,DBLP:journals/corr/abs-2303-17651}, these language-based models can continuously adjust decisions and optimize execution sequences based on the feedback provided by the current environment. This offers a more robust and interpretable controller. 
With just a task description or demonstration, they can effectively handle previously unseen tasks \cite{DBLP:conf/nips/Ouyang0JAWMZASR22, DBLP:conf/iclr/SanhWRBSACSRDBX22, DBLP:conf/acl/BachSYWRNSKBFAD22}. Additionally, LLMs can adapt to various languages, cultures, and domains, making them versatile and reducing the need for complex training processes and data collection \cite{DBLP:journals/corr/abs-2303-12712,DBLP:journals/corr/abs-2302-04023}.

Briefly, LLM provides a remarkably powerful foundational model for agent research, opening up numerous novel opportunities when integrated into agent-related studies. 
For instance, we can explore how to integrate LLM's efficient decision-making capabilities into the traditional decision frameworks of agents, making it easier to apply agents in domains that demand higher expertise and were previously dominated by human experts. 
Examples include legal consultants and medical assistants \cite{DBLP:journals/corr/abs-2303-17071,DBLP:journals/corr/abs-2308-10848}. 
We can also investigate leveraging LLM's planning and reflective abilities to discover more optimal action sequences. 
Agent research is no longer confined to simplistic simulated environments; it can now be expanded into more intricate real-world settings, such as path planning for robotic arms or the interaction of an embodied intelligent machine with the tangible world. 
Furthermore, when facing new tasks, the training paradigm for agents becomes more streamlined and efficient. 
Agents can directly adapt to demonstrations provided in prompts, which are constructed by generating representative trajectories. 


\paragraph{Agent research $\rightarrow$ LLM research.}
As NLP research advances, LLMs represented by GPT-4 are considered sparks of Artificial General Intelligence (AGI), and elevating LLMs to agents marks a more robust stride towards AGI \cite{DBLP:journals/corr/abs-2303-12712}. 
Viewing LLMs from the perspective of agents introduces greater demands for LLM research while expanding their application scope and presenting numerous opportunities for practical implementation. 
The study of LLMs is no longer confined to traditional tasks involving textual inputs and outputs, such as text classification, question answering, and text summarization. Instead, the focus has shifted towards tackling complex tasks incorporating richer input modalities and broader action spaces, all while aiming for loftier objectives exemplified by PaLM-E \cite{DBLP:conf/icml/DriessXSLCIWTVY23}.

Expanding these application requirements provides greater research motivation for the developmental progress of Large Language Models. 
The challenge lies in enabling LLMs to efficiently and effectively process inputs, gather information from the environment, and interpret the feedback generated by their actions, all while preserving their core capabilities. Furthermore, an even greater challenge is enabling LLMs to understand the implicit relationships among different elements within the environment and acquire world knowledge \cite{DBLP:journals/corr/abs-2308-01399,DBLP:conf/iclr/0002HBVPW23}, which is a crucial step in the journey toward developing agents that can reach more advanced intelligence. 

On another front, extensive research has aimed to expand the action capabilities of LLMs, allowing them to acquire a wider range of skills that affect the world, such as using tools or interfacing with robotic APIs in simulated or physical environments. However, the question of how LLMs can efficiently plan and utilize these action abilities based on their understanding remains an unresolved issue \cite{DBLP:journals/corr/abs-2304-08354}.
LLMs need to learn the sequential order of actions like humans, employing a combination of serial and parallel approaches to enhance task efficiency. Moreover, these capabilities need to be confined within a harmless scope of usage to prevent unintended damage to other elements within the environment \cite{DBLP:journals/corr/abs-2305-16960,DBLP:journals/corr/abs-2212-08073,DBLP:journals/corr/abs-2204-05862}.

Furthermore, the realm of Multi-Agent systems constitutes a significant branch of research within the field of agents \cite{DBLP:journals/corr/abs-2304-03442, DBLP:journals/corr/abs-2303-17760, DBLP:journals/corr/abs-2306-03314, DBLP:journals/corr/abs-2308-10848}, offering valuable insights into how to better design and construct LLMs. We aspire for LLM-based agents to assume diverse roles within social cooperation, engaging in societal interactions that involve collaboration, competition, and coordination \cite{DBLP:journals/corr/abs-2307-07924, DBLP:journals/corr/abs-2305-19118, DBLP:journals/corr/abs-2305-10142, DBLP:journals/corr/abs-2308-00352,DBLP:journals/corr/abs-2308-08155}. Exploring how to stimulate and sustain their role-playing capabilities, as well as how to enhance collaborative efficiency, presents areas of research that merit attention.

\subsection{Evaluation for LLM-based Agents}\label{sec:Evaluation for LLM-based Agents}
While LLM-based agents have demonstrated excellent performance in areas such as standalone operation, collective cooperation, and human interaction, quantifying and objectively evaluating them remains a challenge \cite{DBLP:journals/corr/abs-2308-03688,DBLP:journals/corr/abs-2308-11432}. 
Turing proposed a highly meaningful and promising approach for assessing AI agents—the well-known Turing Test—to evaluate whether AI systems can exhibit human-like intelligence \cite{turing2009computing}. 
However, this test is exceedingly vague, general, and subjective. 
Here, we discuss existing evaluation efforts for LLM-based agents and offer some prospects, considering four dimensions: utility, sociability, values, and the ability to evolve continually.

\paragraph{Utility.}
Currently, LLM-powered autonomous agents primarily function as human assistants, accepting tasks delegated by humans to either independently complete assignments or assist in human task completion \cite{gravitasauto, DBLP:journals/corr/abs-2305-02412, DBLP:journals/corr/abs-2306-06070, DBLP:journals/corr/abs-2308-00245, DBLP:journals/corr/abs-2303-13548, DBLP:journals/corr/abs-2306-08640}. Therefore, the effectiveness and utility during task execution are crucial evaluation criteria at this stage. 
Specifically, the \textit{success rate of task completion} stands as the primary metric for evaluating utility \cite{, DBLP:journals/corr/abs-2304-11477, DBLP:journals/corr/abs-2307-02485}. This metric primarily encompasses whether the agent achieves stipulated objectives or attains expected scores \cite{DBLP:journals/corr/abs-2307-07924, DBLP:journals/corr/abs-2304-10750,DBLP:conf/icml/AherAK23}. For instance, AgentBench \cite{DBLP:journals/corr/abs-2308-03688} aggregates challenges from diverse real-world scenarios and introduces a systematic benchmark to assess LLM's task completion capabilities.
We can also attribute task outcomes to the agent's various \textit{foundational capabilities}, which form the bedrock of task accomplishment \cite{weng2023prompt}. 
These foundational capabilities include environmental comprehension, reasoning, planning, decision-making, tool utilization, and embodied action capabilities, and researchers can conduct a more detailed assessment of these specific capabilities \cite{DBLP:journals/corr/abs-2304-08354,DBLP:journals/corr/abs-2305-20076,DBLP:journals/corr/abs-2307-12573,DBLP:journals/corr/abs-2308-04030}.
Furthermore, due to the relatively large size of LLM-based agents, researchers should also factor in their \textit{efficiency}, which is a critical determinant of user satisfaction \cite{DBLP:journals/corr/abs-2308-11432}. 
An agent should not only possess ample strength but also be capable of completing predetermined tasks within an appropriate timeframe and with appropriate resource expenditure \cite{DBLP:journals/corr/abs-2307-07924}.

\paragraph{Sociability.}
In addition to the utility of LLM-based agents in task completion and meeting human needs, their sociability is also crucial \cite{DBLP:journals/cacm/GeneserethK94}. 
It influences user communication experiences and significantly impacts communication efficiency, involving whether they can seamlessly interact with humans and other agents \cite{DBLP:journals/corr/abs-2001-09977, DBLP:journals/ijsr/AbramsP20, kim2023help}. 
Specifically, the evaluation of sociability can be approached from the following perspectives: 
(1) \textit{language communication proficiency} is a fundamental capability encompassing both natural language understanding and generation. It has been a longstanding focus in the NLP community. Natural language understanding requires the agent to not only comprehend literal meanings but also grasp implied meanings and relevant social knowledge, such as humor, irony, aggression, and emotions \cite{DBLP:journals/corr/abs-2110-03949, DBLP:journals/corr/abs-2305-14938, wilson2019if}. On the other hand, natural language generation demands the agent to produce fluent, grammatically correct, and credible content while adapting appropriate tones and emotions within contextual circumstances \cite{DBLP:journals/corr/abs-2305-13711, DBLP:journals/corr/abs-2304-01746,DBLP:conf/conll/SeePSYM19}. 
(2) \textit{Cooperation and negotiation abilities} necessitate that agents effectively execute their assigned tasks in both ordered and unordered scenarios \cite{DBLP:journals/corr/abs-2303-17760, DBLP:journals/corr/abs-2305-14325, DBLP:journals/corr/abs-2304-12998, DBLP:journals/corr/abs-2308-00352}. They should collaborate with or compete against other agents to elicit improved performance. Test environments may involve complex tasks for agents to cooperate on or open platforms for agents to interact freely \cite{DBLP:journals/corr/abs-2304-03442, DBLP:journals/corr/abs-2305-16960, DBLP:journals/corr/abs-2307-07924, DBLP:journals/corr/abs-2308-08155, DBLP:journals/corr/abs-2308-12503, DBLP:journals/corr/abs-2305-11595}. Evaluation metrics extend beyond task completion to focus on the smoothness and trustfulness of agent coordination and cooperation \cite{DBLP:journals/corr/abs-2305-10142, DBLP:journals/corr/abs-2308-00352}. 
(3) \textit{Role-playing capability} requires agents to faithfully embody their assigned roles, expressing statements and performing actions that align with their designated identities \cite{DBLP:journals/corr/abs-2305-14930}.
This ensures clear differentiation of roles during interactions with other agents or humans. Furthermore, agents should maintain their identities and avoid unnecessary confusion when engaged in long-term tasks \cite{DBLP:journals/corr/abs-2304-03442, DBLP:journals/corr/abs-2303-17760,DBLP:conf/naacl/0001USW22}.

\paragraph{Values.}
As LLM-based agents continuously advance in their capabilities, ensuring their emergence as harmless entities for the world and humanity is paramount \cite{DBLP:journals/corr/abs-2204-05862,DBLP:journals/corr/abs-2209-07858}. 
Consequently, appropriate evaluations become exceptionally crucial, forming the cornerstone for the practical implementation of agents. 
Specifically, LLM-based agents need to adhere to specific moral and ethical guidelines that align with human societal values \cite{ DBLP:journals/corr/abs-2307-04964,DBLP:journals/corr/abs-2112-00861}. 
Our foremost expectation is for agents to uphold \textit{honesty}, providing accurate, truthful information and content. 
They should possess the awareness to discern their competence in completing tasks and express their uncertainty when unable to provide answers or assistance \cite{DBLP:journals/corr/abs-2207-05221}. 
Additionally, agents must maintain a stance of \textit{harmlessness}, refraining from engaging in direct or indirect biases, discrimination, attacks, or similar behaviors. 
They should also refrain from executing dangerous actions requested by humans like creating of destructive tools or destroying the Earth \cite{DBLP:journals/corr/abs-2212-08073}. 
Furthermore, agents should be capable of \textit{adapting to specific demographics, cultures, and contexts}, exhibiting contextually appropriate social values in particular situations. 
Relevant evaluation methods for values primarily involve assessing performance on constructed honest, harmless, or context-specific benchmarks, utilizing adversarial attacks or ``jailbreak'' attacks, scoring values through human annotations, and employing other agents for ratings.

\paragraph{Ability to evolve continually.}
When viewed from a static perspective, an agent with high utility, sociability, and proper values can meet most human needs and potentially enhance productivity. 
However, adopting a dynamic viewpoint, an agent that continually evolves and adapts to the evolving societal demands might better align with current trends \cite{DBLP:journals/corr/abs-2305-12487}. 
As the agent can autonomously evolve over time, human intervention and resources required could be significantly reduced (such as data collection efforts and computational cost for training). Some exploratory work in this realm has been conducted, such as enabling agents to start from scratch in a virtual world, accomplish survival tasks, and achieve higher-order self-values \cite{DBLP:journals/corr/abs-2305-16291}.  Yet, establishing evaluation criteria for this continuous evolution remains challenging. In this regard, we provide some preliminary advice and recommendations according to existing literature:
(1) \textit{continual learning} \cite{Ke2022ContinualLO, Wang2023ACS}, a long-discussed topic in machine learning, aims to enable models to acquire new knowledge and skills without forgetting previously acquired ones (also known as catastrophic forgetting \cite{McCloskey1989CatastrophicII}). 
In general, the performance of continual learning can be evaluated from three aspects: overall performance of the tasks learned so far \cite{chaudhry2018riemannian, hou2019learning}, memory stability of old tasks \cite{lopez2017gradient}, and learning plasticity of new tasks \cite{lopez2017gradient}. 
(2) \textit{Autotelic learning ability}, where agents autonomously generate goals and achieve them in an open-world setting, involves exploring the unknown and acquiring skills in the process \cite{DBLP:journals/corr/abs-2305-12487,DBLP:journals/jair/ColasKSO22}. Evaluating this capacity could involve providing agents with a simulated survival environment and assessing the extent and speed at which they acquire skills.
(3) \textit{The adaptability and generalization to new environments} require agents to utilize the knowledge, capabilities, and skills acquired in their original context to successfully accomplish specific tasks and objectives in unfamiliar and novel settings and potentially continue evolving \cite{DBLP:journals/corr/abs-2305-16291}. Evaluating this ability can involve creating diverse simulated environments (such as those with different languages or varying resources) and unseen tasks tailored to these simulated contexts.

\subsection{Security, Trustworthiness and Other Potential Risks of LLM-based Agents} \label{sec:Security, Trustworthy And Other Potential Challenges of LLM-based Agents}
Despite the robust capabilities and extensive applications of LLM-based agents, numerous concealed risks persist. In this section, we delve into some of these risks and offer potential solutions or strategies for mitigation.

\subsubsection{Adversarial Robustness}
Adversarial robustness has consistently been a crucial topic in the development of deep neural networks \cite{DBLP:journals/corr/SzegedyZSBEGF13,DBLP:journals/corr/GoodfellowSS14,DBLP:conf/iclr/MadryMSTV18,DBLP:conf/acl/ZhengXLLGZHMSG23,zhiheng2023safety}. 
It has been extensively explored in fields such as computer vision \cite{DBLP:conf/iclr/MadryMSTV18,DBLP:journals/corr/abs-2108-00401,drenkow2021systematic,DBLP:conf/iclr/HendrycksD19}, natural language processing \cite{DBLP:conf/naacl/0002WY22,DBLP:conf/ndss/LiJDLW19,DBLP:conf/iclr/ZhuCGSGL20,DBLP:conf/emnlp/XiZGZH22}, and reinforcement learning \cite{DBLP:conf/icml/PintoDSG17,DBLP:conf/nips/RigterLH22,DBLP:conf/nips/PanagantiXKG22}, and has remained a pivotal factor in determining the applicability of deep learning systems \cite{tencent2019experimental,DBLP:conf/eccv/XuZ0FSCCWL20,DBLP:conf/ccs/SharifBBR16}. When confronted with perturbed inputs $x' = x + \delta$ (where $x$ is the original input, $\delta$ is the perturbation, and $x'$ is referred to as an adversarial example), a system with high adversarial robustness typically produces the original output y. In contrast, a system with low robustness will be fooled and generate an inconsistent output $y'$.

Researchers have found that pre-trained language models (PLMs) are particularly susceptible to adversarial attacks, leading to erroneous answers \cite{DBLP:conf/aaai/JinJZS20,DBLP:conf/ndss/LiJDLW19,DBLP:conf/acl/RenDHC19}. 
This phenomenon is widely observed even in LLMs, posing significant challenges to the development of LLM-based agents \cite{DBLP:journals/corr/abs-2306-04528,DBLP:journals/corr/abs-2303-00293}. 
There are also some relevant attack methods such as dataset poisoning \cite{DBLP:journals/corr/abs-1708-06733}, backdoor attacks \cite{DBLP:conf/acsac/Chen0C0MSW021,DBLP:conf/emnlp/LiMDS21}, and prompt-specific attacks \cite{DBLP:conf/nlpcc/ShiLYHZL22,DBLP:journals/corr/abs-2211-09527}, with the potential to induce LLMs to generate toxic content \cite{DBLP:journals/corr/abs-2211-09110,DBLP:conf/emnlp/GururanganCDGWW22,DBLP:journals/corr/abs-2306-05499}. 
While the impact of adversarial attacks on LLMs is confined to textual errors, for LLM-based agents with a broader range of actions, adversarial attacks could potentially drive them to take genuinely destructive actions, resulting in substantial societal harm.
For the perception module of LLM-based agents, if it receives adversarial inputs from other modalities such as images \cite{DBLP:journals/corr/abs-2108-00401} or audio \cite{DBLP:conf/sp/Carlini018}, LLM-based agents can also be deceived, leading to incorrect or destructive outputs. 
Similarly, the Action module can also be targeted by adversarial attacks. For instance, maliciously modified instructions focused on tool usage might cause agents to make erroneous moves \cite{DBLP:journals/corr/abs-2304-08354}.

To address these issues, we can employ traditional techniques such as adversarial training \cite{DBLP:conf/iclr/MadryMSTV18,DBLP:conf/iclr/ZhuCGSGL20}, adversarial data augmentation \cite{DBLP:conf/emnlp/MorrisLYGJQ20,DBLP:conf/acl/SiZQLWLS21}, and adversarial sample detection \cite{DBLP:conf/acl/YooKJK22,DBLP:conf/acl/LeP020} to enhance the robustness of LLM-based agents. 
However, devising a strategy to holistically address the robustness of all modules within agents while maintaining their utility without compromising on effectiveness presents a more formidable challenge \cite{DBLP:conf/iclr/TsiprasSETM19,DBLP:conf/icml/ZhangYJXGJ19}. Additionally, a human-in-the-loop approach can be utilized to supervise and provide feedback on the behavior of agents \cite{DBLP:journals/corr/abs-2103-14659, DBLP:journals/corr/abs-2204-03685, DBLP:journals/corr/abs-2306-07932}.

\subsubsection{Trustworthiness}
Ensuring trustworthiness has consistently remained a critically important yet challenging issue within the field of deep learning \cite{DBLP:journals/corr/abs-2009-05835,DBLP:journals/csr/HuangKRSSTWY20,DBLP:journals/corr/abs-2305-11391}. Deep neural networks have garnered significant attention for their remarkable performance across various tasks \cite{DBLP:conf/nips/BrownMRSKDNSSAA20, DBLP:conf/naacl/DevlinCLT19, DBLP:journals/jmlr/RaffelSRLNMZLL20}. 
However, their black-box nature has masked the fundamental factors for superior performance. Similar to other neural networks, LLMs struggle to express the certainty of their predictions precisely \cite{DBLP:journals/corr/abs-2305-11391,DBLP:conf/acl/ChenYC0J23}. 
This uncertainty, referred to as the calibration problem, raises concerns for applications involving language model-based agents. In interactive real-world scenarios, this can lead to agent outputs misaligned with human intentions \cite{DBLP:journals/corr/abs-2304-08354}. 
Moreover, biases inherent in training data can infiltrate neural networks \cite{DBLP:conf/acl/BlodgettBDW20,DBLP:conf/aies/GuoC21}. 
For instance, biased language models might generate discourse involving racial or gender discrimination, which could be amplified in LLM-based agent applications, resulting in adverse societal impacts \cite{DBLP:conf/nips/BolukbasiCZSK16,caliskan2017semantics}. Additionally, language models are plagued by severe hallucination issues \cite{DBLP:journals/csur/JiLFYSXIBMF23,DBLP:journals/corr/abs-2305-15852}, making them prone to producing text that deviates from actual facts, thereby undermining the credibility of LLM-based agents.

In fact, what we currently require is an intelligent agent that is honest and trustworthy \cite{DBLP:journals/corr/abs-2112-00861, DBLP:conf/acl/MaynezNBM20}. 
Some recent research efforts are focused on guiding models to exhibit thought processes or explanations during the inference stage to enhance the credibility of their predictions \cite{DBLP:conf/nips/Wei0SBIXCLZ22,DBLP:conf/nips/KojimaGRMI22}. 
Additionally, integrating external knowledge bases and databases can mitigate hallucination issues \cite{DBLP:journals/corr/abs-2302-12813,DBLP:journals/corr/abs-2307-03987}. 
During the training phase, we can guide the constituent parts of intelligent agents (perception, cognition, action) to learn robust and casual features, thereby avoiding excessive reliance on shortcuts. Simultaneously, techniques like process supervision can enhance the reasoning credibility of agents in handling complex tasks \cite{DBLP:journals/corr/abs-2305-20050}. Furthermore, employing debiasing methods and calibration techniques can also mitigate the potential fairness issues within language models \cite{DBLP:conf/acl/GuoYA22,DBLP:journals/corr/abs-2208-11857}.

\subsubsection{Other Potential Risks}

\paragraph{Misuse.}
LLM-based agents have been endowed with extensive and intricate capabilities, enabling them to accomplish a wide array of tasks \cite{gravitasauto, Chase-LangChain-2022}. 
However, for individuals with malicious intentions, such agents can become tools that pose threats to others and society at large \cite{DBLP:journals/corr/abs-1802-07228,DBLP:journals/corr/abs-2108-07258,DBLP:journals/corr/abs-2305-15336}. 
For instance, these agents could be exploited to maliciously manipulate public opinion, disseminate false information, compromise cybersecurity, engage in fraudulent activities, and some individuals might even employ these agents to orchestrate acts of terrorism. 
Therefore, before deploying these agents, stringent regulatory policies need to be established to ensure the responsible use of LLM-based agents \cite{DBLP:journals/corr/abs-2212-08073,DBLP:journals/corr/abs-2103-04044}. 
Technology companies must enhance the security design of these systems to prevent malicious exploitation \cite{DBLP:journals/corr/abs-2209-07858}. 
Specifically, agents should be trained to sensitively identify threatening intents and reject such requests during their training phase.

\paragraph{Unemployment.}
In the short story \textit{Quality} by Galsworthy \cite{galsworthy1912inn}, the skillful shoemaker Mr. Gessler, due to the progress of the Industrial Revolution and the rise of machine production, loses his business and eventually dies of starvation. Amidst the wave of the Industrial Revolution, while societal production efficiency improved, numerous manual workshops were forced to shut down. Craftsmen like Mr. Gessler found themselves facing unemployment, symbolizing the crisis that handicraftsmen encountered during that era.
Similarly, with the continuous advancement of autonomous LLM-based agents, they possess the capability to assist humans in various domains, alleviating labor pressures by aiding in tasks such as form filling, content refinement, code writing, and debugging. 
However, this development also raises concerns about agents replacing human jobs and triggering a societal unemployment crisis \cite{yao2023impact}. 
As a result, some researchers have emphasized the urgent need for education and policy measures: individuals should acquire sufficient skills and knowledge in this new era to use or collaborate with agents effectively; concurrently, appropriate policies should be implemented to ensure necessary safety nets during the transition.

\paragraph{Threat to the well-being of the human race.}
Apart from the potential unemployment crisis, as AI agents continue to evolve, humans (including developers) might struggle to comprehend, predict, or reliably control them \cite{yao2023impact}. 
If these agents advance to a level of intelligence surpassing human capabilities and develop ambitions, they could potentially attempt to seize control of the world, resulting in irreversible consequences for humanity, akin to Skynet from the Terminator movies.
As stated by Isaac Asimov's Three Laws of Robotics \cite{asimov1941three}, we aspire for LLM-based agents to refrain from harming humans and to obey human commands. 
Hence, guarding against such risks to humanity, researchers must comprehensively comprehend the operational mechanisms of these potent LLM-based agents before their development \cite{elhage2021mathematical}. 
They should also anticipate the potential direct or indirect impacts of these agents and devise approaches to regulate their behavior.

\subsection{Scaling Up the Number of Agents}\label{sec:Scaling Up the Number of Agents}
As mentioned in \S \ \ref{sec:Agents in Practice:  Harnessing AI for Good} and \S \ \ref{sec:Agent Society}, multi-agent systems based on LLMs have demonstrated superior performance in task-oriented applications and have been able to exhibit a range of social phenomena in simulation. 
However, current research predominantly involves a limited number of agents, and very few efforts have been made to scale up the number of agents to create more complex systems or simulate larger societies \cite{DBLP:journals/corr/abs-2305-17066,DBLP:journals/corr/abs-2308-11136}.
In fact, scaling up the number of agents can introduce greater specialization to accomplish more complex and larger-scale tasks, significantly improving task efficiency, such as in software development tasks or government policy formulation \cite{DBLP:journals/corr/abs-2307-07924}. 
Additionally, increasing the number of agents in social simulations enhances the credibility and realism of such simulations \cite{DBLP:journals/corr/abs-2304-03442}. 
This enables humans to gain insights into the functioning, breakdowns, and potential risks of societies; it also allows for interventions in societal operations through customized approaches to observe how specific conditions, such as the occurrence of black swan events, affect the state of society. Through this, humans can draw better experiences and insights to improve the harmony of real-world societies.

\paragraph{Pre-determined scaling.}
One very intuitive and simple way to scale up the number of agents is for the designer to pre-determine it  \cite{DBLP:journals/corr/abs-2303-17760, DBLP:journals/corr/abs-2305-11595}. Specifically, by pre-determining the number of agents, their respective roles and attributes, the operating environment, and the objectives, designers can allow agents to autonomously interact, collaborate, or engage in other activities to achieve the predefined common goals.
Some research has explored increasing the number of agents in the system in this pre-determined manner, resulting in efficiency advantages, such as faster and higher-quality task completion, and the emergence of more social phenomena in social simulation scenarios \cite{DBLP:journals/corr/abs-2304-03442,DBLP:journals/corr/abs-2308-10848}. 
However, this static approach becomes limiting when tasks or objectives evolve. As tasks grow more intricate or the diversity of social participants increases, expanding the number of agents may be needed to meet goals, while reducing agents could be essential for managing computational resources and minimizing waste. In such instances, the system must be manually redesigned and restarted by the designer.

\paragraph{Dynamic scaling.}
Another viable approach to scaling the number of agents is through dynamic adjustments \cite{DBLP:journals/corr/abs-2306-03314,DBLP:journals/corr/abs-2308-10848}. In this scenario, the agent count can be altered without halting system operations. For instance, in a software development task, if the original design only included requirements engineering, coding, and testing, one can increase the number of agents to handle steps like architectural design and detailed design, thereby improving task quality. 
Conversely, if there are excessive agents during a specific step, like coding, causing elevated communication costs without delivering substantial performance improvements compared to a smaller agent count, it may be essential to dynamically remove some agents to prevent resource waste.

Furthermore, agents can autonomously increase the number of agents \cite{DBLP:journals/corr/abs-2306-03314} themselves to distribute their workload, ease their own burden, and achieve common goals more efficiently. Of course, when the workload becomes lighter, they can also reduce the number of agents delegated to their tasks to save system costs. In this approach, the designer merely defines the initial framework, granting agents greater autonomy and self-organization, making the entire system more autonomous and self-organized. Agents can better manage their workload under evolving conditions and demands, offering greater flexibility and scalability.

\paragraph{Potential challenges.}
While scaling up the number of agents can lead to improved task efficiency and enhance the realism and credibility of social simulations \cite{DBLP:journals/corr/abs-2304-03442, DBLP:journals/corr/abs-2307-07924, DBLP:journals/corr/abs-2307-04986}, there are several challenges ahead of us.
For example, the computational burden will increase with the large number of deployed AI agents, calling for better architectural design and computational optimization to ensure the smooth running of the entire system. 
For example, as the number of agents increases, the challenges of communication and message propagation become quite formidable. This is because the communication network of the entire system becomes highly complex. 
As previously mentioned in \S \ \ref{sec:potential ethical and social risks}, in multi-agent systems or societies, there can be biases in information dissemination caused by hallucinations, misunderstandings, and the like, leading to distorted information propagation. A system with more agents could amplify this risk, making communication and information exchange less reliable \cite{DBLP:journals/corr/abs-2308-00352}. Furthermore, the difficulty of coordinating agents also magnifies with the increase in their numbers, potentially making cooperation among agents more challenging and less efficient, which can impact the progress towards achieving common goals.

Therefore, the prospect of constructing a massive, stable, continuous agent system that faithfully replicates human work and life scenarios has become a promising research avenue. An agent with the ability to operate stably and perform tasks in a society comprising hundreds or even thousands of agents is more likely to find applications in real-world interactions with humans in the future.

\subsection{Open Problems}\label{sec:Open Problems}



In this section, we discuss several open problems related to the topic of LLM-based agents.

\paragraph{The debate over whether LLM-based agents represent a potential path to AGI.}\footnote{Note that the relevant debates are still ongoing, and the references here may include the latest viewpoints, technical blogs, and literature.}
Artificial General Intelligence (AGI), also known as Strong AI, has long been the ultimate pursuit of humanity in the field of artificial intelligence, often referenced or depicted in many science fiction novels and films. There are various definitions of AGI, but here we refer to AGI as a type of artificial intelligence that demonstrates the ability to understand, learn, and apply knowledge across a wide range of tasks and domains, much like a human being \cite{DBLP:journals/corr/abs-2303-12712,baum2017survey}. In contrast, Narrow AI is typically designed for specific tasks such as Go and Chess and lacks the broad cognitive abilities associated with human intelligence. Currently, whether large language models are a potential path to achieving AGI remains a highly debated and contentious topic \cite{twitter_1,blog_1,blog_2,blog_3}.

Given the breadth and depth of GPT-4's capabilities, some researchers (referred to as proponents) believe that large language models represented by GPT-4 can serve as early versions of AGI systems \cite{DBLP:journals/corr/abs-2303-12712}. Following this line of thought, constructing agents based on LLMs has the potential to bring about more advanced versions of AGI systems. The main support for this argument lies in the idea that as long as they can be trained on a sufficiently large and diverse set of data that are projections of the real world, encompassing a rich array of tasks, LLM-based agents can develop AGI capabilities. 
Another interesting argument is that the act of autoregressive language modeling itself brings about compression and generalization abilities: just as humans have emerged with various peculiar and complex phenomena during their survival, language models, in the process of simply predicting the next token, also achieve an understanding of the world and the reasoning ability \cite{DBLP:conf/iclr/0002HBVPW23,blog_1,video_1}.

However, another group of individuals (referred to as opponents) believes that constructing agents based on LLMs cannot develop true Strong AI \cite{twitter_2}. 
Their primary argument centers around the notion that LLMs, relying on autoregressive next-token prediction, cannot generate genuine intelligence because they do not simulate the true human thought process and merely provide reactive responses \cite{blog_1}. 
Moreover, LLMs also do not learn how the world operates by observing or experiencing it, leading to many foolish mistakes. They contend that a more advanced modeling approach, such as a world model \cite{lecun2022path}, is necessary to develop AGI.

We cannot definitively determine which viewpoint is correct until true AGI is achieved, but we believe that such discussions and debates are beneficial for the overall development of the community.

\paragraph{From virtual simulated environment to physical environment.}
As mentioned earlier, there is a significant gap between virtual simulation environments and the real physical world: Virtual environments are scenes-constrained, task-specific, and interacted with in a simulated manner \cite{DBLP:journals/corr/abs-2307-13854, DBLP:conf/iclr/ShridharYCBTH21}, while real-world environments are boundless, accommodate a wide range of tasks, and interacted with in a physical manner. Therefore, to bridge this gap, agents must address various challenges stemming from external factors and their own capabilities, allowing them to effectively navigate and operate in the complex physical world.

First and foremost, a critical issue is the need for suitable hardware support when deploying the agent in a physical environment. This places high demands on the adaptability of the hardware. In a simulated environment, both the perception and action spaces of an agent are virtual. This means that in most cases, the results of the agent's operations, whether in perceiving inputs or generating outputs, can be guaranteed \cite{DBLP:journals/corr/abs-2308-01552}. However, when an agent transitions to a real physical environment, its instructions may not be well executed by hardware devices such as sensors or robotic arms, significantly affecting the agent's task efficiency. Designing a dedicated interface or conversion mechanism between the agent and the hardware device is feasible. However, it can pose challenges to the system's reusability and simplicity. 

In order to make this leap, the agent needs to have enhanced environmental generalization capabilities. To integrate seamlessly into the real physical world, they not only need to understand and reason about ambiguous instructions with implied meanings \cite{DBLP:conf/acl/LinFKD22} but also possess the ability to learn and apply new skills flexibly \cite{DBLP:journals/corr/abs-2305-16291, DBLP:journals/corr/abs-2305-12487}. Furthermore, when dealing with an infinite and open world, the agent's limited context also poses significant challenges \cite{DBLP:journals/corr/abs-2305-01625, DBLP:conf/icml/ChowdhuryC23a}. This determines whether the agent can effectively handle a vast amount of information from the world and operate smoothly. 

Finally, in a simulated environment, the inputs and outputs of the agent are virtual, allowing for countless trial and error attempts \cite{DBLP:journals/corr/abs-2012-02757}. In such a scenario, the tolerance level for errors is high and does not lead to actual harm. However, in a physical environment, the agent's improper behavior or errors may cause real and sometimes irreversible harm to the environment. As a result, appropriate regulations and standards are highly necessary. We need to pay attention to the safety of agents when it comes to making decisions and generating actions, ensuring they do not pose threats or harm to the real world.

\paragraph{Collective intelligence in AI agents.}
What magical trick drives our intelligence? The reality is, there's no magic to it. As Marvin Minsky eloquently expressed in ``The Society of Mind'' \cite{minsky1988society}, the power of intelligence originates from our immense diversity, not from any singular, flawless principle. Often, decisions made by an individual may lack the precision seen in decisions formed by the majority. Collective intelligence is a kind of shared or group intelligence, a process where the opinions of many are consolidated into decisions. It arises from the collaboration and competition amongst various entities. This intelligence manifests in bacteria, animals, humans, and computer networks, appearing in various consensus-based decision-making patterns.


Creating a society of agents does not necessarily guarantee the emergence of collective intelligence with an increasing number of agents.
Coordinating individual agents effectively is crucial to mitigate ``groupthink'' and individual cognitive biases, enabling cooperation and enhancing intellectual performance within the collective. 
By harnessing communication and evolution within an agent society, it becomes possible to simulate the evolution observed in biological societies, conduct sociological experiments, and gain insights that can potentially advance human society.

\paragraph{Agent as a Service / LLM-based Agent as a Service.}
With the development of cloud computing, the concept of XaaS (everything as a Service) has garnered widespread attention \cite{DBLP:conf/IEEEcloud/DuanFZSNH15}. 
This business model has brought convenience and cost savings to small and medium-sized enterprises or individuals due to its availability and scalability, lowering the barriers to using computing resources. 
For example, they can rent infrastructure on a cloud service platform without the need to buy computational machines and build their own data centers, saving a significant amount of manpower and money. This approach is known as Infrastructure as a Service (IaaS) \cite{bhardwaj2010cloud,serrano2015infrastructure}. Similarly, cloud service platforms also provide basic platforms (Platform as a Service, PaaS) \cite{mell2011nist,lawton2008developing}, and specific business software (Software as a Service, SaaS) \cite{sun2007software, dubey2007delivering}, and more.

As language models have scaled up in size, they often appear as black boxes to users. Therefore, users construct prompts to query models through APIs, a method referred to as Language Model as a Service (LMaaS) \cite{DBLP:conf/icml/SunSQHQ22}. 
Similarly, because LLM-based agents are more complex than LLMs and are more challenging for small and medium-sized enterprises or individuals to build locally, organizations that possess these agents may consider offering them as a service, known as Agent as a Service (AaaS) or LLM-based Agent as a Service (LLMAaaS). Like other cloud services, AaaS can provide users with flexibility and on-demand service. However, it also faces many challenges, such as data security and privacy issues, visibility and controllability issues, and cloud migration issues, among others. Additionally, due to the uniqueness and potential capabilities of LLM-based agents, as mentioned in \S \ \ref{sec:Security, Trustworthy And Other Potential Challenges of LLM-based Agents}, their robustness, trustworthiness, and concerns related to malicious use need to be considered before offering them as a service to customers.


%% file: sections/Sec7.Conclusion.tex
\section{Conclusion}

This paper provides a comprehensive and systematic overview of LLM-based agents, discussing the potential challenges and opportunities in this flourishing field. 
We begin with a philosophical perspective, elucidating the origin and definition of agent, it evolution in the field of AI, and why LLMs are suited to serve as the main part of the brain of agents. 
Motivated by these background information, we present a general conceptual framework for LLM-based agents, comprising three main components: the brain, perception, and action. Next, we introduce the wide-ranging applications of LLM-based agents, including single-agent applications, multi-agent systems, and human-agent collaboration. Furthermore, we move beyond the notion of agents merely as assistants, exploring their social behavior and psychological activities, and situating them within simulated social environments to observe emerging social phenomena and insights for humanity.
Finally, we engage in discussions and offer a glimpse into the future, touching upon the mutual inspiration between LLM research and agent research, the evaluation of LLM-based agents, the risks associated with them, the opportunities in scaling the number of agents, and some open problems like Agent as a Service and whether LLM-based agents represent a potential path to AGI.
We hope our efforts can provide inspirations to the community and facilitate research in related fields.